\documentclass[10pt,twocolumn,letterpaper]{article}

\usepackage[pagenumbers]{cvpr}      %

\usepackage{graphicx}
\usepackage{amsmath}
\usepackage{amssymb}
\usepackage{booktabs}
\usepackage{xcolor}
\usepackage{multirow}
\usepackage{bbm}

\usepackage[pagebackref,breaklinks,colorlinks]{hyperref}

\usepackage[capitalize]{cleveref}
\crefname{section}{Sec.}{Secs.}
\Crefname{section}{Section}{Sections}
\Crefname{table}{Table}{Tables}
\crefname{table}{Tab.}{Tabs.}

\def\cvprPaperID{0000} %
\def\confName{CVPR}
\def\confYear{2022}

\usepackage{soul}

\DeclareMathOperator{\erf}{erf}
\DeclareMathOperator{\Cov}{Cov}
\DeclareMathOperator{\StdDev}{StdDev}
\DeclareMathOperator{\Entropy}{Entropy}

\begin{document}

\title{Scene Uncertainty and the Wellington Posterior of Deterministic Image Classifiers}

\author{Stephanie Tsuei\\
UCLA\\
{\tt\small stephanietsuei@ucla.edu}
\and
Aditya Golatkar\\
UCLA\\
{\tt\small adityagolatkar@ucla.edu}
\and
Stefano Soatto\\
UCLA\\
{\tt\small soatto@cs.ucla.edu}
}
\maketitle

\begin{abstract}
We propose a method to estimate the uncertainty of the outcome of an image classifier on a given input datum. Deep neural networks commonly used for image classification are deterministic maps from an input image to an output class. As such, their outcome on a given datum involves no uncertainty, so we must specify what variability we are referring to when defining, measuring and interpreting uncertainty, and attributing ``confidence'' to the outcome. To this end, we introduce the Wellington Posterior, which is the distribution of outcomes that {\em would} have been obtained in response to data that {\em could} have been generated by the same scene that produced the given image. Since there are infinitely many scenes that could have generated any given image, the Wellington Posterior involves inductive transfer from scenes other than the one portrayed. We explore the use of data augmentation, dropout, ensembling, single-view reconstruction, and model linearization to compute a Wellington Posterior. Additional methods include the use of conditional generative models such as generative adversarial networks, neural radiance fields, and conditional prior networks. We test these methods against the empirical posterior obtained by performing inference on multiple images of the same underlying scene. These developments are only a small step towards assessing the reliability of deep network classifiers in a manner that is compatible with safety-critical applications and human interpretation.
\end{abstract}

\section{Introduction}
\label{sec:intro}

It appears that Deep Neural Networks (DNNs) can classify images as well as humans, at least as measured by popular benchmarks, yet small perturbations of the images can cause changes in the predicted class. Even excluding adversarial perturbations, simply classifying consecutive frames in a video shows variability inconsistent with the reported error rate 
(see \cref{fig:spread_is_uncertainty}). So, how much should we trust image classifiers? How {\em confident} should we be of the outcome they render on a given image?
There is a substantive literature on uncertainty quantification, including work characterizing the (epistemic and aleatoric) uncertainty of trained classifiers (Sect. \ref{sec:related}). Such uncertainty is a property of the {\em classifier}, not of the {\em outcome} of inference on particular datum. 
We are instead interested in ascertaining how confident to be in the response of a particular DNN model to a particular image, not generally how well the classifier performs on images from a given class:
Say we have an image $x$, and a DNN that computes a discriminant vector $y = f(x)\in {\mathbb R}^K$ with as many components as the number of classes $K$ ({\em e.g.,} logits or softmax vector) from which it returns the estimated label $\hat k = \arg\max_k y_k = $``cat.'' How sure are we that there is a cat in {\em this} image? %
If faced with the question ``are you sure?'' a human would take a second look, or capture a new image, to either confirm or profess doubt. But a DNN classifier would return the same answer, correct or not, since most real-world deep networks in use today are deterministic maps $f$ from the input $x$ to the output $y$. 

Since the classifier is deterministic, and the image is given, the first {\bf key question} we must address is {\em with respect to what variability should uncertainty be defined and evaluated.}  To this end, we introduce the Wellington Posterior (WP) of a deterministic image classifier, which is {\em the distribution of outcomes that {\em would} have been obtained in response to data that {\em could} have been generated by the same scene that produced the given image.} Simply put, there are no cats in images, only pixels. Cats are in the {\em scene}, about which images provide evidence. The question of whether we are ``sure'' of the outcome of the an image classifier is therefore of counterfactual nature: Had we been given {\em different} images of the same scene, would an image-based classifier have returned the same outcome?

Formally, given an image $x$, if we could compute the posterior probability of the estimated label, $\hat k$ from which we can then measure confidence intervals, entropy, and other statistics commonly used to measure uncertainty, what we are after is {\em not} $P(\hat k | x)$. Instead, it is $P(\hat k | \{x'\})$ where $x' \sim p(x' | S)$, where $S$ is the scene that yielded the given image $x$. Since the image $x$ is compatible with infinitely many scenes and we only have access to a single image at test-time, the true scene $S$ can not be known.\footnote{Consider the filming of the \emph{Lord of the Rings}, which used projection to make a subset of the adult-sized actors in a single scene appear to be the height of children.} Since $P(\hat k | x')$ is unknowable, we will settle for $p(\hat k | S(x))$, where $S(x)$ is a set containing one or more of the scenes that could have generated $x$. In the following, $p(\hat k | S(x))$ and $p(y | S(x))$ are both called the \emph{Wellington Posterior}.\footnote{We name the quantity $p(y |  S(x))$ after the Duke of Wellington's  suggestion that strategic decision making, needed to defeat Napoleon, required {\em guessing what is at the other side of the hill}. Such a guess cannot be based directly on the given data, for one cannot see behind the hill, but on hypotheses induced from by having seen behind {\em other} hills before.} The scene $S$, whether real or imagined, is the vehicle that allows one to {\em guess what is not known ($S$) from what is known ($x$)}, which is the inductive process. The crux of the matter to compute the Wellington Posterior $p(y|S(x))$ is to characterize the set of scenes (or at least {\em a} set of scenes) that could have generated the given image(s), which we discuss in Sect.~\ref{sec:method}.

\subsection{Related Work}
\label{sec:related}

There is a large amount of literature on uncertainty quantification for neural networks, especially in recent years. Uncertainty, and the methods that estimate and measure it, is often sorted into one of two categories: \emph{epistemic}, or a model's lack of knowledge, and \emph{aleatoric}, uncertainty inherent in the data-generating process. Many works, including ours, do not fit cleanly into either category. Below, we sort a sample of methods for uncertainty quantification in image classification by answering the question \emph{it estimates uncertainty with respect to what?}

\paragraph{To the weights.}
The data-fitting capacity of modern deep networks is so large that all epistemic uncertainty can be captured as uncertainties in the values of the weights. This is formalized in Bayesian neural nets \cite{BNNBook,BNN_survey,bayesian_svi,ll_svi}. At inference time, the output prediction and uncertainty is computed using Bayes rule, rather than a simple forward pass. This is computationally intractable for models the size of modern deep nets, but can be approximated using Monte-Carlo test-time dropout \cite{mc_dropout}, deep ensembles \cite{lakshminarayanan_simple_2017}, or even a mix of the two \cite{Durasov_2021_CVPR}.

\paragraph{To the pixels in the image.}
In image classification, the only aleatoric uncertainty considered is the noise caused by the cameras capturing the image. So far, there has been far less work estimating aleatoric uncertainty than epistemic uncertainty. Two methods that address this problem are Assumed Density Filtering \cite{gast_lightweight_2018} and test-time dropout \cite{kendall_what_2017}. \cite{loquercio_general_2020} implements both \cite{gast_lightweight_2018} and \cite{mc_dropout} to estimate both aleatoric and epistemic uncertainty.

\paragraph{To a test dataset.}
Works on calibration ignore the distinction between aleatoric and epistemic uncertainty and focus on achieving the frequentist notion of \emph{calibration} --- confidence scores are calibrated when exactly X\% of all samples given a confidence of X\% are correctly classified. The quality of a network's confidence scores is given by the Expected Calibration Error (ECE) metric and several other variants \cite{nixon_measuring_2019}.

It was first noted in \cite{guo_calibration_2017} that modern deep neural networks trained with the cross-entropy loss are incredibly overconfident when the maximum value of the softmax vector is taken as a measure of confidence. \cite{guo_calibration_2017} compared and proposed several methods for post-hoc adjustment of the softmax vector. It is also common to use Monte-Carlo dropout \cite{mc_dropout} and deep ensembles \cite{lakshminarayanan_simple_2017} to compute more calibrated softmax vectors as well. More recent work has focused on ensuring that calibrated confidences remain calibrated in the presence of OOD data \cite{ovadia_can_2019, Tomani_2021_CVPR, pmlr-v139-zhou21b}. Other work has extended calibration to conformal prediction \cite{angelopoulos_uncertainty_2020, barber_predictive_2021}, where a model predicts a set of classes whose probabilities sum to a desired confidence level rather than a single class.

\paragraph{Other categories.}
\cite{malinin_predictive_2018, evidential_deep_learning} incorporate the likelihood that a test image is from the same distribution as the training images into the training and inference process. \cite{EDL_with_GANs} uses GANs to improve upon \cite{evidential_deep_learning}. A common feature of these methods is interpreting the softmax vector as the parameters of a Dirichlet distribution rather than a categorical distribution. \cite{pmlr-v119-joo20a} models uncertainty to specifically address noise and errors in ground-truth labels.

\paragraph{To the scene.}
The counterfactual ideas we present in this work have appeared elsewhere, but have not been formalized. \cite{hendrycks_benchmarking_2018} and \cite{imagenetvidrobust} find that image classifiers are not able to correctly classify all the frames of short videos. \cite{wang_aleatoric_2019} use test-time data augmentation in the same way we do, except in the context of segmentation of medical images. Related in spirit to our approach is \cite{CLUE}, which adds a layer of counterfactual reasoning to quantify the sensitivity of uncertainty estimates to changes in the input.

\subsection{Contributions}

Unlike work described above, ours does not propose a new measure of uncertainty nor a new way to calibrate the discriminant to match empirical statistics: We use standard statistics computed from the ``posterior probability'', such as covariance and entropy, to measure uncertainty. The core of our work aims to specify {\em with respect to what posterior} to measure uncertainty. Since deep network image classifiers used in the real world are deterministic, the choice is consequential, and yet seldom addressed explicitly in the existing literature. The Wellington Posterior is introduced to explicitly characterize the variability with respect to which uncertainty is measured. This is our {\bf first contribution}.

The Wellington Posterior is not general and does not apply to any data type. It is specific to images of natural scenes. No matter how many images we are given at inference time, there are infinitely many different scenes that could have generated them, including real and virtual scenes. For example, consider an image of a person standing in a room vs. a person standing in front of a background image of a room. Therefore, there are many different ways of constructing the  Wellington Posterior. Our \textbf{second contribution} is a description of possible ways to compute the Wellington Posterior and experiments that their accuracy. For the datasets and methods we choose, we \emph{compare the accuracy of each method against an empirical paragon}. Lessons learned from experiments detailed in Section \ref{sec:experiments} and Appendix \ref{sec:scene_entropy} are the \textbf{third contribution}.

\section{Method}
\label{sec:method}

We start by introducing the nomenclature used throughout the rest of the paper:\\
-- The {\bf scene} $S$ is an abstraction of the physical world. It can be thought of as a sample from some distribution $S\sim p_S$ that is unknown and arguably unknowable. The scene itself (not just its distribution) is arguably unknowable but for some of its ``attributes'' manifest in sensory data.\\
-- An {\bf attribute} $k$ is a characteristic of the scene that belongs to a finite set ({\em e.g.,} names of objects), $k(S) \in \{1, \dots, K\}$. Note that there can be many scenes that share the same attribute(s) ({\bf intrinsic variability}). For instance, $k$ can be the label ``cat'' and $p_S(\cdot|k)$ is the distribution of scenes that contain a ``cat.'' Continuous, but finitely-parametrized, attributes are also possible, for instance related to shape or illumination. \\
-- {\bf Extrinsic variability} $g_t$ is an unknown transformation of the scene that changes its manifestation (measurements, see next point) but not its attributes. It can be thought of as a sample from some nuisance distribution $g_t \sim p_g$. For instance, extrinsic variability could be due to the vantage point of the camera, the illumination, partial occlusion, sensor noise, quantization {\em etc.}, none of which depends on whether the scene is labeled ``cat''. Note that there can be spurious correlations between the attribute and nuisance variability: An indoor scene is more likely to contain a cat than a beach scene. Nevertheless, if there were a cat on the beach, we would want our classifier to say so with confidence. The fact that nuisance variables can correlate with attributes on a given dataset may engender confusion between intrinsic and extrinsic variability. To be clear, phenomena that generate intrinsic variability would not exist in the absence of the attribute of interest. The pose, color and shape of a cat do not exist without the cat. Conversely, ambient illumination (indoor vs. outdoor) exists regardless of whether there is a cat, even if correlated with its presence. The effect of nuisance variability on confidence, unlike intrinsic variability, is correlational, rather than causal. \\
-- A {\bf measurement} $x_t$ is a known function of both the scene $S$ (and therefore its attributes) and the nuisances. We will assume that there is a generative model that, if the scene $S$ was known, and if the nuisances $g_t$ were known, would yield a measurement up to some residual (white, zero-mean, homoscedastic Gaussian) noise $n_t$
    \begin{equation}
        x_t = h(g_t, S)+ n_t.
    \end{equation}
For example, $h$ can be thought of as a graphics engine where all variables on the right hand-side are given.\\
-- The {\bf discriminant} $y = f(x)$ is a deterministic function of the measurement that can be used to infer some attributes of the scene. For instance, $y = P(k | x) \in {\mathbb S}^{K-1}$ is the Bayesian discriminant (posterior probability). More in general, $y$ could be any element of a vector (embedding) space.\\
-- The {\bf estimated class} $\hat k(x)$ is the outcome of a classifier, for instance $\hat k = \arg\max_k [f(x)]_k$.
Given an image $x$ and a classifier $\hat k(\cdot)$, we reduce questions of confidence and uncertainty to the posterior probability $P(\hat k = k | x)$.  In the absence of any variability in the estimator $f$, defining uncertainty in the estimate $\hat k$ requires {\em assuming} some kind of variability. The Wellington Posterior hinges on the following assumptions:
\begin{itemize}
    \item The class $k$ is an attribute of the scene $S$ and is independent of intrinsic and extrinsic variability, by their definition.
    \item We posit that, when asking ``how confident we are about the class $\hat k(x)$'' we do {\em not} refer to the uncertainty of the class given that image, which is zero. Instead, we refer to uncertainty of the estimated class {\em with respect to the variability of all possible images of the same scene $S$ which could have been obtained} by changing nuisance (extrinsic) variability.
\end{itemize}
In other words, if in response to an image, a classifier returns the label ``cat,'' the question is {\em not} how sure to be about whether there is a cat in the image.  The question is how sure to be that there is a cat {\em in the scene portrayed by the image.} For instance, if instead of the given image, one was given a slightly different one, captured slightly earlier or a little later, and the classifier returned ``dog,'' would one be less confident in the answer than if it had also returned ``cat''?  Intuitively yes. Hypothetical repeated trials would involve not running the same image through the classifier over and over, but capturing different images of the same scene, and running each through the classifier. Of course, different images obtained by adding  noise would be a special case where the world is static and the only nuisance variability is due to sensor noise.

Intrinsic variability does not figure in the definition of scene uncertainty or the Wellington Posterior. The fact that we are given {\em one} image implies that we are interested in {\em one} scene, the one portrayed in the image. Even though there are infinitely many scenes compatible with it, the given image defines an equivalence class of extrinsic variability. So, the question of how sure we are of the answer ``cat'' given an image is not how frequently the classifier correctly returns the label ``cat'' on different images {\em of different scenes} that contain different cats. It is a question about {\em the particular scene portrayed by the image we are given}, with the given cat in it. The goal is {\em not}  $P(\hat k | k)$, which would be how frequently we say ``cat'' when there is one (in some scene). We are interested in {\em this} scene, the one portrayed by the image. Written as a Markov chain we have
\begin{equation}
    k \rightarrow S \rightarrow x \rightarrow y \rightarrow \hat k
\end{equation}
where the first arrow includes intrinsic variability (a particular attribute is shared by many scenes) and the second arrow includes nuisance/extrinsic variability (a particular scene can generate infinitely many images). The last two arrows are deterministic. We are interested {\em only} in the variability in the second arrow. To compute the Wellington Posterior $P(\hat k | S(x))$, we observe that
\begin{equation}
\begin{aligned}
    &P(\hat k = k | S(x)) \\ &= \int P(\hat k | x) dP(x |S) \\ &= 
    \int \delta(\hat k(x)-k) dP(x|S) \\ 
    &= \int \delta(\arg\min f(\underbrace{h(g, S)}_x) - k)dP(g) 
    \\ &\simeq \frac{1}{T}\sum_{t=1}^T \delta(\arg\min f(\underbrace{h(g_t, S)}_{x_t})-k).
\label{eq:EmpiricalSceneConfidence} 
\end{aligned}
\end{equation}
That is, given samples from the nuisance variability $g_t$, or sample images generated by changing nuisance variability, $x_t$, we can compute the probability of a particular label by counting the frequency of that label in response to different nuisance variability. We defer the question of whether the samples given are fair or sufficiently exciting. In the expression above, $f$ is computed by the given DNN classifier, $g_t$ is from a chosen class of nuisance transformations, and $h$ is an image formation model that is also chosen by the designer of the experiment. What remains to be determined is how to create a set of scenes $S(x)$ from the given image $x$.

As we defined it, the scene is an abstraction of the physical world. Such abstraction can live inside the memory of a computer. Since a scene is only observed through images of it, if a synthetic scene generates images that are indistinguishable from those captured of a physical scene, the real and synthetic scenes -- while entirely different objects -- are equivalent for the purpose of computing the Wellington Posterior. Thus a ``scene'' could be any generative model that can produce images that are indistinguishable from real ones {\em including} the given one. Different images are then obtained by perturbing the scene with nuisance variability.

Depending on how we measure ``indistinguishable,'' and how sophisticated the class of nuisance variables and interventions we adopt, we have a hierarchy of increasingly complex models. In addition, depending on how broadly we sample nuisance variables, that is depending on $p(g)$, we have a more or less representative sample from the Wellington Posterior. Below we outline the modeling choices tested in Sect. \ref{sec:experiments} and in the Appendix.

\subsection{Proposed Measures of Scene Uncertainty}
\label{sec:measuring_uncertainty}

We propose to use the following as measures of uncertainty that can be computed from the Wellington Posterior. In the text below, assume that for a sample of $N$ images from a single scene, we have logits $y_i$, softmax vectors $s_i$, and predictions $\hat k_i$ for $i=1, \dots N$.
\begin{itemize}
    \item \textbf{Logit Spread:} Frobenius norm of the covariance matrix of the logits, $\| \Cov(y) \|_F$.
    \item \textbf{Softmax Spread:} Standard deviation of cosine similarity distance between the softmax vectors, $\StdDev(\arccos(s \cdot \bar s))$, where $\bar s$ is the mean softmax vector.
    \item \textbf{Percent Non-Mode:} The percentage of predictions $\hat k_i$ that are not equal to $\bar k$, $P(\hat k \ne \bar k)$, where $\bar k$ is the most often predicted class within that scene.
    \item \textbf{Scene Entropy:} Entropy of the histogram of the $\hat k_i$ within a scene, $\Entropy(P(\hat k | S))$.
\end{itemize}

\subsection{Modeling the Wellington Posterior}
\label{sec:modeling}

This section details all the methods we use to model the Wellington Posterior in Section \ref{sec:experiments} and details others that we do not use. It is not our intention to give an exhaustive account of all possible ways, and it is beyond the scope of this paper to determine which is the ``best'' method, for that depends on the application ({\em e.g.}, closed-loop operation vs. batch post-processing), the available data ({\em e.g.}, a large static dataset vs. a simulator), the run-time constraints, etc.

\subsubsection{Image Generation Methods.}

\paragraph{Data Augmentation.} The simplest model we consider interprets the scene as a flat plane on which the given image is painted. Correspondingly, different images $x'$ can be generated from $x$ via {\em data augmentation}, which consists of typically small group transformations of the image itself: $\{x'_t = g_t x \ | \ g_t \in G\}$ where $G = {\cal A}({\mathbb R}^2)\times {\cal A}({\mathbb R})$ is the group of diffeomorphisms of the domain of the image, approximated by local affine planar transformations, and affine transformations of the range space of the image, also known as {\em contrast transformations}. These include small translations, rotations, and rescaling of the image, as well as changes of the colormap. In addition to affine domain and range transformations, one can also add i.i.d. Gaussian noise and paste small objects in the image, to simulate occlusion nuisances. Data augmentation is typically used to train classifiers to be robust, or insensitive, to the class of chosen transformations, which correspond to a rudimentary model of the scene, so we chose it as the {\em baseline} in our experiments in Sect. \ref{sec:experiments}. 

\paragraph{Explicit 3D scene reconstruction.} The model of the scene implicit in data augmentation does not take into account parallax, occlusions, and other fundamental phenomena of image formation. One step more general, we could use knowledge of the shape and topology of other scenes, manifest in a training set, to predict (one or more) scenes from a single image. Since there are infinitely many, the one(s) we predict is a function of the inductive biases implicit and explicit in the method. There there are hundreds of single-view reconstruction methods \cite{fu2021single}, including so-called ``photo pop-up'' \cite{hoiem2005automatic} used for digital refocusing and computational photography, well beyond what we can survey in this paper. %
Formally, $S(x) \sim p_S(\cdot | x, {\cal D})$ is a scene compatible with the given image $x$, which is represented as a distribution depending on a dataset ${\cal D}$ of scenes $S_i$ and corresponding images $\{x_{ij}\}_{j = 1}^{N_i}$ from it. These can be obtained through some ground truth mechanism, for instance a range sensor or a simulation engine.

\paragraph{Conditional Prior Networks.} Since the purpose of reconstructing the scene in three dimensions is to allow us to move the scene and warp the image, we could seek directly for warpings of the image that are compatible with likely 3D scenes. While, again, there are infinitely many such scenes, we can use a pre-trained network as inductive bias. Indeed, there are already methods to generate a distribution of optical flow fields $v:{\mathbb R}^2 \rightarrow {\mathbb R}^2$ that are compatible with the given image $x$, that is $P(v|x, {\cal D})$, called Conditional Prior Networks (CPNs) \cite{yang2018conditional}. %

\paragraph{Conditional Generative Adversarial Networks.} Finally, Conditional Generative Adversarial Networks, or any method that generates other similar images or video conditioned on an image, may also be a source of imputation \cite{clark2019adversarial}. In particular, \cite{cGAN} was designed for data augmentation.

\paragraph{Neural Radiance Fields (NERFs).} NERFs \cite{mildenhall_nerfs} are {\em transductive} generative models trained from multiple images of a specific scene, typically with known pose, that allow generating additional images from an arbitrary viewpoint, essentially by sampling and interpolating the plenoptic function \cite{adelson_book} without explicit representation of objects or even surfaces. Because of the Data Processing Inequality, the information contained in the infinitely many images that a NERF can generate is not one bit more than that contained in the images used for training it. From this perspective, a NERF is just a regularizer. If posed images were available, one could use them to compute the empirical distribution of the discriminant, possibly regularized with different methods, not necessarily those most suitable for rendering. The situation may chance as NERFs are further developed to embody {\em inductive} power, by generalizing across different scenes, but the state-of-the-art is still far from the point where we can instantiate a ``conditional NERF'' from a single sample, leveraging a training set of disjoint images, which is the setting for which the Wellington Posterior is defined.

\subsection{Model-Based Uncertainty.}
All the models above aim to generate a sample from the distribution $P(x'|S(x))$. Next, we note that any arbitrary additive perturbation in an input image $x$, $\Delta x$, may be equivalently expressed as an additive perturbation in the weights of a network's first layer, $\Delta w$. These additive perturbations may also be expressed as more complex perturbations in deeper layers of the network. This dual relationship between perturbations in the image and perturbations in the weights suggest that techniques for model-based uncertainty, such as Monte-Carlo Dropout \cite{mc_dropout} and deep ensembles \cite{lakshminarayanan_simple_2017}, may also be used to produce distributions $P(y|S(x))$ and $P(\hat k|S(x))$.

\paragraph{Linear Quadratic Fine-Tuning (LQF).}
\label{sec:LQF_main}

 Recent development in network linearization suggest that it is possible to perturb the weights of a trained network locally to perform novel tasks essentially as well as non-linear optimization/fine-tuning \cite{achille_lqf_2020}. Such linearization is with respect to perturbations of the weights of the model. Therefore, we use the closed-form analytical Jacobian of the linearized network to compute first and second-order statistics (mean and covariance) of the discriminant, without the need for sampling. The first-order approximation of the network around the pre-trained weights $w_0$ is
\begin{equation}
    f_{w}(x) = f_{w_0}(x) + \nabla_w f_{w_0}(x) \rvert_{w=w_0} (w - w_0).
    \label{eq:lqf_form}
\end{equation}
In equation \eqref{eq:lqf_form}, $\bar w := w - w_0$ has a closed-form solution that is a function of the training data. Then, equation \eqref{eq:lqf_form} can be used to define a stochastic network with weights distributed according to $w \sim \mathcal{N}(w_0 + {\bar{w}}; \Sigma_w)$ for any desired covariance $\Sigma_w$. Correspondingly, the discriminant and logits $y$ can be assumed to be distributed as
\begin{equation}
\begin{aligned}
    y &= \mathcal N(\bar y, \Sigma_y) \\
    \bar y &=  f_{w_0}(x) + \nabla_w f_{w_0}(x) \cdot \bar w \\
    \Sigma_y &= \nabla_w f_{w_0}(x) \cdot \Sigma_w \cdot \nabla_w f_{w_0}(x)^T
\end{aligned}
\label{eq:lqfcov}
\end{equation}
Given a diagonal or block-diagonal value of $\Sigma_w$, we compute $\Sigma_y$ using $2K$ forward passes and $K$ backward passes per input image over the neural network. This is not a fast computation, but requires far less resources than a naive implementation. More implementation details are given in \cref{sec:lqf_compute_sigmay}.

Next, given $\mathcal N(\bar y, \Sigma_y)$, the probability that class $\hat k$ is the predicted class is given by:
\begin{equation}
P(k=\hat k|S(x)) = \int \phi_{w,x}(y) \mathbbm{1}\{ \hat k=\arg\max(y) \} dy.
\label{eq:fullintegral}
\end{equation}
The above is a $K$-dimensional integral. The lower limit of integration in every dimension is $-\infty$ while the upper limit is variable. For $K=3$, $y=[y_1,y_2,y_3]$, and $\hat k=1$, equation \eqref{eq:fullintegral} is:

\begin{equation}
\begin{aligned}
    & \frac{1}{(2\pi)^{K/2} |\Sigma_y|^{1/2}} \int_{-\infty}^{\infty} \int_{-\infty}^{y_1} \dots \int_{-\infty}^{y_1} \cdot \\
    & \prod_{i=1}^K \exp{\left (  \frac{-1}{2 \sigma_i^2} (y_i - \mu_y)^2  \right )} d y_K \dots d y_2 dy_1
\end{aligned}
\end{equation}
Next, introduce the change of variables $e_i = y_i - \bar y_i$ and let $u_i = y_1 - \bar y_i$. The above integral then becomes:
\begin{equation}
\begin{aligned}
    & \frac{1}{(2\pi)^{K/2} |\Sigma_y|^{1/2}} \int_{-\infty}^{\infty} \int_{-\infty}^{u_2} \dots \int_{-\infty}^{u_K} \cdot \\
    & \prod_{i=1}^K \exp{\left (  \frac{-1}{2 \sigma_i^2} e_i^2  \right )} d e_K \dots d e_2 de_1 .
\end{aligned}
\label{eq:int0}
\end{equation}
Using the identity
\begin{equation}
    \int \exp \left ( a x^2 \right ) dx = \frac{\sqrt{\pi} \erf(\sqrt{a}x) }{ 2 \sqrt{a}}
\end{equation}
and the fact that $\erf(-\infty) = -1$ for $i=2, \dots K$, we arrive at the following value of $P(k=1|S(x))$:
\begin{equation}
\begin{aligned}
    & \frac{1}{(2\pi)^{K/2} |\Sigma_z|^{1/2}} \int_{-\infty}^{\infty} \exp{\left (  -\frac{1}{2 \sigma_{\hat k}^2}(z_{\hat k}-\bar z_{\hat k})^2 \right )} \cdot \\
    & \prod_{k \neq \hat k} \left ( \frac{\sqrt{\pi} \left ( \erf \left (\sqrt{\frac{1}{2 \sigma_k^2}} u_{k,\hat k} \right )  +1  \right )  }{2 \sqrt{\frac{1}{2 \sigma_k^2}}} \right ) dz_{\hat k}
    \label{eq:lqfintdiagapprox}
\end{aligned}
\end{equation}
\cref{eq:lqfintdiagapprox} has no closed form solution, but can be numerically evaluated using a quadrature method.

For values of $\Sigma_y$ that are not well-approximated by their diagonals, statistics of the distribution in equation \eqref{eq:lqfcov} can be approximated and compared them to the empirical paragon in Section \ref{sec:experiments} through sampling. This is, of course, not ideal, but computationally fast.

\subsection{Pseudo-Uncertainties}
\label{sec:pseudo_uncertainty}

It is common to interpret the value of the discriminant $y = f(x)$ as an approximation of the (log-)posterior probability over classes. However, this is incorrect since the Bayesian posterior is not a minimizer of the empirical risk \cite{hullermeier_aleatoric_2021}. Appendix \ref{sec:scene_entropy} details preliminary work showing that the discriminant $y$, and functions of the discriminant, can not represent the Wellington Posterior.

\section{Experiments}
\label{sec:experiments}

\paragraph{Datasets.}
We performed experiments on two datasets, Objectron \cite{objectron2021}, and the ImageNetVid \cite{imagenetvid} and ImageNetVid-Robust \cite{imagenetvidrobust} datasets. These datasets are ideal for our image classification experiments because they contain multiple images of the same scene and each scene contains a single object of interest that is visible in all the images.
ImageNetVid was created in 2015 and consists of videos with bounding boxes for 30 classes. In 2017, additional annotated videos were added to the ImageNetVid validation dataset and in 2019, a set of frames, mostly selected from the 2017 videos, were human-curated into the ImageNetVid-Robust dataset. Each ``video" in ImageNetVid-Robust consists of 21 frames: 20 regular frames and one designated ``anchor" frame. Temporally, the anchor frame is the middle of the other frames. Human curation guaranteed that the object of interest would be clearly visible in each frame and that the frames would be quite similar. We use ImageNetVid frames as a training set and ImageNetVid-Robust frames as a test set in our experiments. 

Objectron \cite{objectron2021} contains short video clips of scenes with 9 classes: Bikes, Books, Bottles, Cameras, Cereal Boxes, Chairs, Cups, Laptops, and Shoes. Videos in the dataset consist of humans moving a smartphone around a static object. Motions in all videos are quite similar: a human waves a smartphone around a stationary object with large changes in the azimuth angle between a point on the object and the camera, but smaller changes in the elevation between a point on the object and the camera and in the distance from the object to the camera. We selected a subset of 5000 videos for training set, 500 videos for validation set, and 500 for testing --- there is no overlap in video frames between training, validation and test splits. From each video clip, we selected 20 evenly-spaced frames to serve as an empirical paragon and designated the middle frame the anchor image.

In general, evenly distributed frames across a short video clip do not provide a fair sample from the population of all possible images that the scene could have generated, so any resulting statistics cannot be considered proper ground truth. A better experiment would use synthetically generated 3D scenes containing all possible intrinsic and extrinsic variability instead of natural videos. However, this would require manually creating thousands of 3D scenes, each focused on a different object; the cost of this task is prohibitive. Another possibility would be to create  a smaller group synthetic scenes containing the same classes of objects as Objectron and ImageNetVid and use domain adaptation to adapt the synthetic scenes to natural images; the cost prohibitive part would then become manually creating many extrinsic variabilities found in scenes. For example, creating a laptop scene for the Objectron dataset in this manner would require a 3D model of not only the laptop, but the table beneath it and decorations around it. Therefore, an empirical baseline of 20 uniformly sampled frames from each video serves as a first step in evaluating methods to predict scene uncertainty.

So that we would have a fairer test set on which to evaluate our methods, we created two additional tiny datasets to accompany the Objectron dataset: one synthetic and one real. Each dataset consists of two items from each class. The one consisting of real images ``MiniObjectron", has 20 images per scene, captured from a set of positions around the object with low precision. The one consisting of synthetic images, ``SyntheticObjectron", consists of 468 images per scene, captured from a set of positions around the object with perfect precision.

\paragraph{Base Discriminant.} In order to compute the empirical posterior and the various forms of Wellington Posterior, we need a discriminant function $f(\cdot)$, from which to build a classifier. We use the backbone of an ImageNet-pretrained ResNet-50 or ResNet-101 \cite{resnet}, where $f(x)$ is called the vector of logits, whose maximizer is the selected class, and whose normalized exponential is called softmax vector. Fine-tuning and validation for ResNet-50 on Objectron, and ResNet-101 on ImageNetVid, used one frame from each scene. The focus is not to achieve the highest possible accuracy, but to provide a meaningful estimate of uncertainty relative to the variability of different images of the same scene. For this reason, we select the most common, not the highest performing, baseline classifier. For reference, we achieve $>$ 95\% validation accuracy for Objectron and $\sim$ 75\% validation accuracy for ImageNetVid.

\subsection{Baseline Uncertainty Values}
\label{sec:exp_pred_spread_valid}

\begin{figure*}[htp]
\centering
\includegraphics[width=0.24\textwidth]{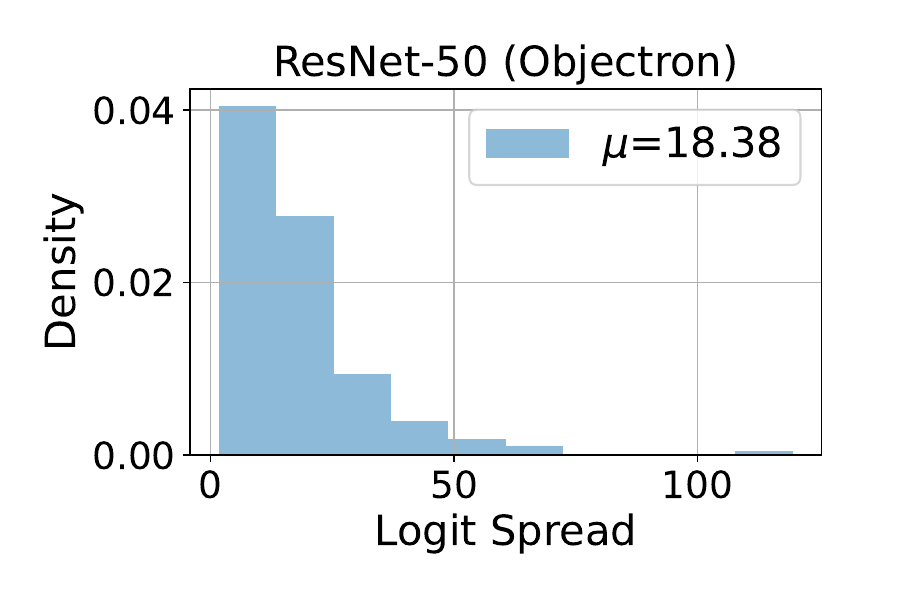}
\includegraphics[width=0.24\textwidth]{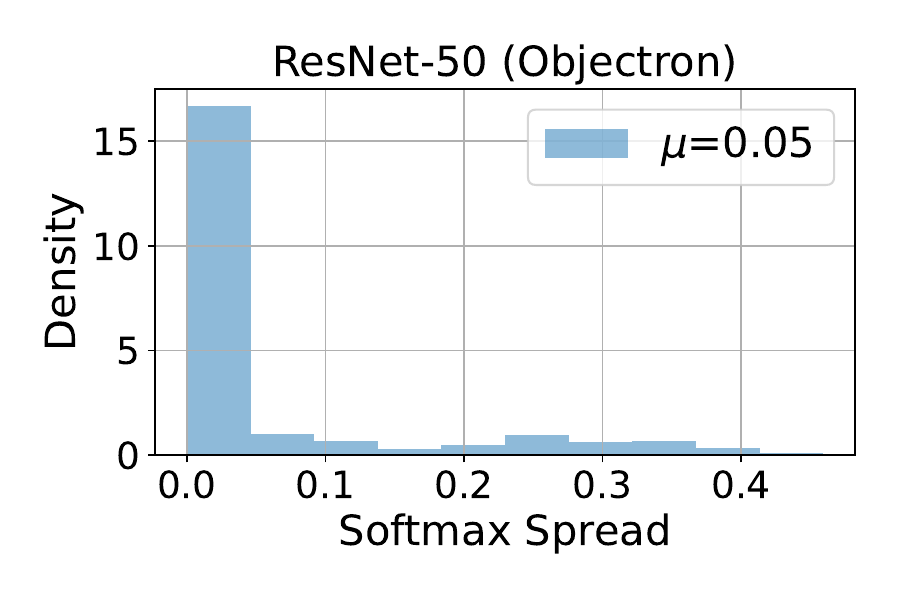}
\includegraphics[width=0.24\textwidth]{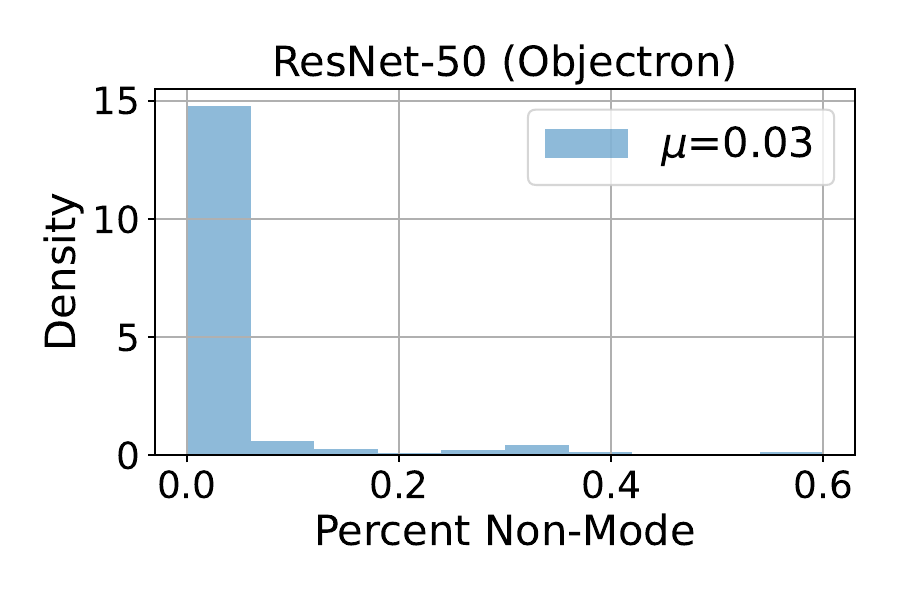}
\includegraphics[width=0.24\textwidth]{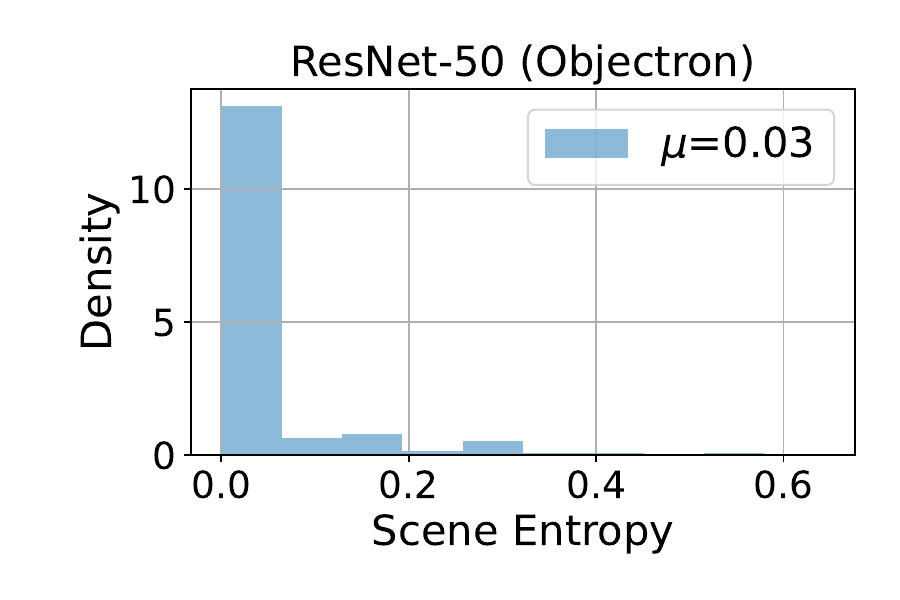}
\includegraphics[width=0.24\textwidth]{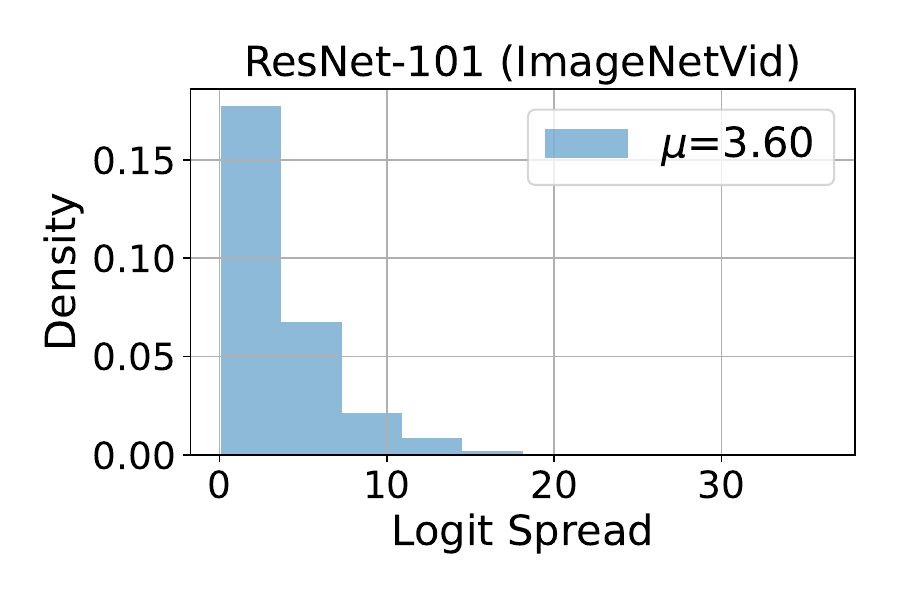}
\includegraphics[width=0.24\textwidth]{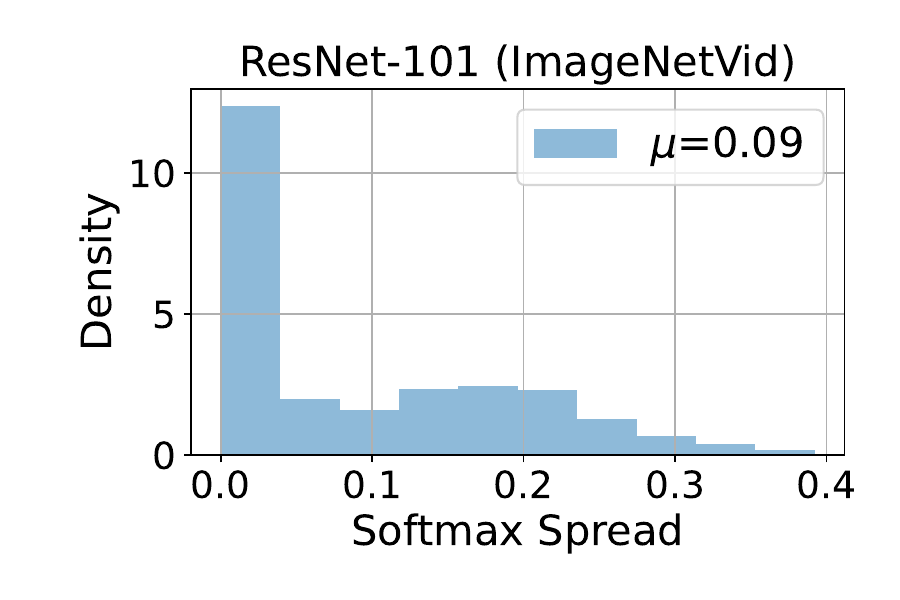}
\includegraphics[width=0.24\textwidth]{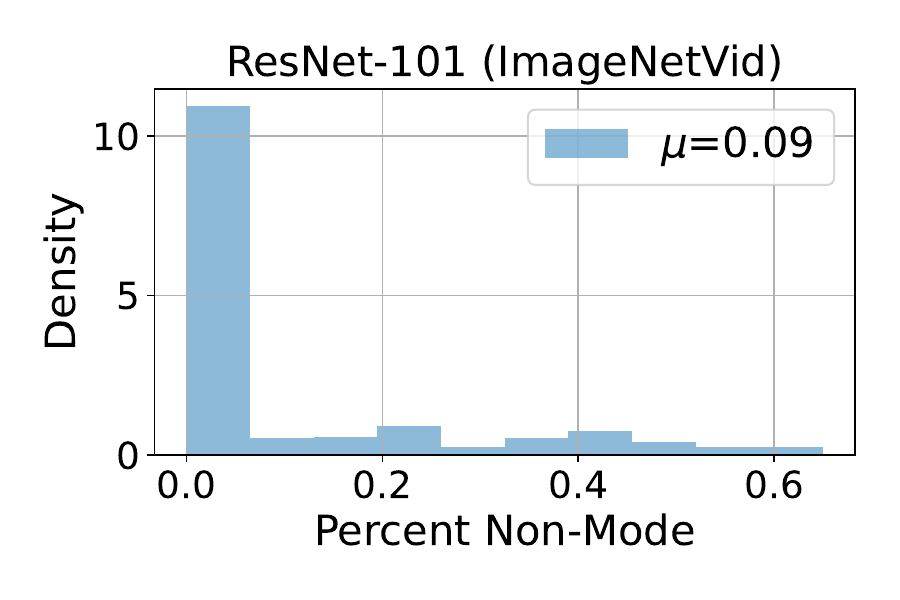}
\includegraphics[width=0.24\textwidth]{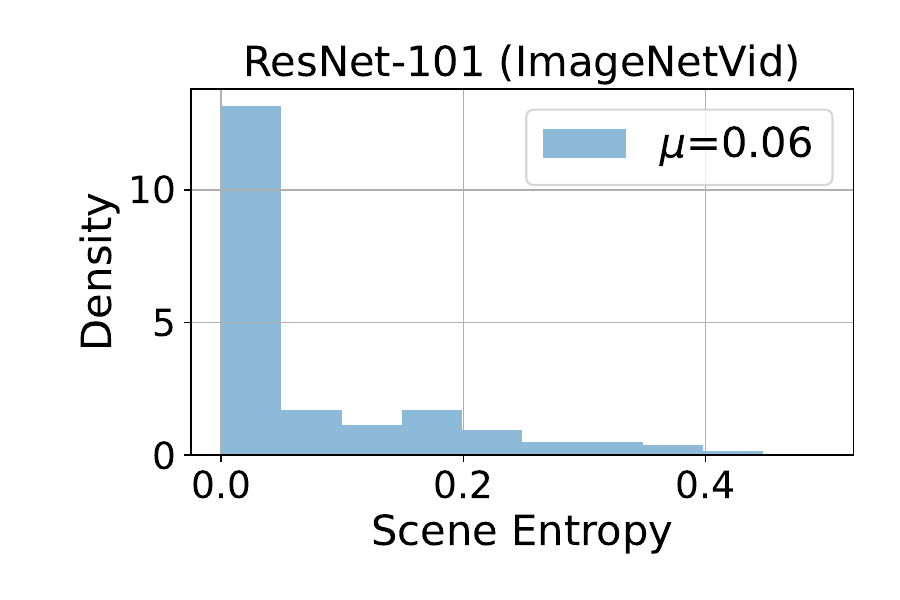}
\includegraphics[width=0.24\textwidth]{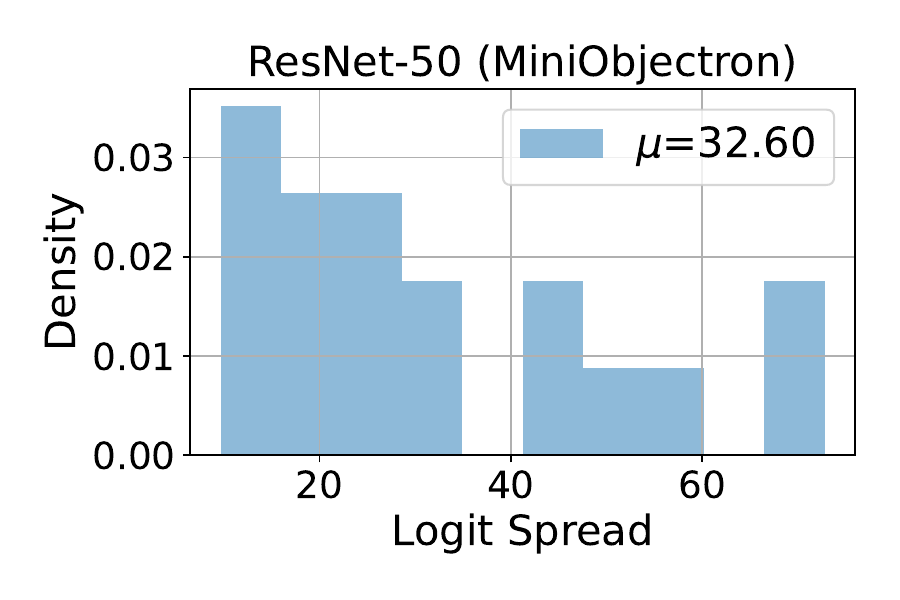}
\includegraphics[width=0.24\textwidth]{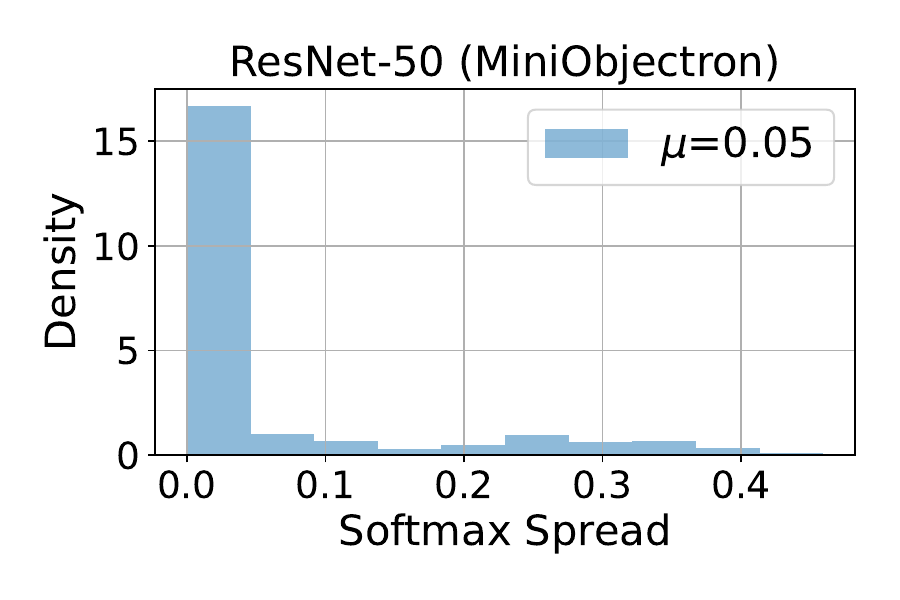}
\includegraphics[width=0.24\textwidth]{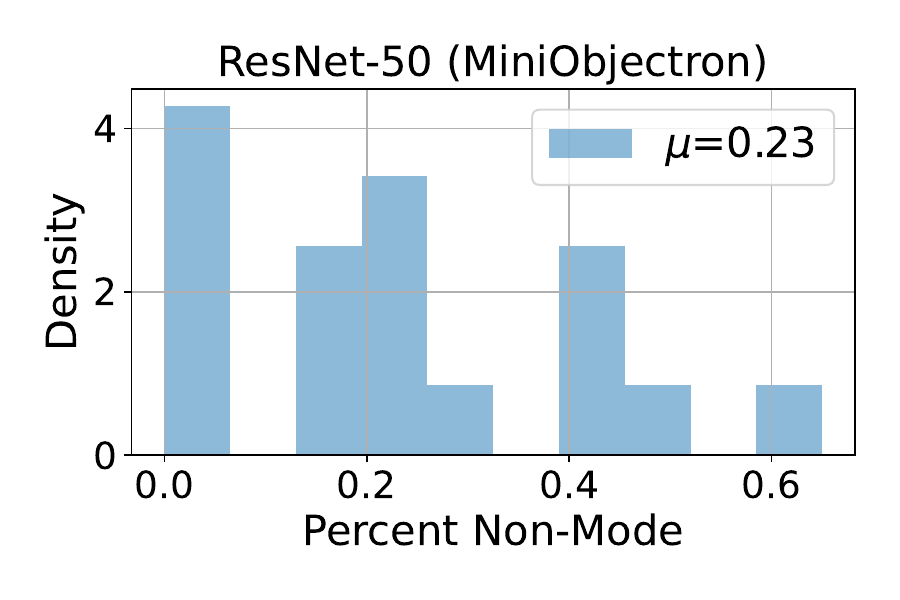}
\includegraphics[width=0.24\textwidth]{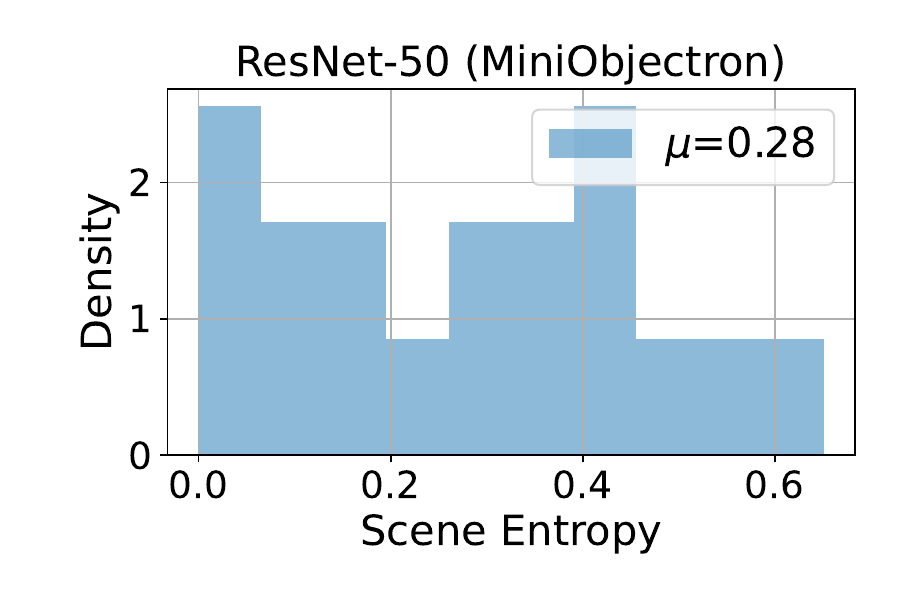}
\includegraphics[width=0.24\textwidth]{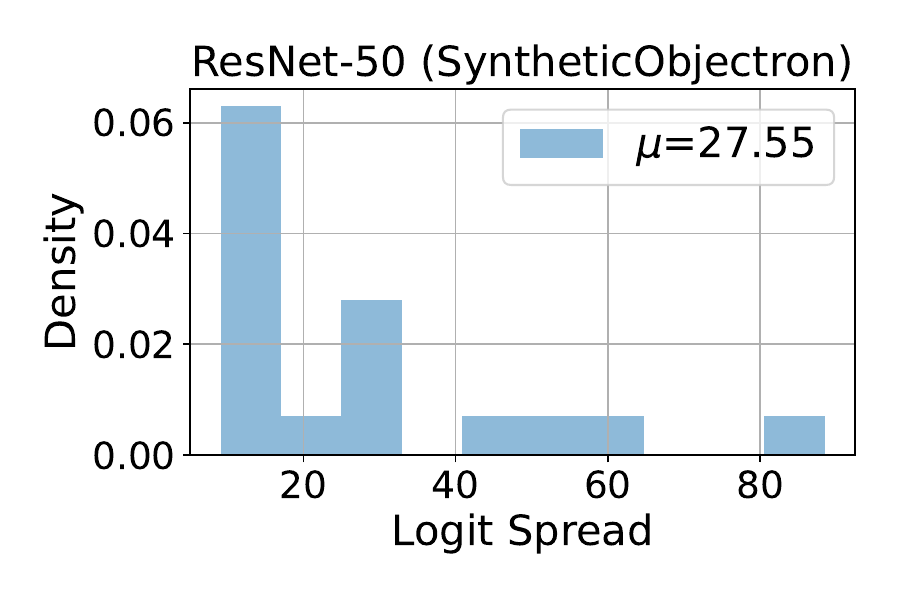}
\includegraphics[width=0.24\textwidth]{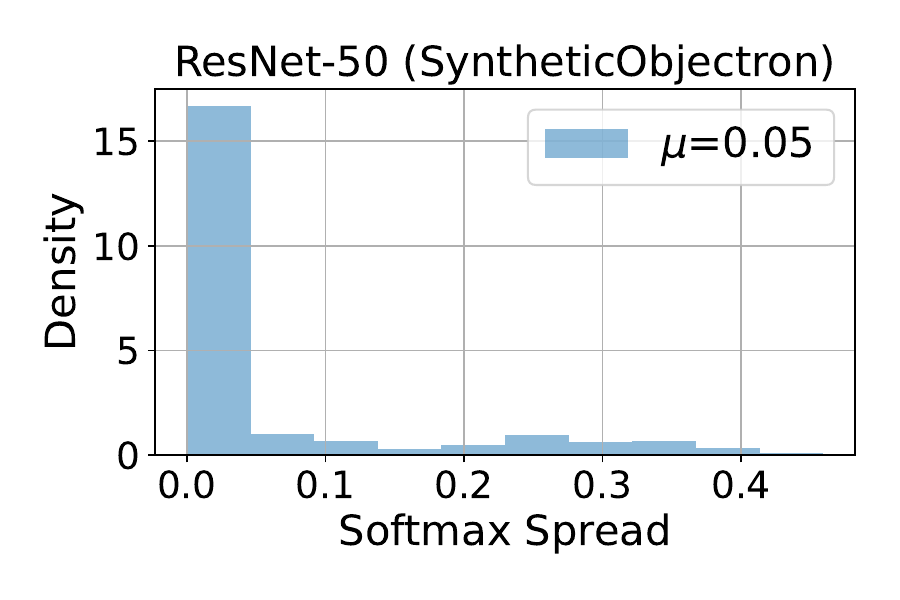}
\includegraphics[width=0.24\textwidth]{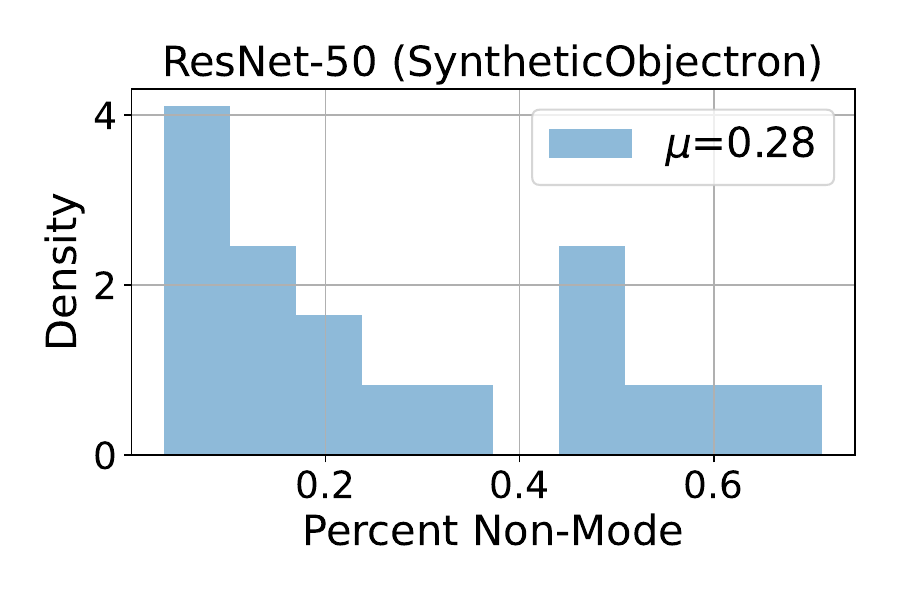}
\includegraphics[width=0.24\textwidth]{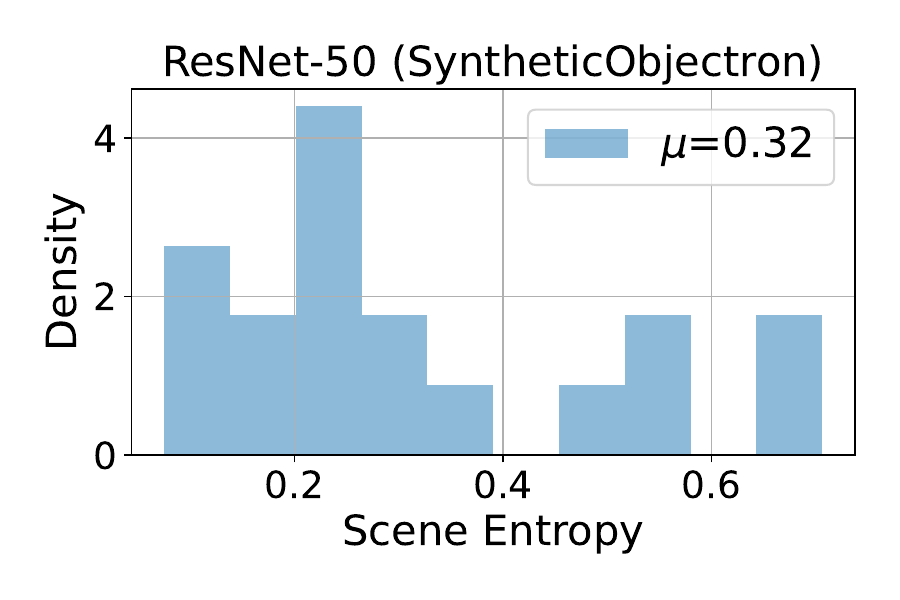}
\caption{
\textbf{Scene Uncertainty is Real.} Histograms of the four measures of scene uncertainty (logit spread, softmax spread, percent non-mode, and scene entropy) from \cref{sec:measuring_uncertainty} for ResNet-50 on Objectron (top row), ResNet-101 on ImageNetVid (second row), ResNet-50 on MiniObjectron (third row), and ResNet-50 on SyntheticObjectron (bottom row). Each point in the histograms above is a scene. The mean of each distribution is shown in the top-left. For the cases when the uncertainty of the network is measured over the test split of the same dataset it was trained on (top two rows) thick bars at the left of a histogram indicates that most images of most scenes are classified identically within a scene and that discriminants are similar. However, there is a long tail in the plots showing that scene uncertainty is a real phenomenon that will be encountered in real systems. In the bottom two rows, where the uncertainty of ResNet-50 was measured on our two ``more fair" constructed datasets, the measures of scene uncertainty are much higher.
}
\label{fig:spread_is_uncertainty}
\end{figure*}

We computed the measures of uncertainty proposed in \cref{sec:measuring_uncertainty} (logit spread, softmax spread, percent non-mode predictions, and scene entropy) over empirical paragon scenes to compute baseline uncertainty values. Baseline values are plotted as a histogram in \cref{fig:spread_is_uncertainty}. Means and standard deviation of baseline values from three separate experiments are shown in \cref{tab:vanilla_spread}. Both \cref{fig:spread_is_uncertainty} and \cref{tab:vanilla_spread} show that scene uncertainty is a real phenomena.

\begin{table*}[htp]
\small
    \centering
    \begin{tabular}{lcccc}
    \toprule
      & \textbf{ImageNetVid} & \textbf{Objectron} & \textbf{MiniObjectron} & \textbf{SyntheticObjectron} \\
    \midrule
    Logit Spread       & 0.5618 $\pm$ 0.0131 & 1.5803 $\pm$ 0.0077 & 32.8264 $\pm$ 0.5193 & 29.7664 $\pm$ 2.0210 \\
    Softmax Spread     & 0.0619 $\pm$ 0.0000 & 0.0418 $\pm$ 0.0000 & 0.2652 $\pm$ 0.0000 & 0.3142 $\pm$ 0.0000 \\
    Percent Non-Mode   & 0.0933 $\pm$ 0.0026 & 0.0315 $\pm$ 0.0019 & 0.2269 $\pm$ 0.0273 & 0.2908 $\pm$ 0.0125 \\
    Scene Entropy      & 0.2129 $\pm$ 0.0041 & 0.0774 $\pm$ 0.0036 & 0.2692 $\pm$ 0.0254 & 0.3403 $\pm$ 0.0135 \\
    \bottomrule
    \end{tabular}
    \caption{This table accompanies \cref{fig:spread_is_uncertainty}. Entries are means and standard deviations of proposed measures of uncertainty (logit spread, softmax spread, percent non-mode predictions, and scene entropy) computed using the empirical paragon. Means and standard deviations come from performing all calculations over all datasets three times using different trained networks. The only difference between networks is the data shuffling used at training time. 
    }
    \label{tab:vanilla_spread}
\end{table*}

\subsection{Computing the Wellington Posterior}
\label{sec:WP_performance}

In our second experiment, we compute Wellington Posteriors using data augmentation, 3D reconstruction, Monte-Carlo dropout, deep ensembles, and LQF. In all methods that required sampling, \textbf{20 generated samples formed the Wellington Posterior}.

We leave methods that require difficult fine-tuning and sampling from a distribution, such as conditional prior networks, and conditional GANs, to future work.Sample images generated by \cite{cGAN} from both ImageNetVid and Objectron are given in Section \ref{sec:gans_are_lulz}, illustrating the need for fine-tuning.

\paragraph{Data Augmentation.}
To generate multiple images from a single anchor frame, our data augmentation experiments uses horizontal flips with $p=0.5$, shifts by up to 20\% of image size in both horizontal and vertical directions, random uniform scaling between $[0.8, 1.2]$, a random rotation between $[-20,20]$ degrees, and a $224 \times 224$ center drop.
In order to avoid artifacts caused by rotations, we use the \texttt{SafeRotation} Albumentations imaging library \cite{buslaev_albumentations_2020} in place of the standard transformations available in Pytorch.

\paragraph{3D Reconstruction.}

For the Objectron dataset, we generated depth maps of anchor frames using \cite{VOICED}. Reprojected images came from perturbing the position of the camera by 5\% of the mean depth measurement in the image in all directions (X, Y, and Z) and by rotating the camera by up to 2 degrees in all directions (roll, pitch, and yaw). Note that Objectron is a 3D tracking dataset and \cite{VOICED} is a depth \emph{completion} network --- we leveraged the sparse depth measurements associated with the anchor frame to produce a depth map, but no other frames. We chose to use monocular depth completion instead of monocular depth estimation in order to maximize the quality of generated images using all the information available; this experiment should therefore be considered an upper-bound on the performance of single-view 3D reconstruction for Objectron.

Since the ImageNetVid-Robust dataset does not contain sparse depth measurements, we attempted to generate depth maps using a pretrained state-of-the-art monocular depth estimation network \cite{Yin2019enforcing}. The pretrained model consistently and incorrectly predicted a near constant depth for many images, showing that ImageNetVid-Robust images are too different from the images in the dataset on which the depth estimation network was trained. 

More details about training \cite{VOICED} on Objectron and on using \cite{Yin2019enforcing} on ImageNetVid are given in Appendix \ref{sec:3D_reproj_imgs}.

\paragraph{Dropout.} Our dropout experiments consisted of training a separate network with 5\% probability of dropout after ReLU activations within the basic and bottleneck blocks of the ResNet architecture. (This is the same probability of dropout used in \cite{ovadia_can_2019}.) Training parameters were otherwise the same as those used to train the baseline networks. To estimate uncertainty at test-time, we leave dropout enabled instead of rescaling the weights, as described in \cite{mc_dropout}. Multiple runs over the same images with dropout enabled then produces multiple discriminants and predictions to form a Wellington Posterior. %

\paragraph{Deep Ensembles.} Each deep ensemble contained 20 networks. Differences in the ensemble came from randomly withholding 5\% of the data and using a different random seed in the dataloader. We found that changing hyperparameters was not necessary to create variation in predictions and accuracy over the validation set.

\paragraph{Linear Quadratic Fine-Tuning.}
The key to using LQF to compute a distribution for the logits $y$ is to first have a value for the covariance of the fine-tuned weights, $\Sigma_w$.
We assume that $\Sigma_w$ is approximately diagonal and compute it as the sample variance of an ensemble of 20 LQF networks that each had a random 5\% of training data withheld. Once we compute a value of $\Sigma_y$ (see \cref{sec:lqf_compute_sigmay}), we can then almost compute $P(\hat k=k|S(x))$, the scene entropy and percent of non-mode predictions analytically using equation \eqref{eq:lqfintdiagapprox}. Since we do not know the distribution of the softmax function of a Gaussian random variable, we compute softmax spread by sampling logits from the distribution $\mathcal N(y, \Sigma_y)$ in order to generate a scene of softmax vectors. 

\paragraph{Metrics.}

We use two metrics to evaluate the ability of the methods in our modeling hierarchy to predict the Wellington Posterior for producing a measure of uncertainty (softmax spread, flip probability, and prediction entropy) against the empirical paragon.
\begin{enumerate}
    \item \textbf{Mean absolute error (MAE)} $(\varepsilon)$ between uncertainty measures for corresponding WP and empirical paragon over a testset
    \item \textbf{Correlation coefficient} $(\rho)$ between uncertainty measures for corresponding WP and empirical paragon over a testset
\end{enumerate}

\paragraph{Results.}

\begin{figure*}[ht]
    \centering
    \includegraphics[width=0.24\textwidth]{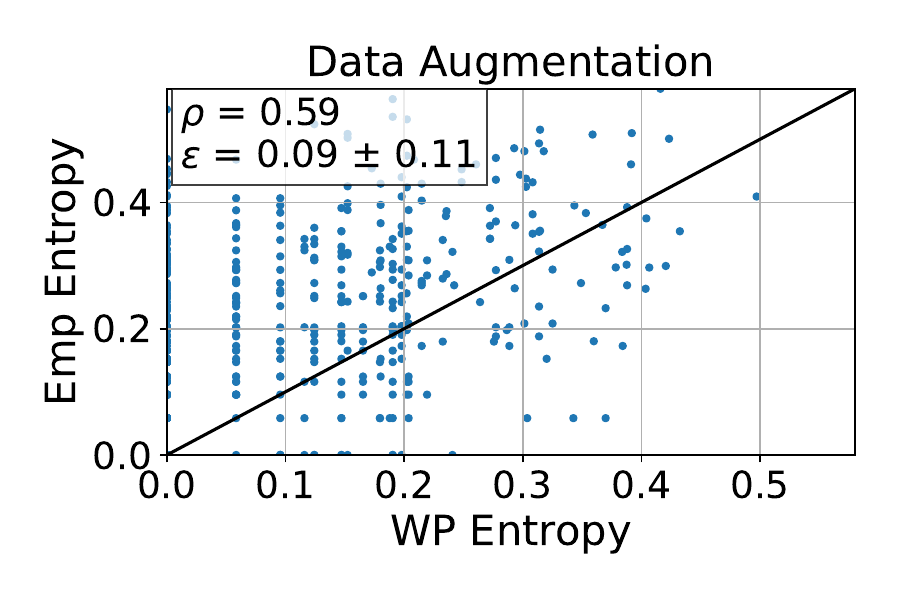}
    \includegraphics[width=0.24\textwidth]{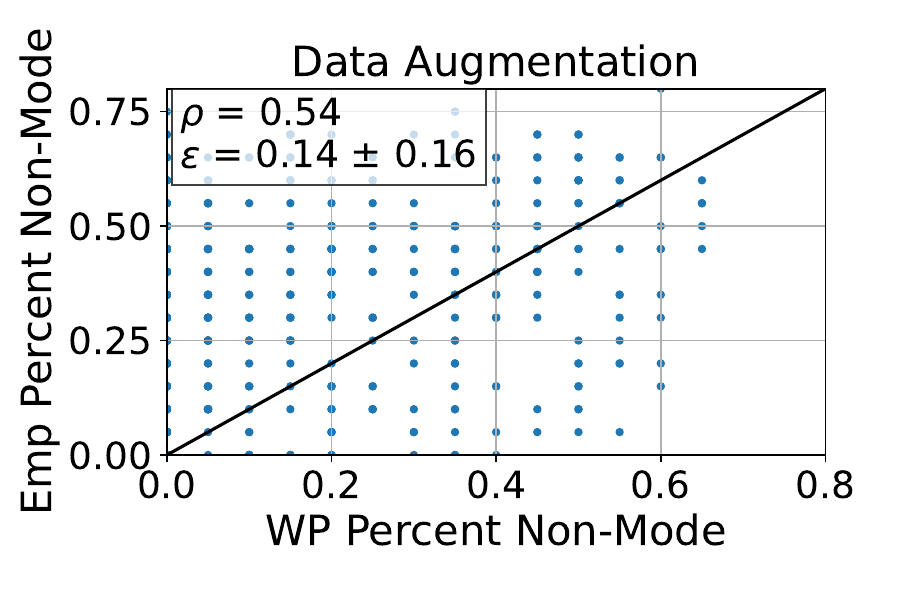}
    \includegraphics[width=0.24\textwidth]{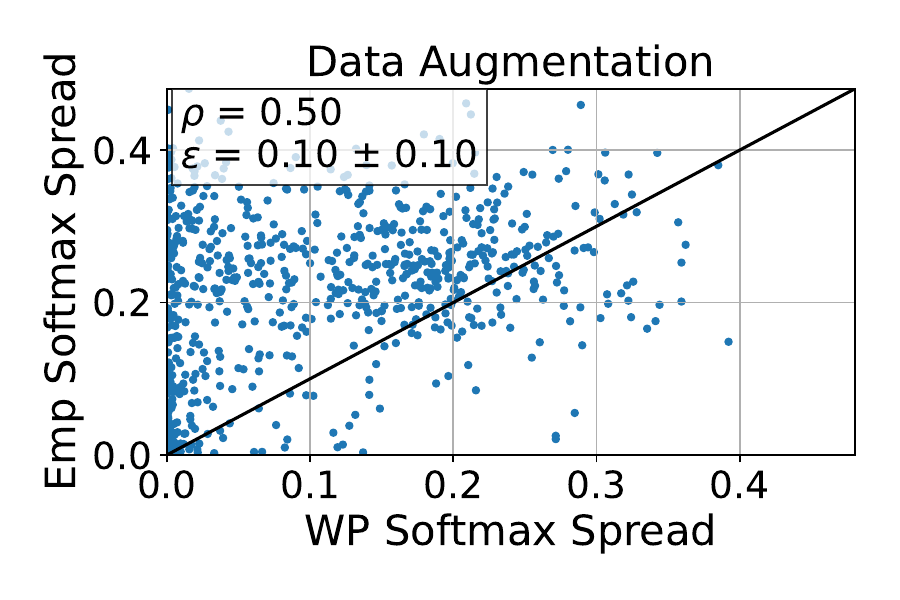}
    \includegraphics[width=0.24\textwidth]{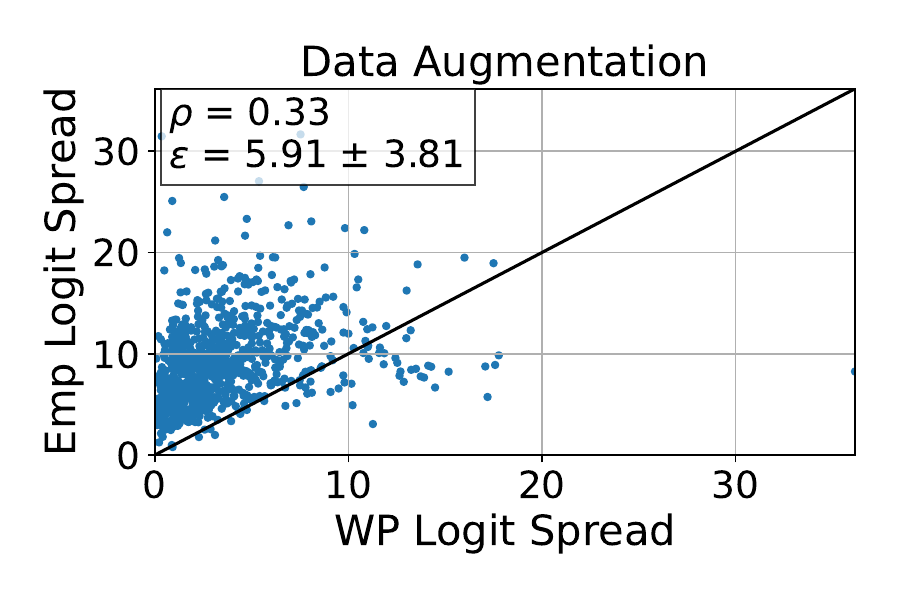}
    \includegraphics[width=0.24\textwidth]{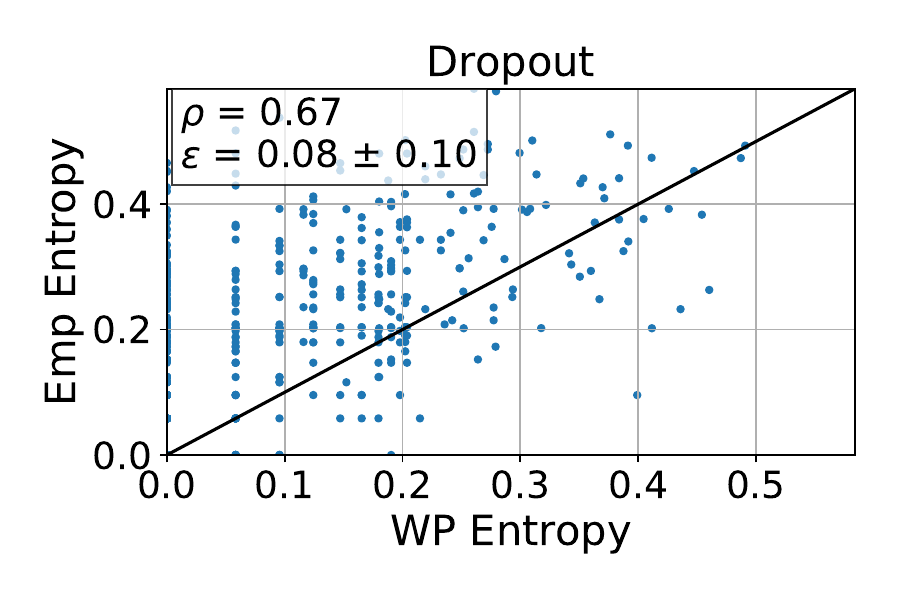}
    \includegraphics[width=0.24\textwidth]{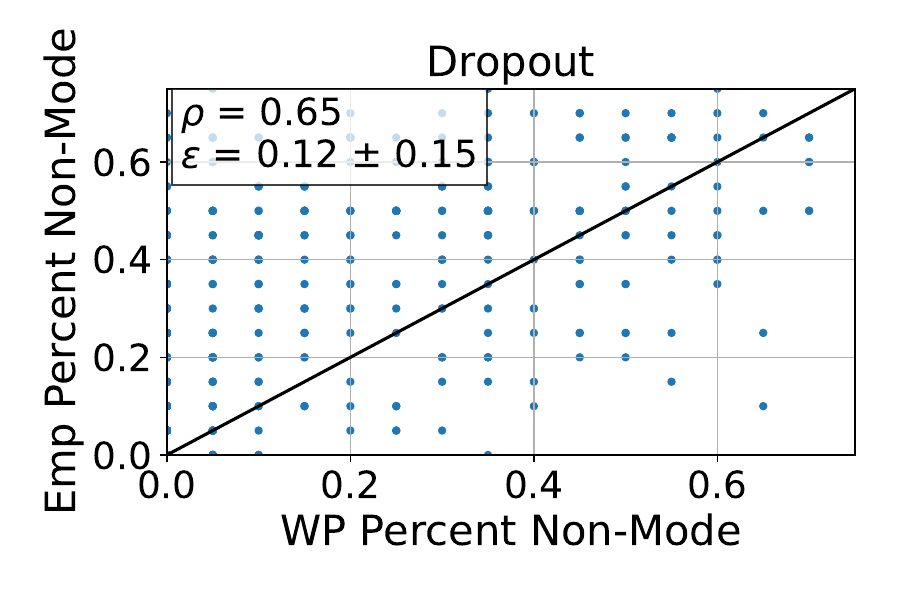}
    \includegraphics[width=0.24\textwidth]{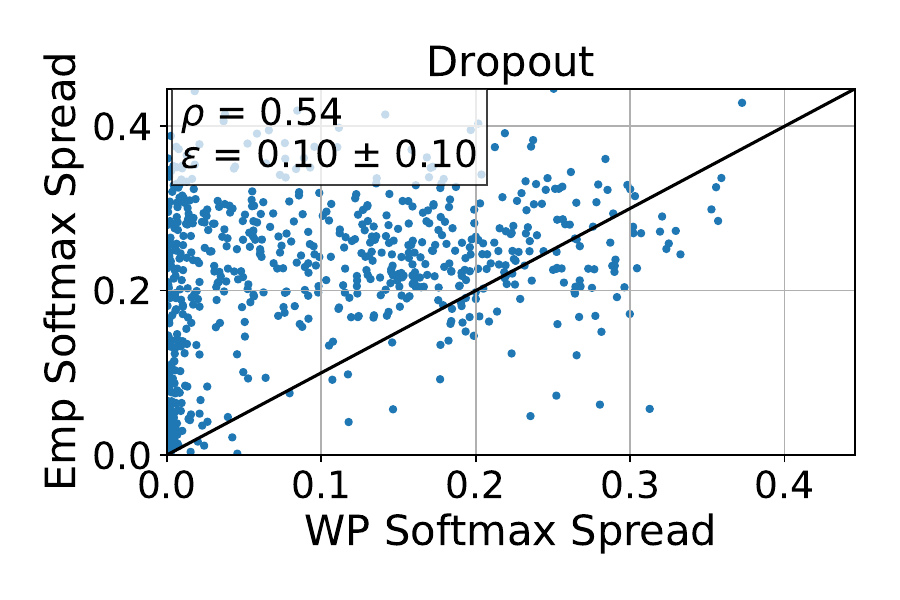}
    \includegraphics[width=0.24\textwidth]{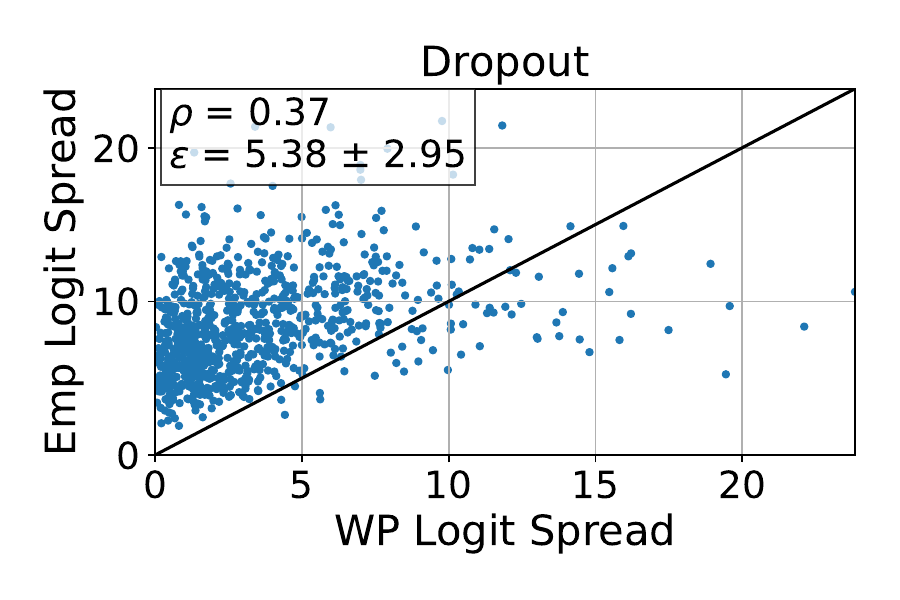}
    \includegraphics[width=0.24\textwidth]{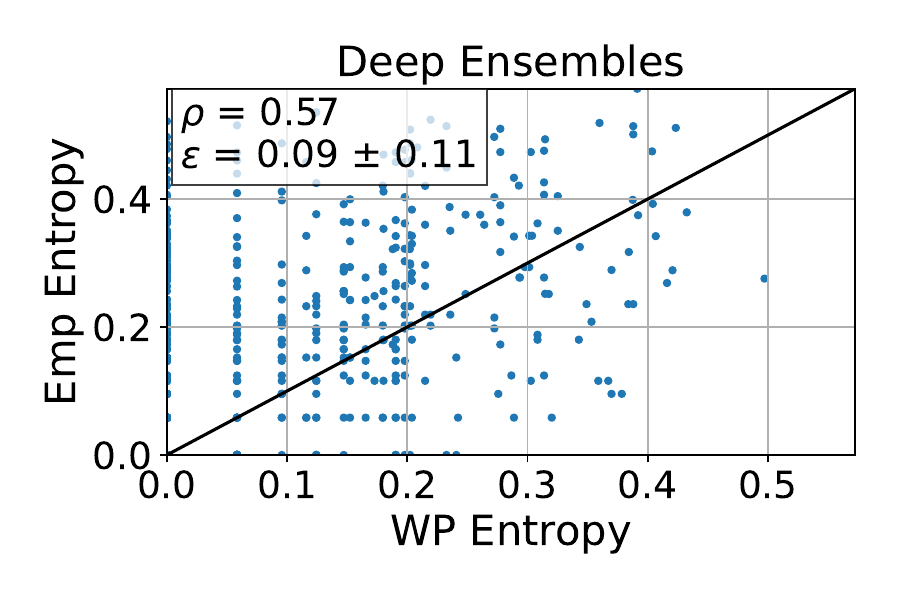}
    \includegraphics[width=0.24\textwidth]{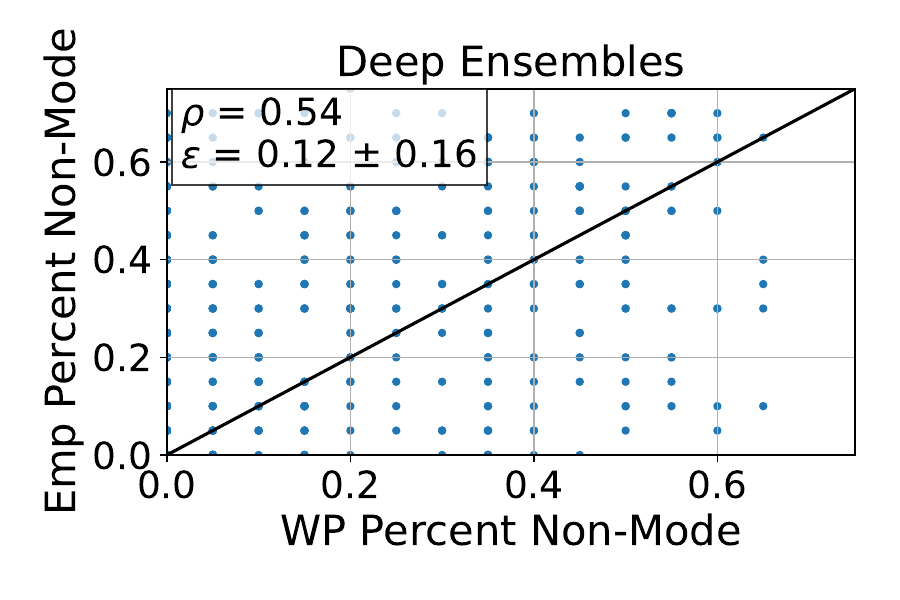}
    \includegraphics[width=0.24\textwidth]{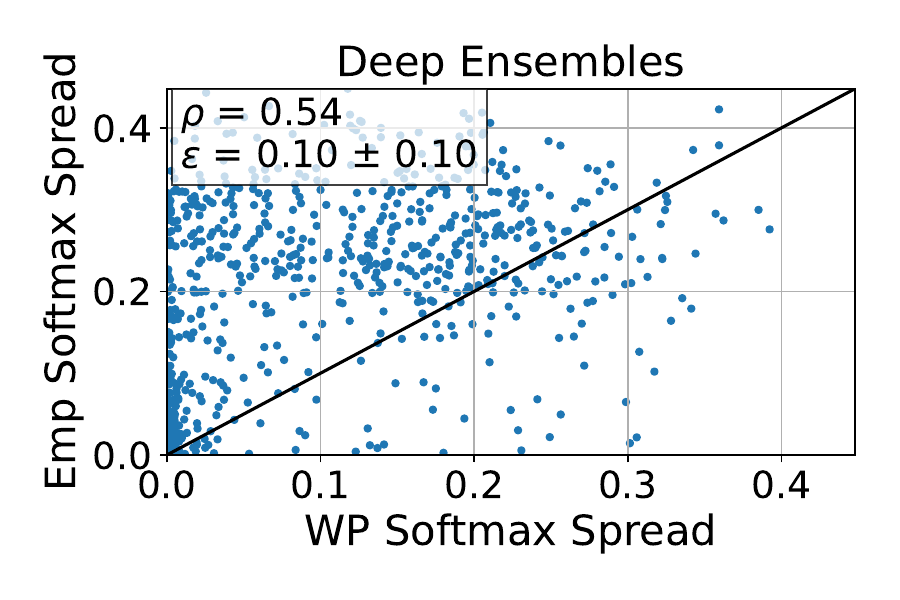}
    \includegraphics[width=0.24\textwidth]{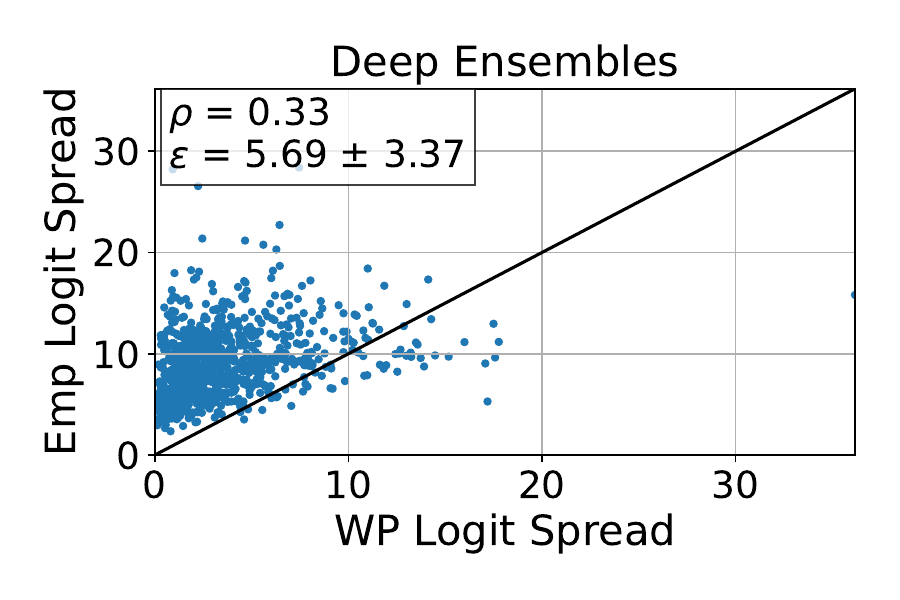}
    \includegraphics[width=0.24\textwidth]{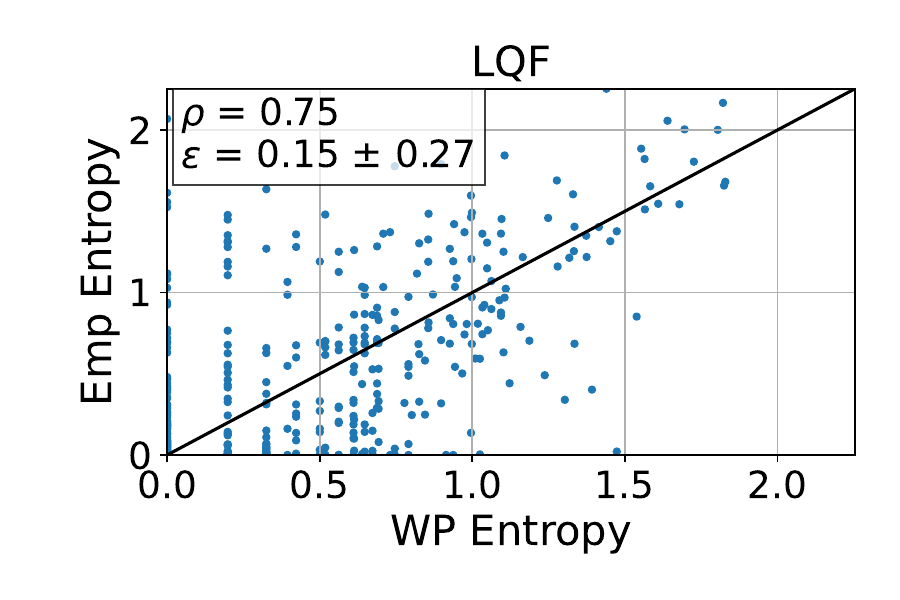}
    \includegraphics[width=0.24\textwidth]{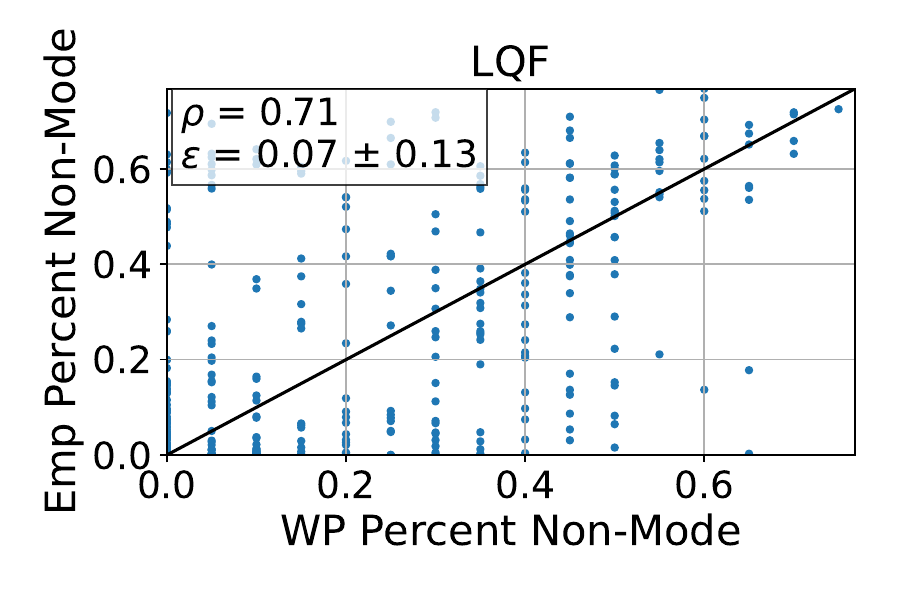}
    \includegraphics[width=0.24\textwidth]{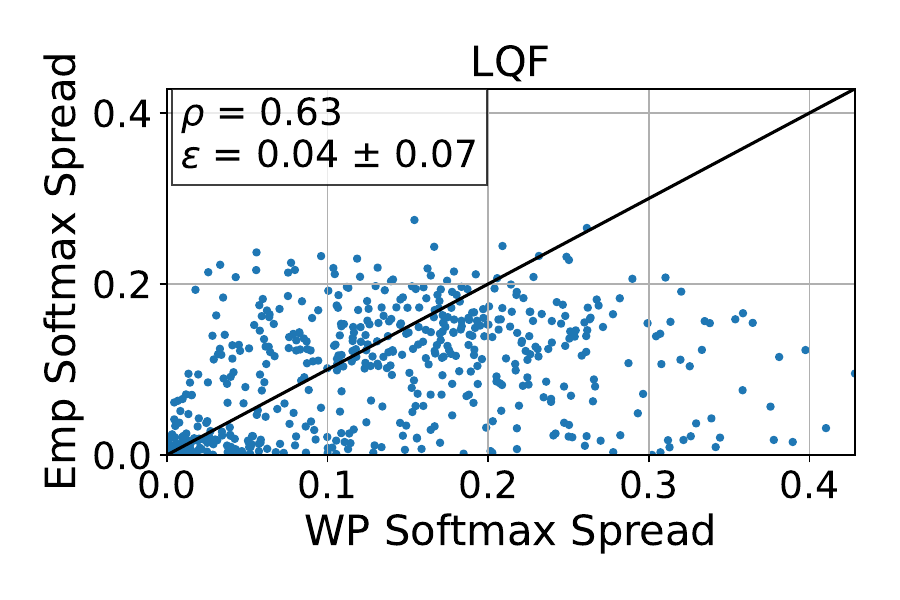}
    \includegraphics[width=0.24\textwidth]{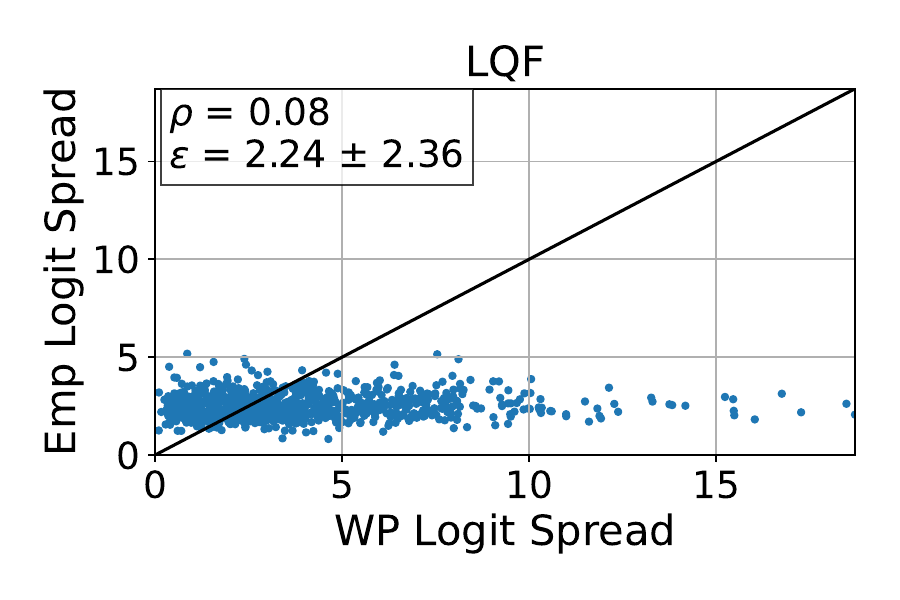}
    \caption{
    \textbf{Empirical Scene Uncertainty vs. Wellington Posterior Scene Uncertainty for ImageNetVid-Robust.} For each scene in the ImageNetVid-Robust dataset, we computed measures of uncertainty (see \cref{sec:measuring_uncertainty}) using the empirical paragon and the Wellington Posterior generated from a single anchor image. The measures of uncertainty are scene entropy (left column), percent non-mode predictions (second column), softmax spread (third column), and logit spread (right column). Methods used to generate the Wellington Posterior are data augmentation (top row), dropout (second row), deep ensembles (third row), and LQF (bottom row). Each point of a plot represents one of the 889 scenes in the ImageNetVid-Robust dataset. The point's x-axis is the uncertainty measure computed from the Wellington Posterior and the point's y-axis is the uncertainty measure computed from the empirical paragon. The box in the top-left corner of each plot displays the correlation coefficient $\rho$ and the mean and standard deviation of absolute error $\varepsilon$ between uncertainties computed using the empirical paragon and the Wellington Posterior. Ideally, we would like to see the points clustered along the diagonal line $y=x$ (plotted in black), which is not what is displayed. Instead, we see many scatters of points that do not appear to form a trend; the standard deviation in absolute error is often as large as the mean error. We also note that the average value of $\varepsilon$ are comparable with the baseline values reported in \cref{tab:vanilla_spread}, further illustrating that there none of these methods are able to generate accurate Wellington Posteriors.
    }
    \label{fig:WP_performance_ImgNetVid}
\end{figure*}

\begin{figure*}[ht]
    \centering
    \includegraphics[width=0.24\textwidth]{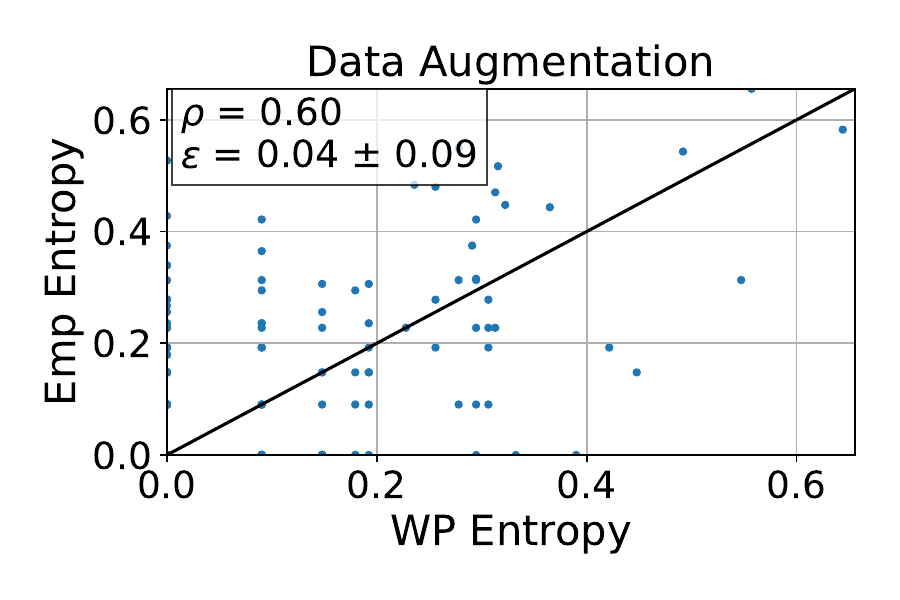}
    \includegraphics[width=0.24\textwidth]{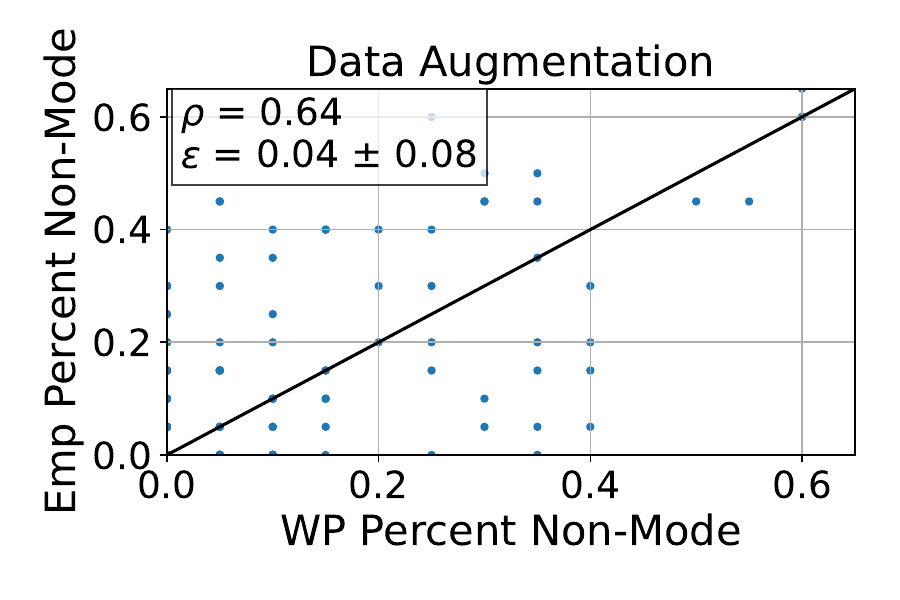}
    \includegraphics[width=0.24\textwidth]{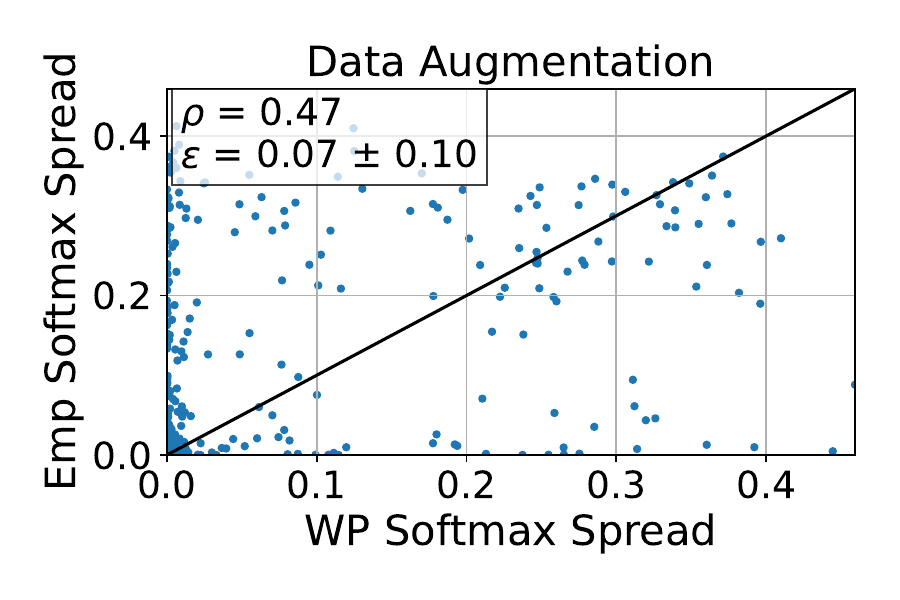}
    \includegraphics[width=0.24\textwidth]{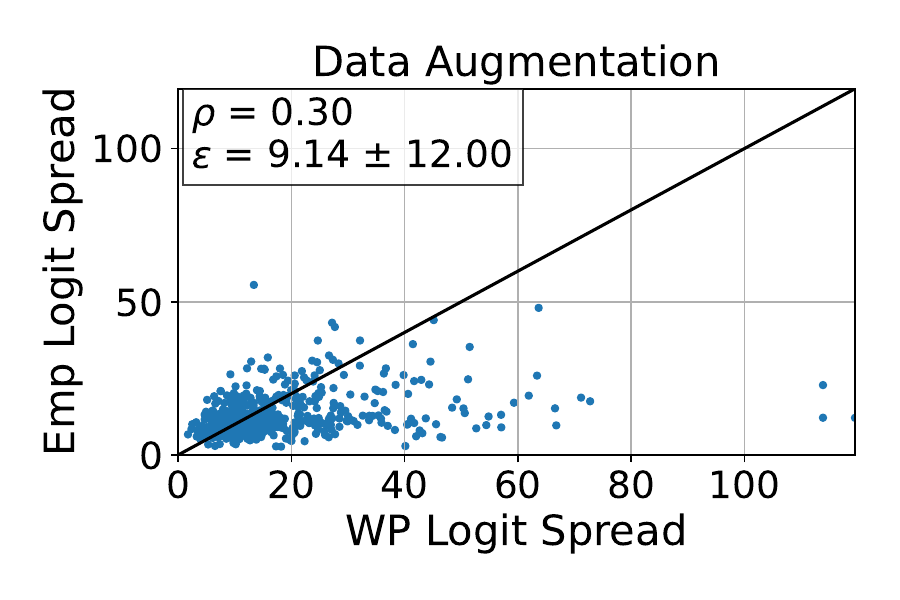}
    
    \includegraphics[width=0.24\textwidth]{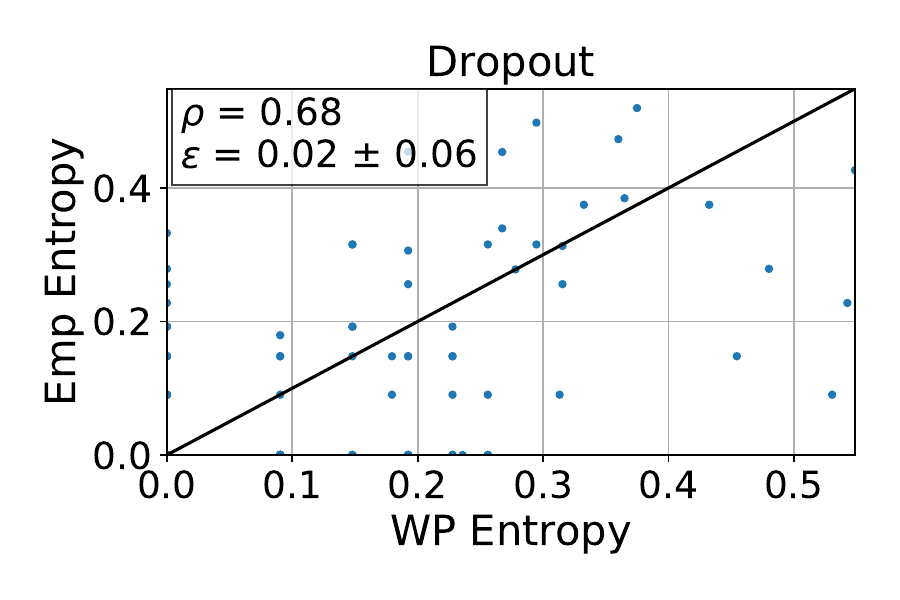}
    \includegraphics[width=0.24\textwidth]{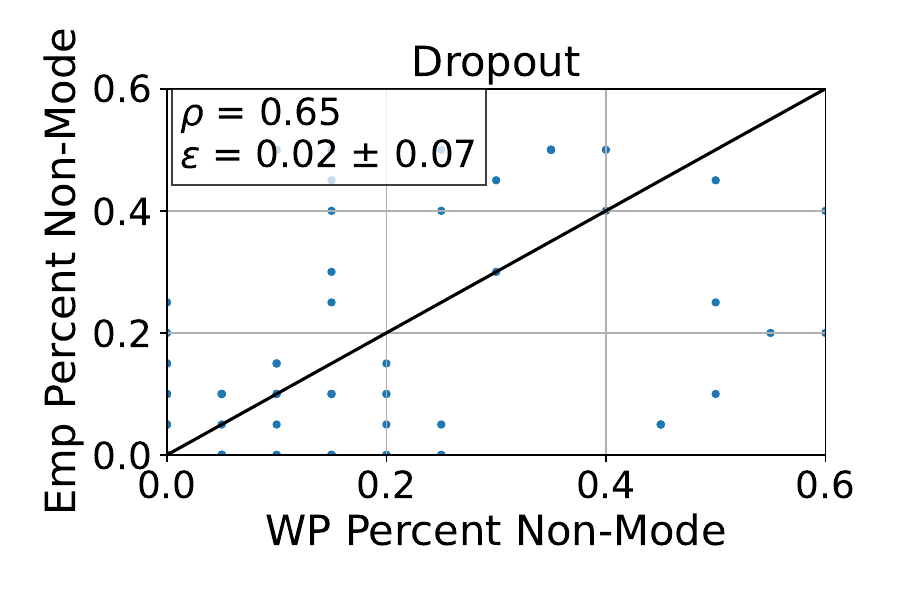}
    \includegraphics[width=0.24\textwidth]{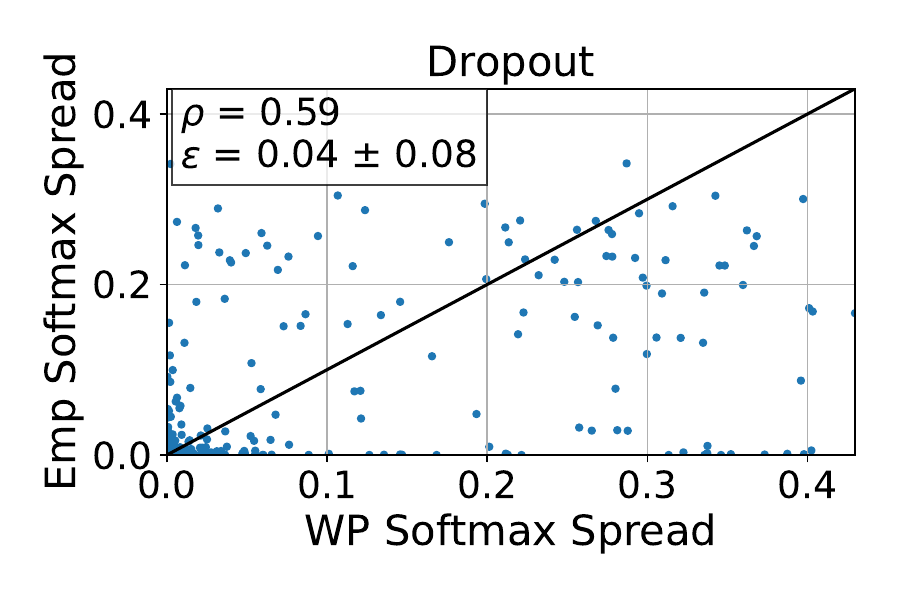}
    \includegraphics[width=0.24\textwidth]{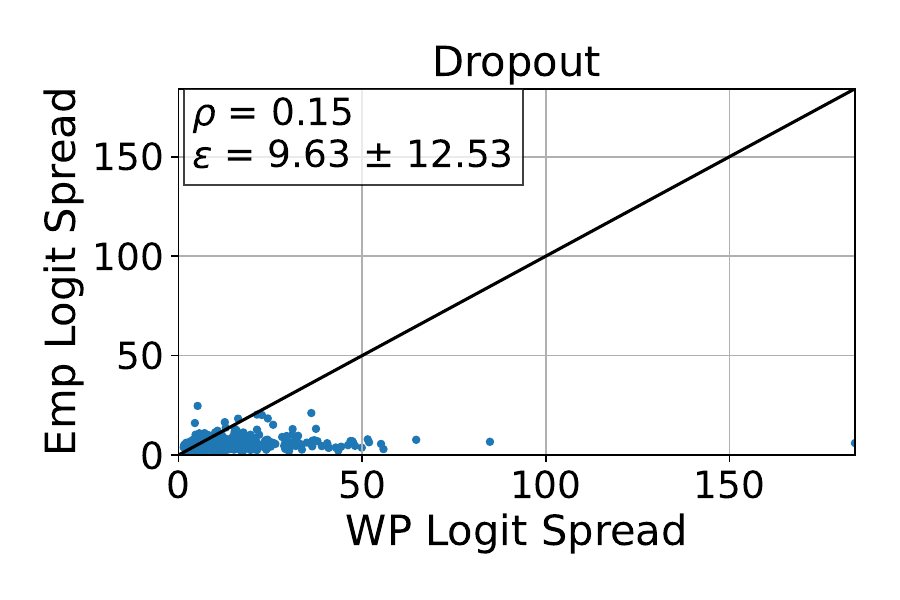}
    
    \includegraphics[width=0.24\textwidth]{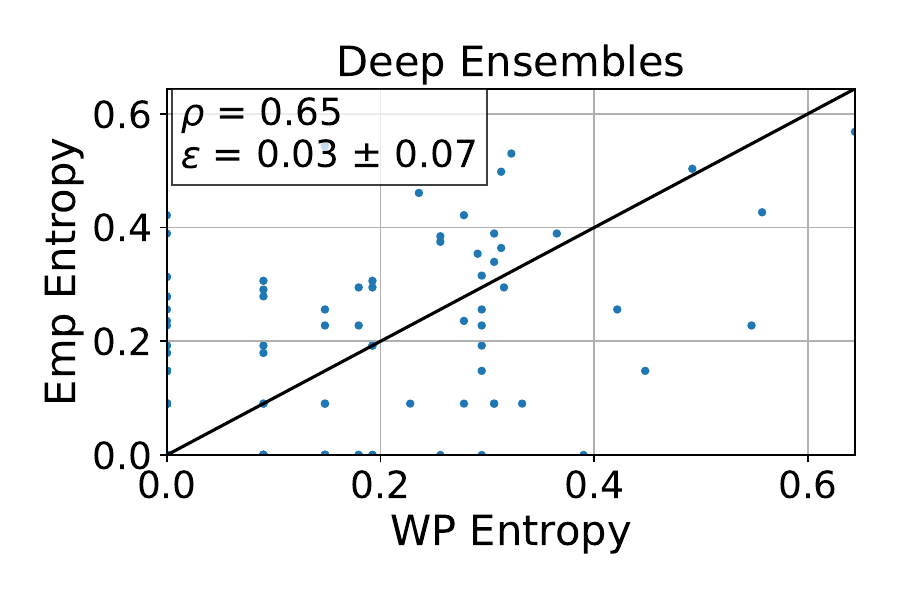}
    \includegraphics[width=0.24\textwidth]{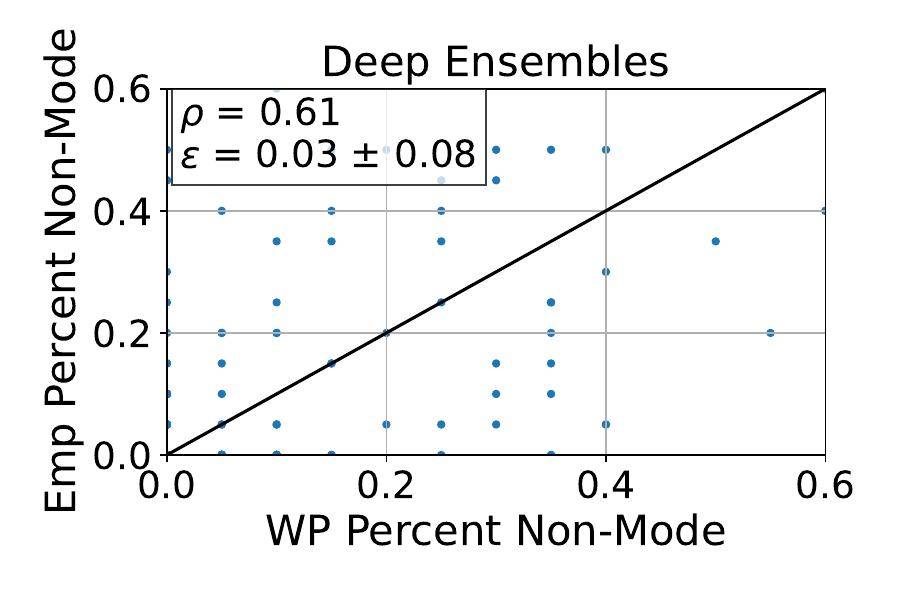}
    \includegraphics[width=0.24\textwidth]{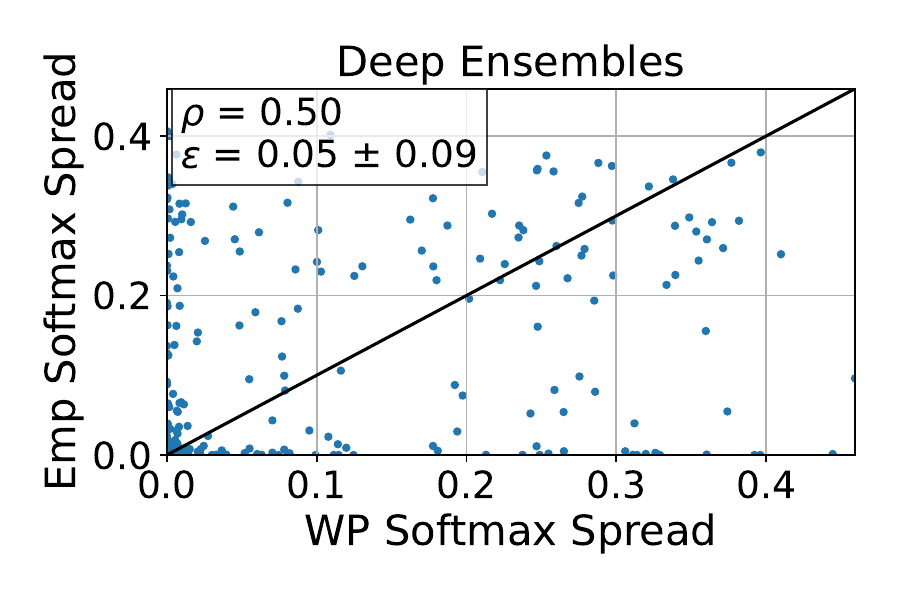}
    \includegraphics[width=0.24\textwidth]{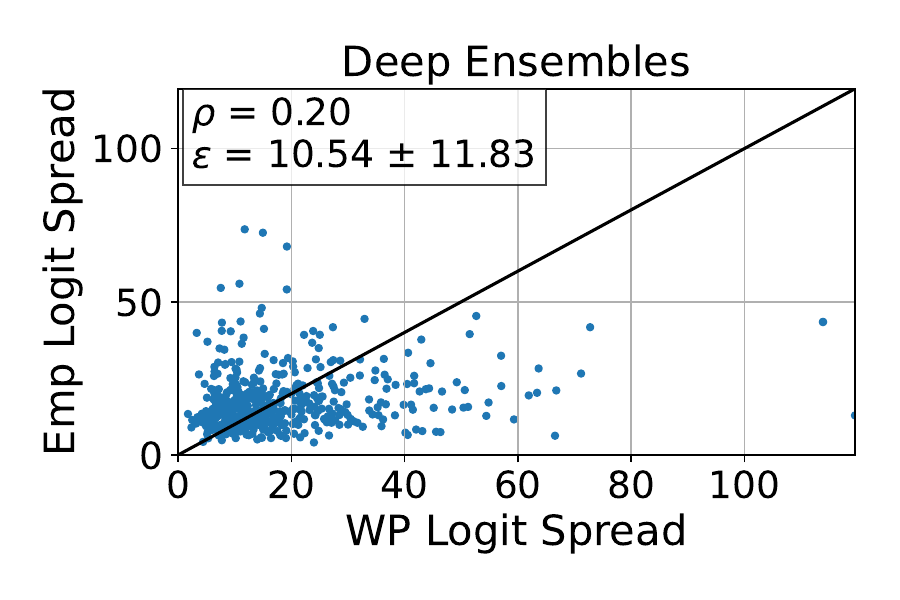}
    
    \includegraphics[width=0.24\textwidth]{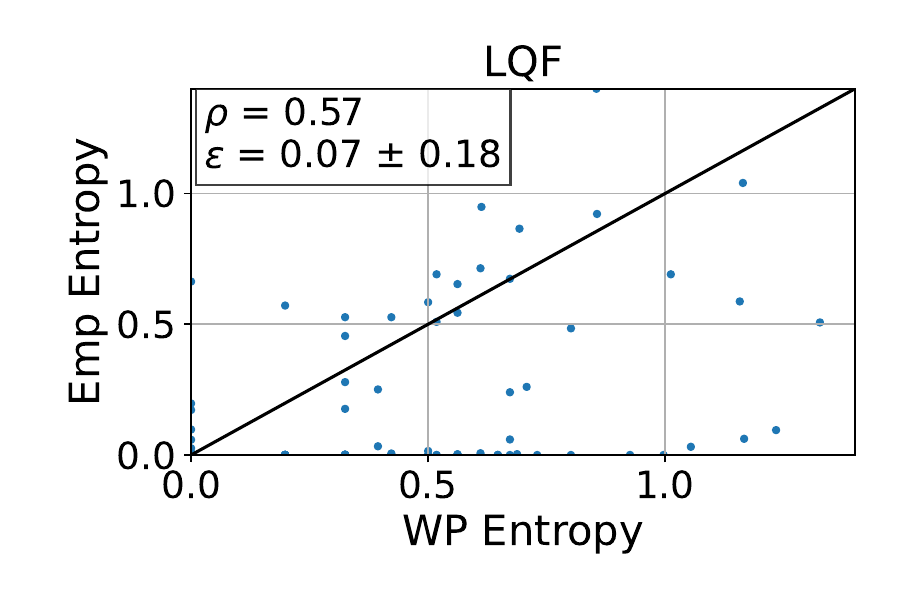}
    \includegraphics[width=0.24\textwidth]{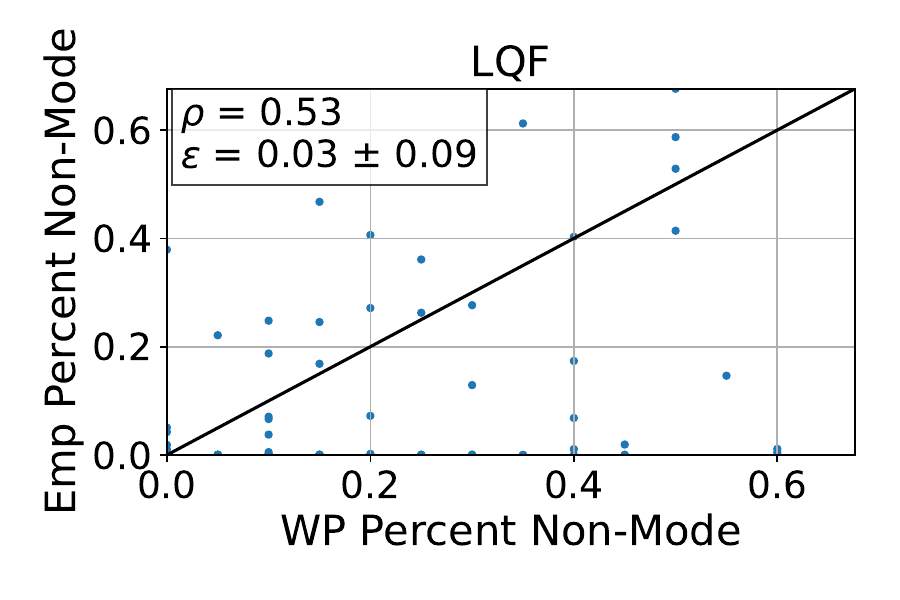}
    \includegraphics[width=0.24\textwidth]{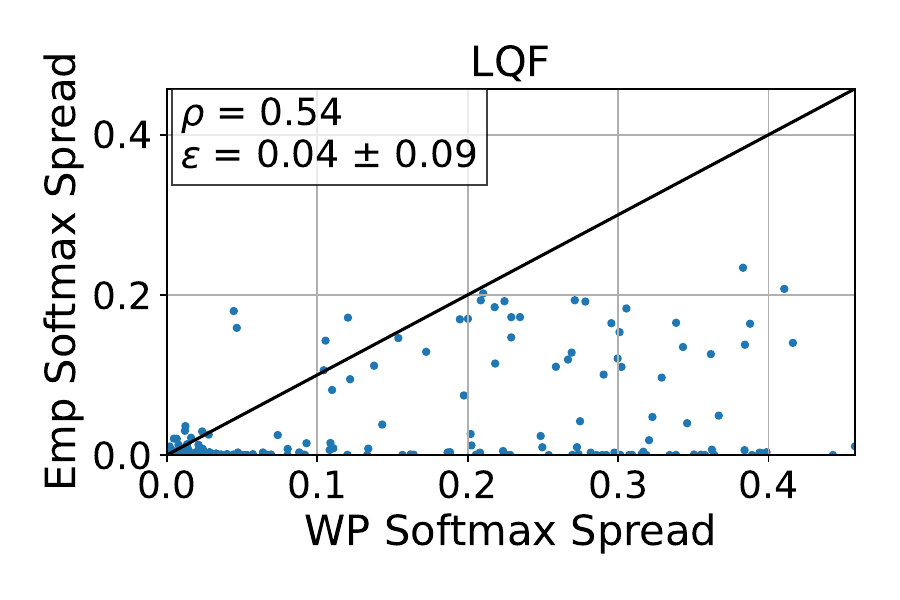}
    \includegraphics[width=0.24\textwidth]{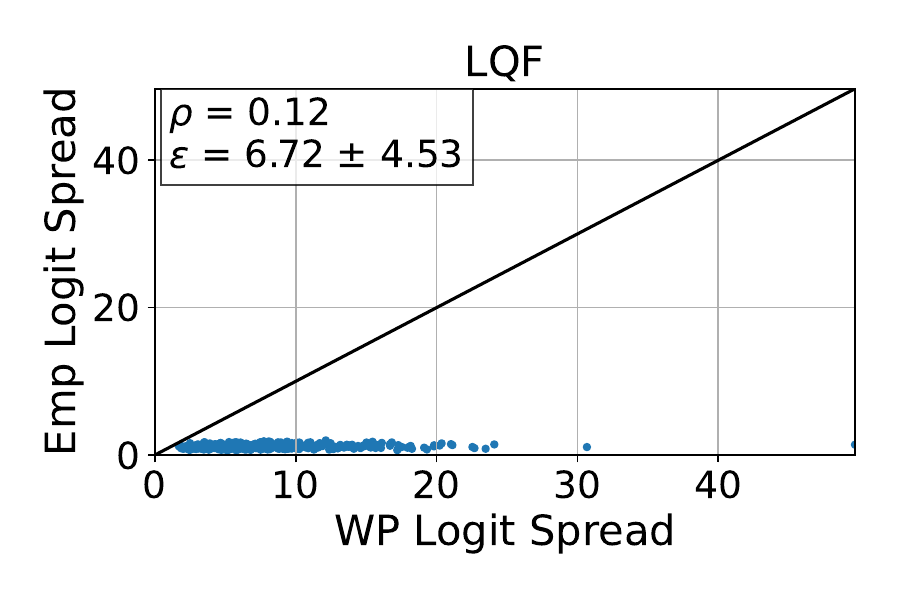}
    
    \includegraphics[width=0.24\textwidth]{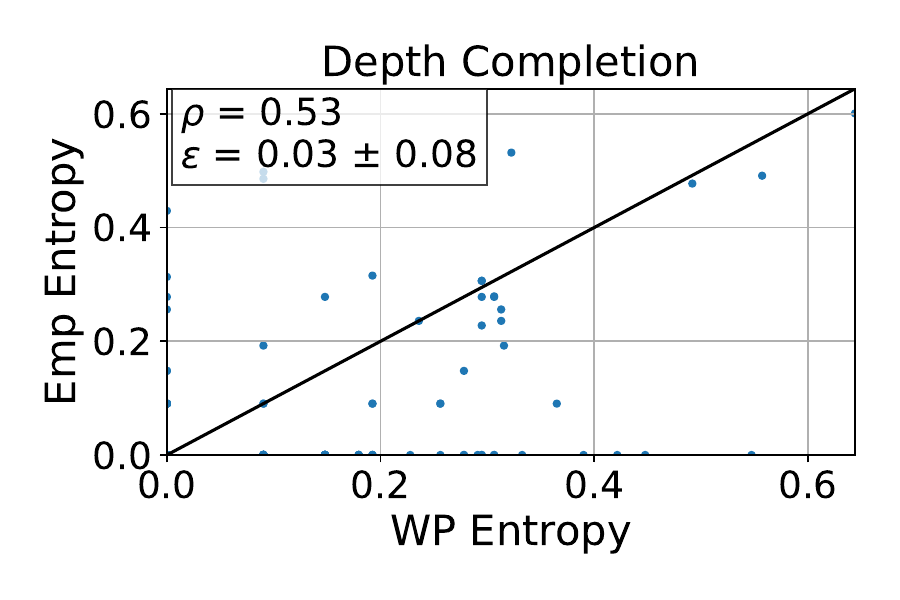}
    \includegraphics[width=0.24\textwidth]{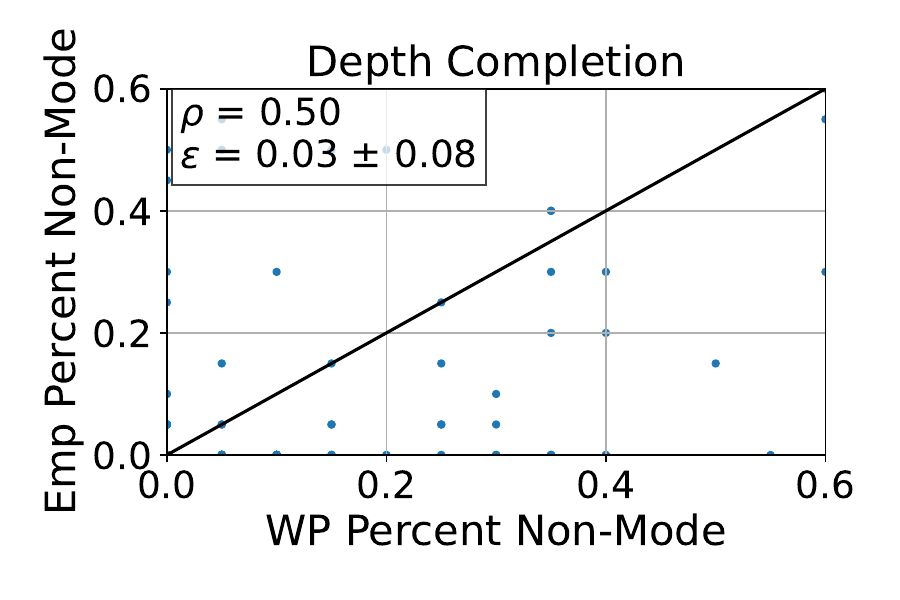}
    \includegraphics[width=0.24\textwidth]{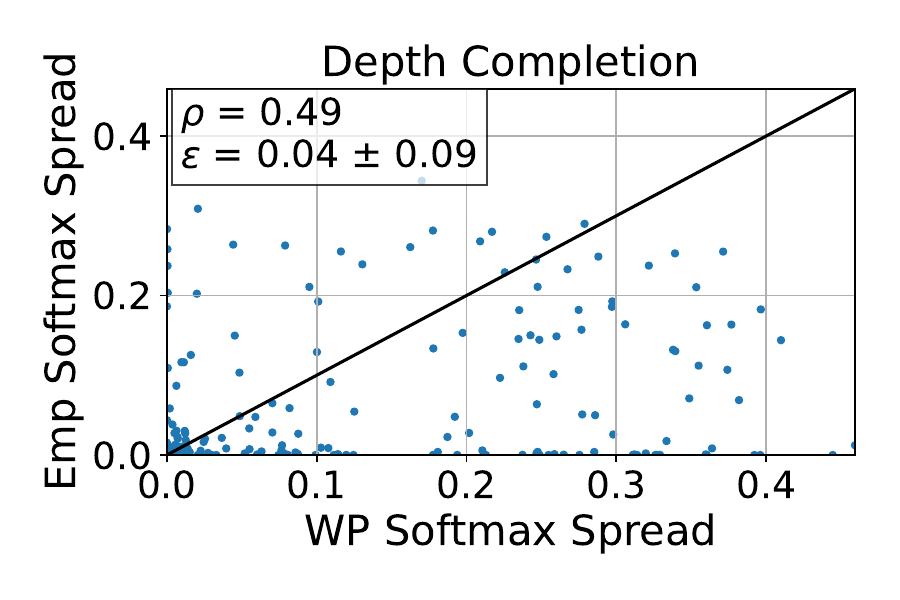}
    \includegraphics[width=0.24\textwidth]{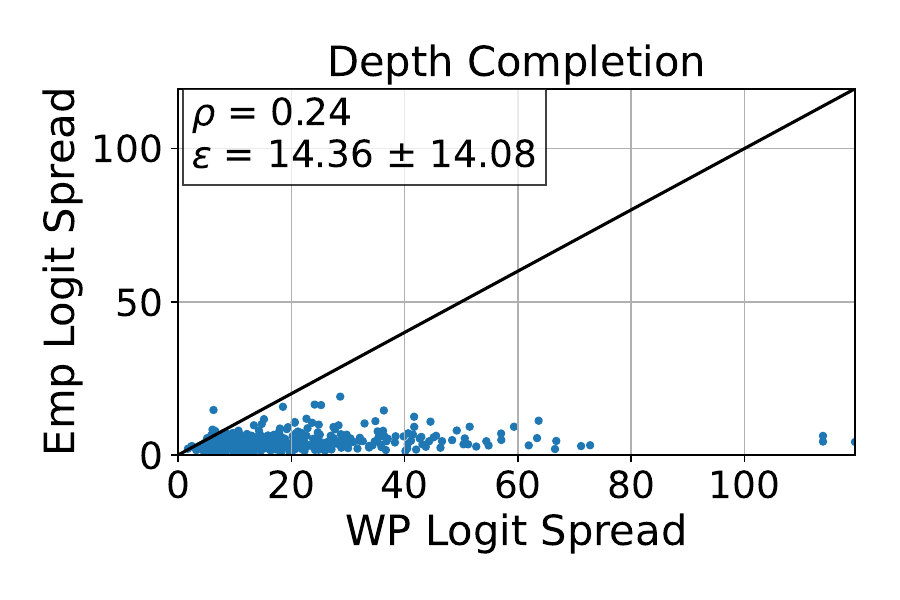}
    \caption{\textbf{Empirical Scene Uncertainty vs. Wellington Posterior Scene Uncertainty for Objectron.} In the above plots, each point is one of 500 scenes from Objectron's test set run through a ResNet-50 network trained on its training set. The point's x-value is an uncertainty measure computed from the Wellington Posterior and the point's y-axis is the uncertainty measure computed from the empirical paragon. The correlation coefficient $\rho$ and the MAE $\varepsilon$ are also displayed. What is notable is that values of $\varepsilon$ are very low (good) while values of $\rho$ are also quite low (bad). This is explained by the fact that most points are located at or very close to the origin --- in these points, all frames in the empirical paragon and all classifications in the WP are the same. Outside of these high-accuracy scenes, there appears to be only weak correlation between uncertainties computed from the WP and uncertainties computed from the empirical paragon.}
    \label{fig:WP_performance_Objectron}
\end{figure*}

\begin{figure*}[ht]
    \centering
    \includegraphics[width=0.24\textwidth]{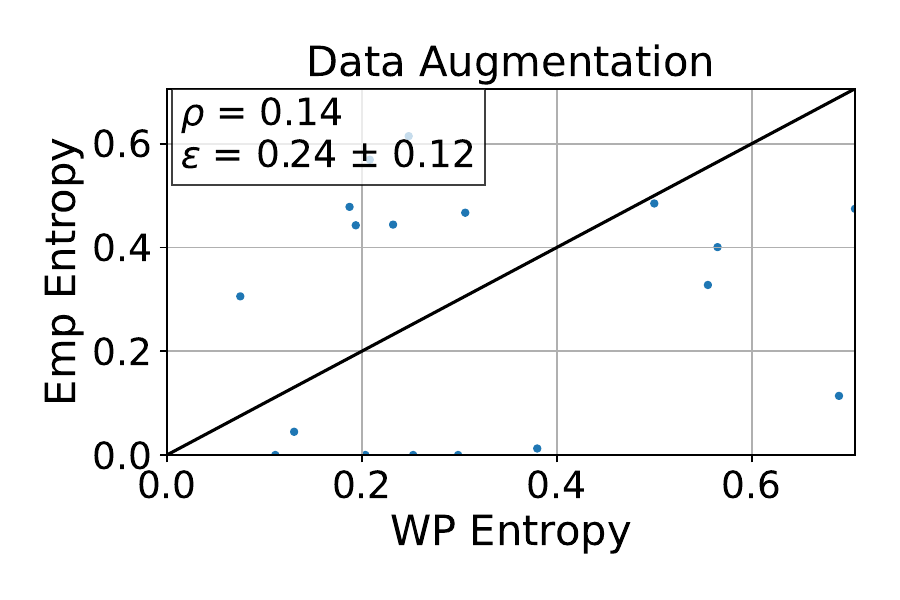}
    \includegraphics[width=0.24\textwidth]{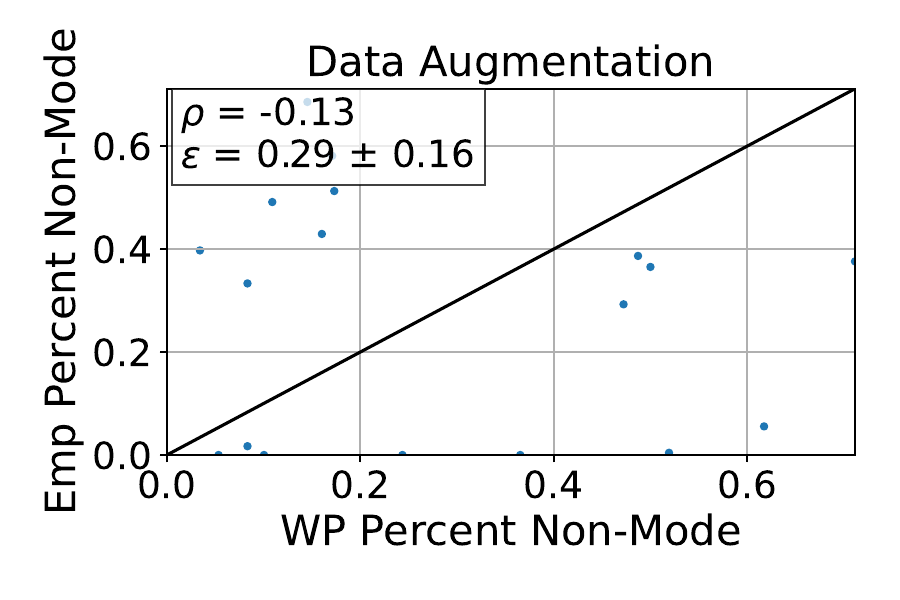}
    \includegraphics[width=0.24\textwidth]{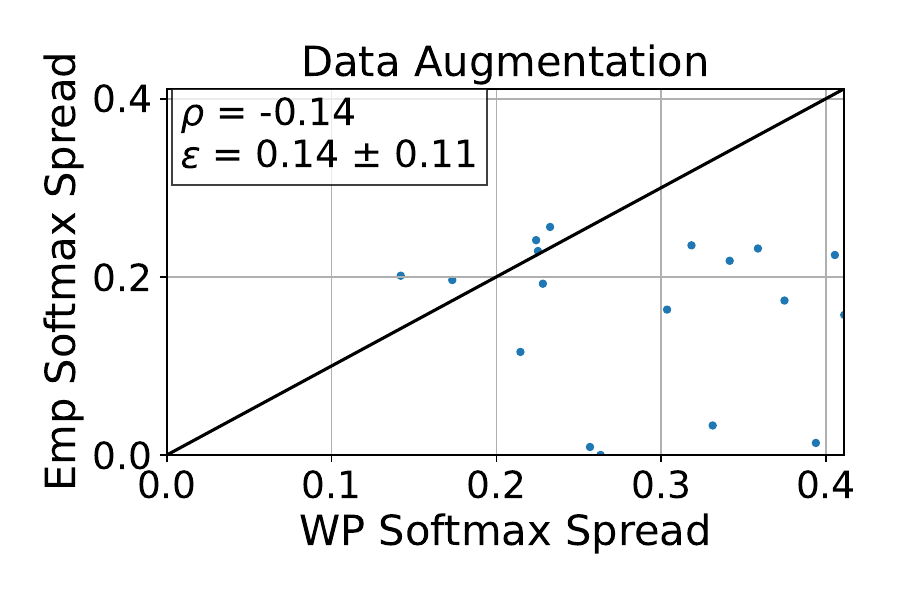}
    \includegraphics[width=0.24\textwidth]{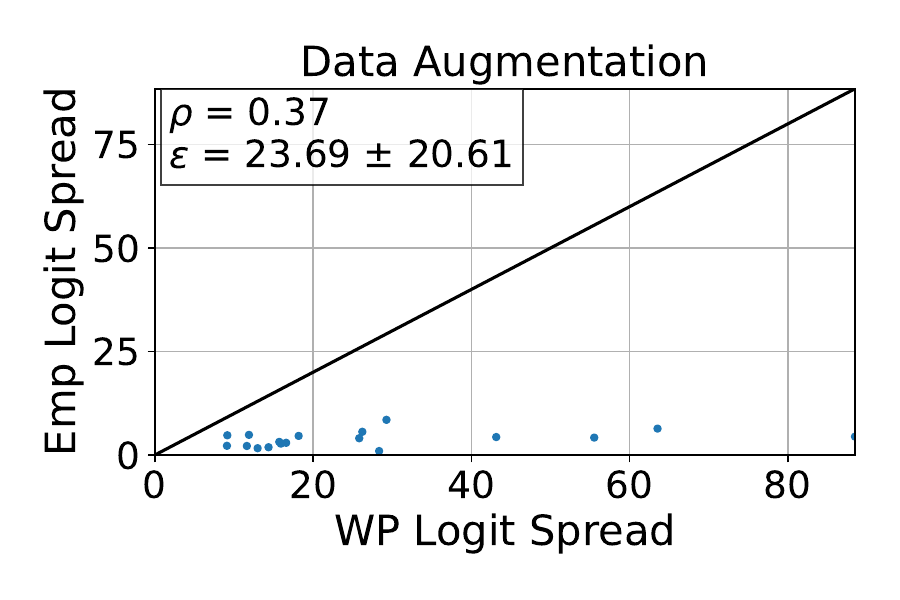}
    
    \includegraphics[width=0.24\textwidth]{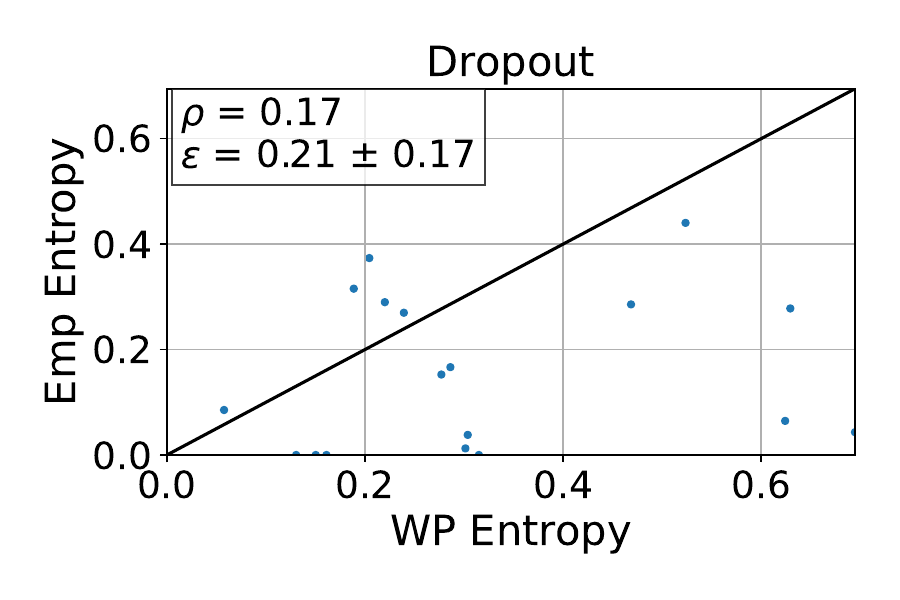}
    \includegraphics[width=0.24\textwidth]{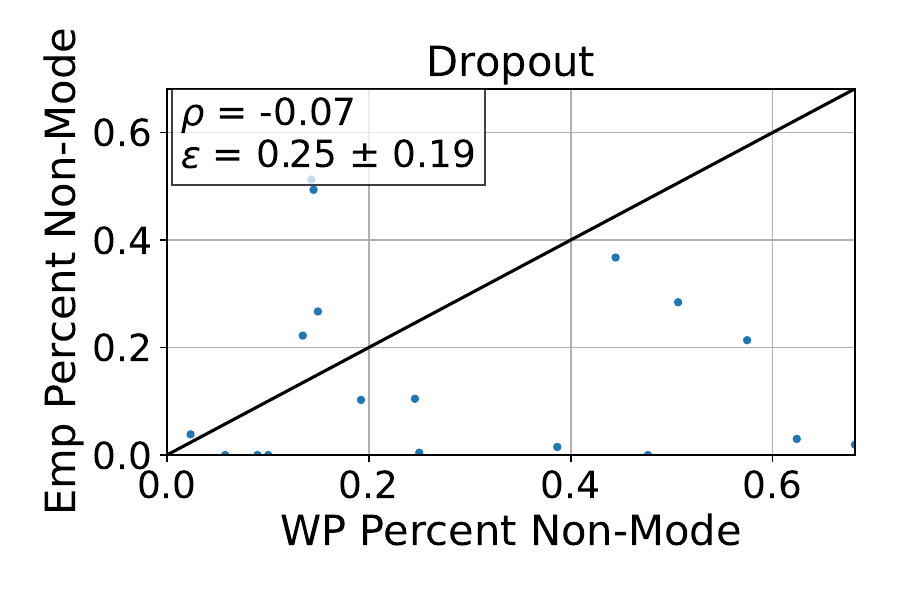}
    \includegraphics[width=0.24\textwidth]{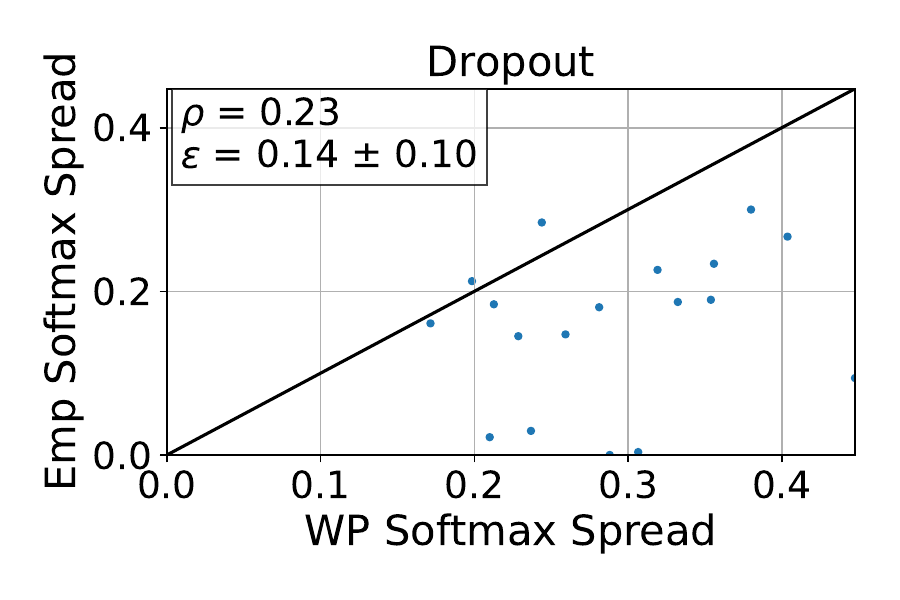}
    \includegraphics[width=0.24\textwidth]{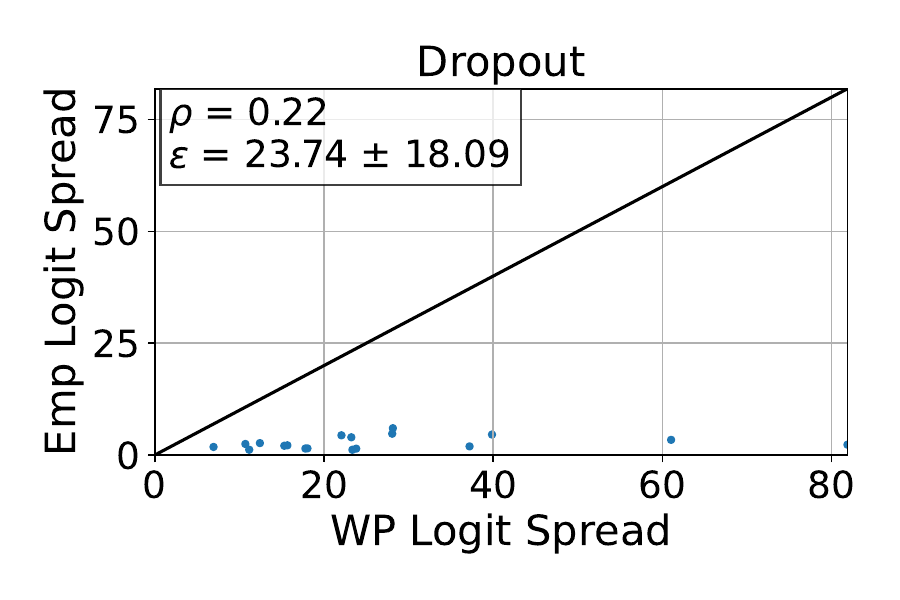}
    
    \includegraphics[width=0.24\textwidth]{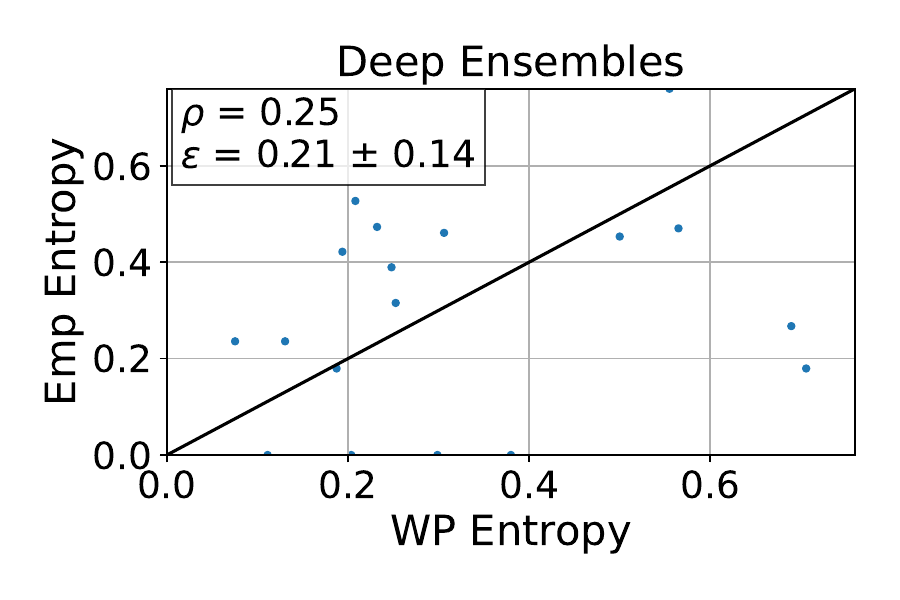}
    \includegraphics[width=0.24\textwidth]{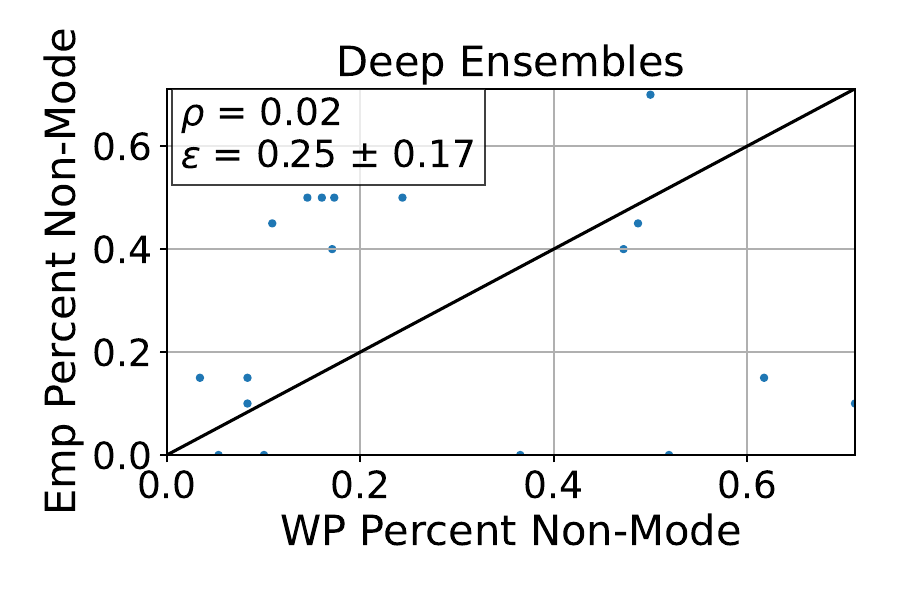}
    \includegraphics[width=0.24\textwidth]{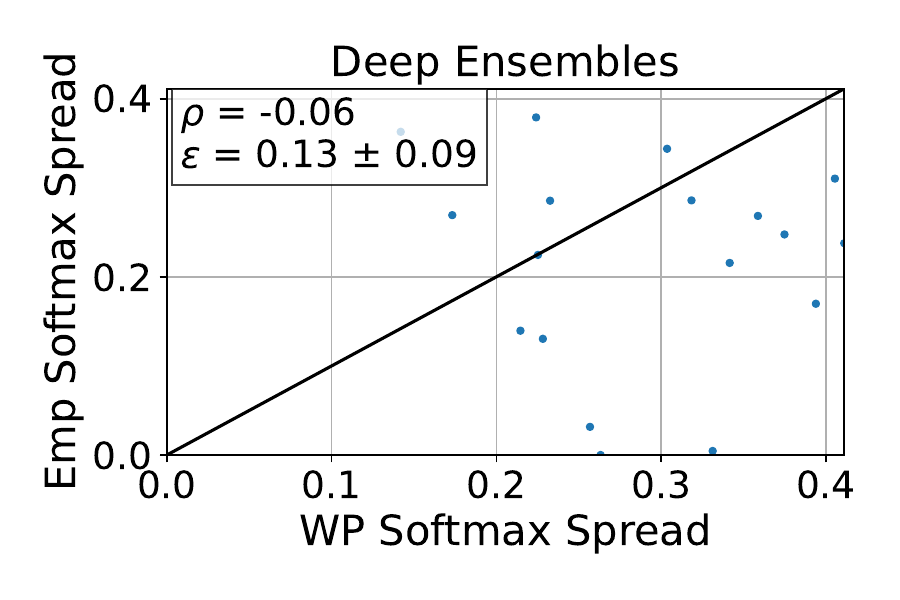}
    \includegraphics[width=0.24\textwidth]{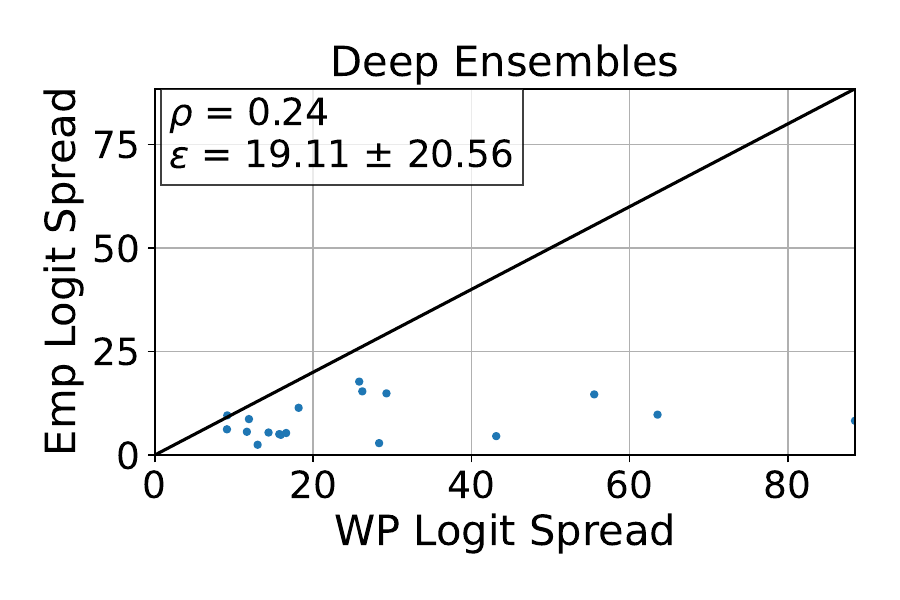}
    
    \includegraphics[width=0.24\textwidth]{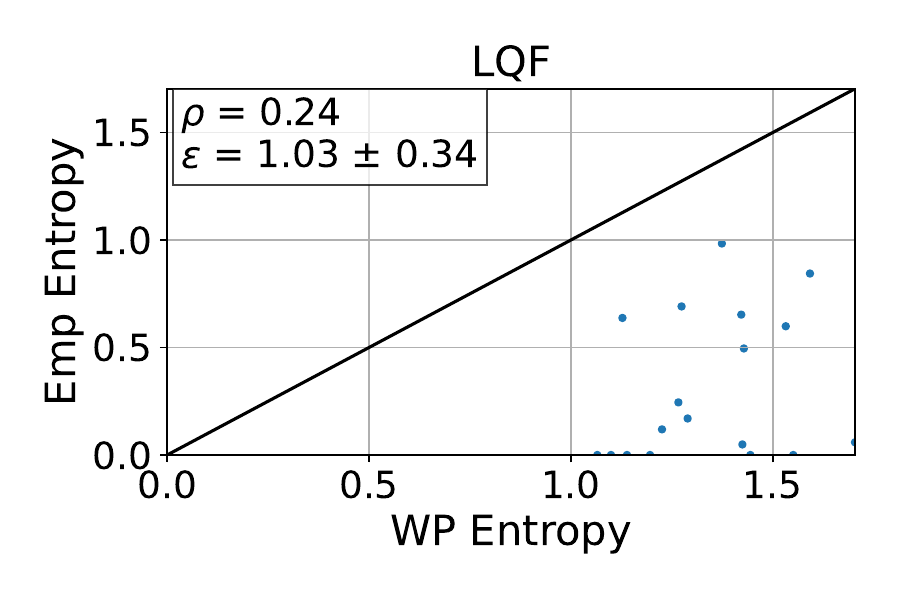}
    \includegraphics[width=0.24\textwidth]{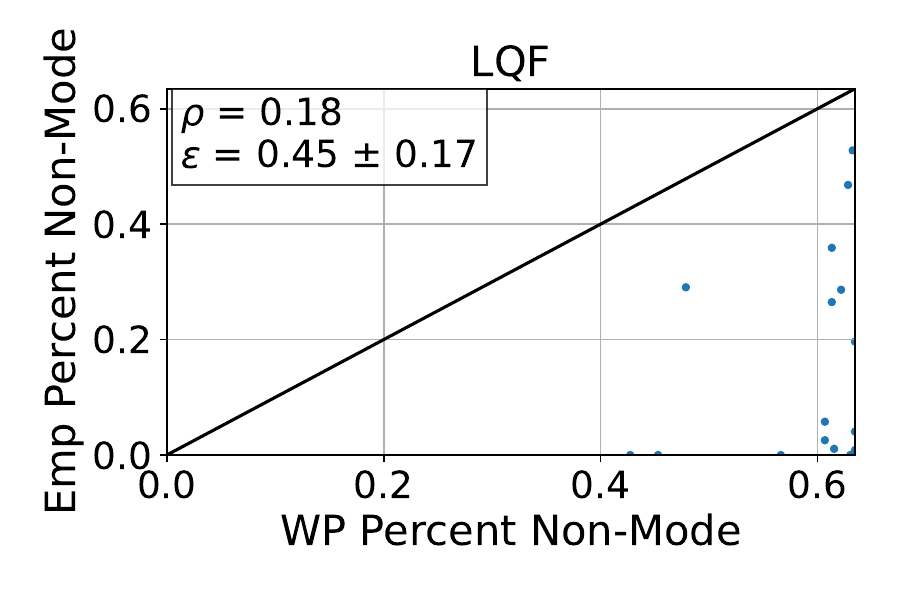}
    \includegraphics[width=0.24\textwidth]{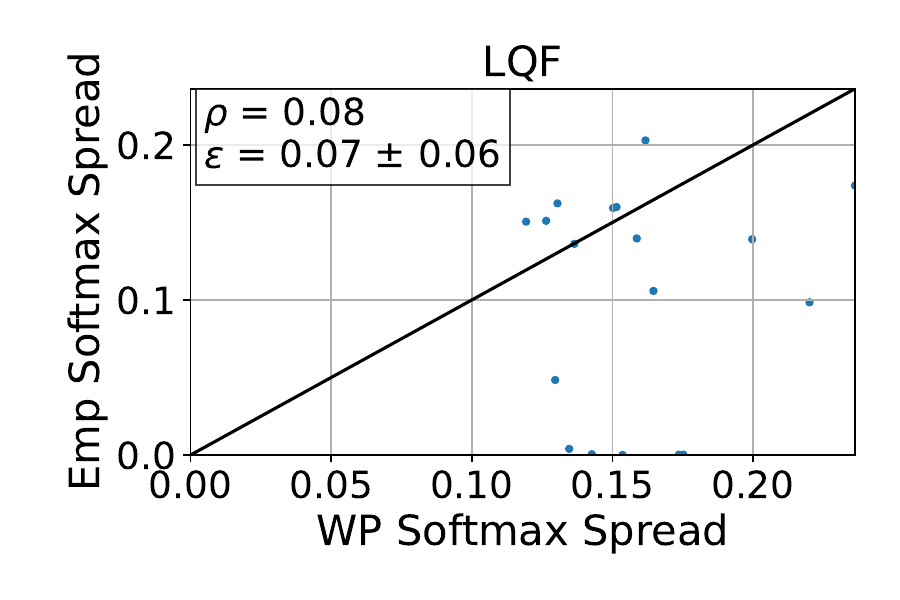}
    \includegraphics[width=0.24\textwidth]{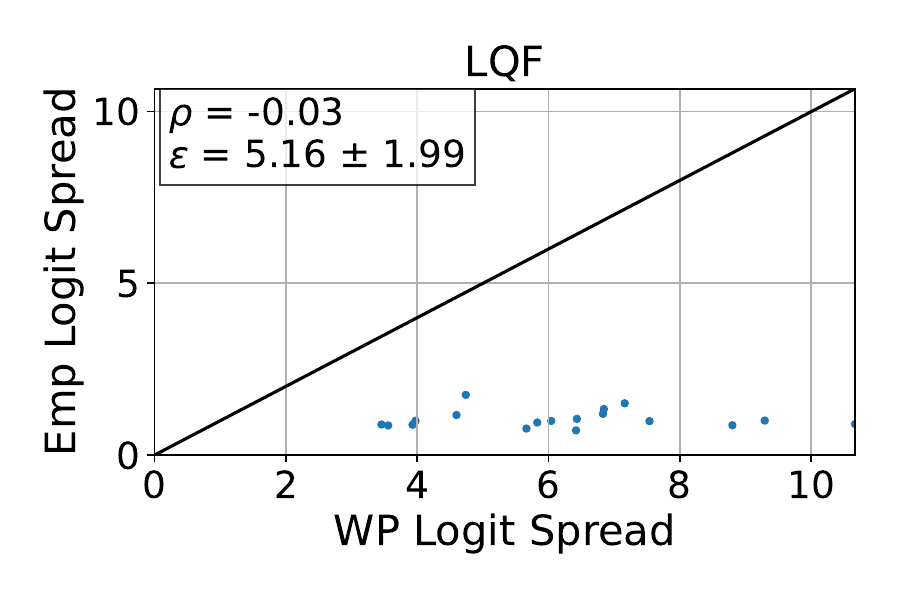}
    
    \caption{\textbf{Empirical Scene Uncertainty vs. Wellington Posterior Scene Uncertainty for Synthetic Objectron.} In the above plots, each point is one of 18 scenes from Synthetic Objectron run through a ResNet-50 network trained on its training set. The point's x-value is an uncertainty measure computed from the Wellington Posterior and the point's y-axis is the uncertainty measure computed from the empirical paragon. The correlation coefficient $\rho$ and the MAE $\varepsilon$ are also displayed. Both MAE and correlation coefficients are poor.}
    \label{fig:WP_performance_SyntheticObjectron}
\end{figure*}

\begin{figure*}[ht]
    \centering
    \includegraphics[width=0.24\textwidth]{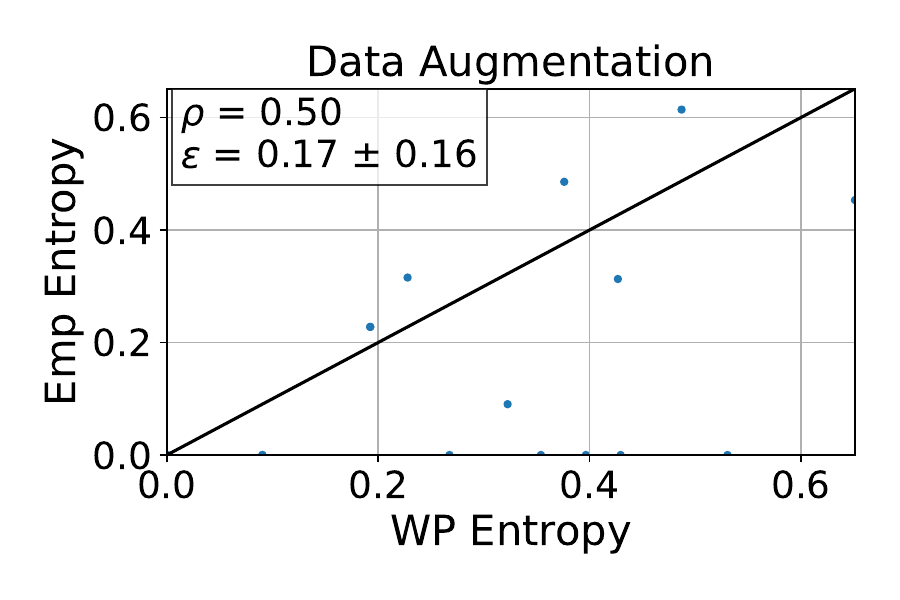}
    \includegraphics[width=0.24\textwidth]{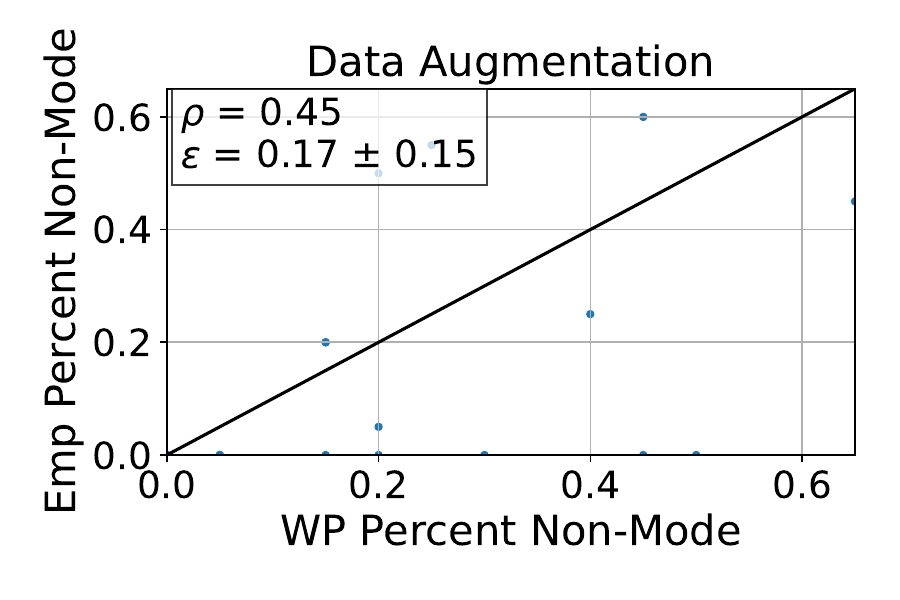}
    \includegraphics[width=0.24\textwidth]{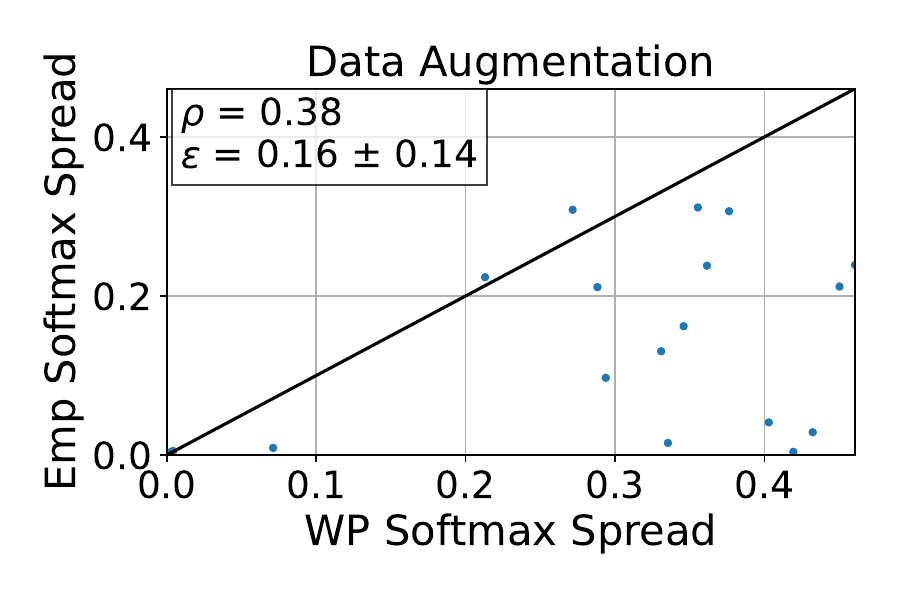}
    \includegraphics[width=0.24\textwidth]{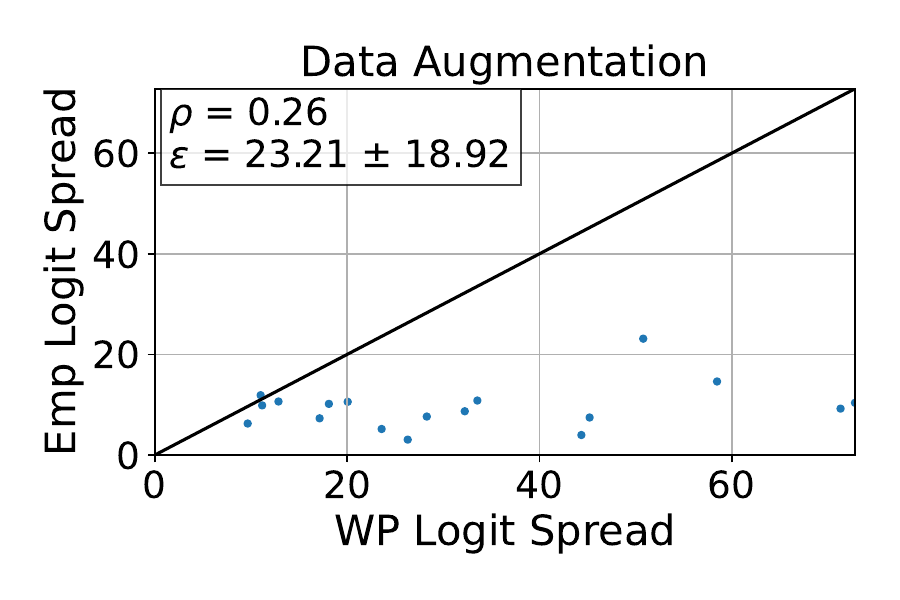}
    
    \includegraphics[width=0.24\textwidth]{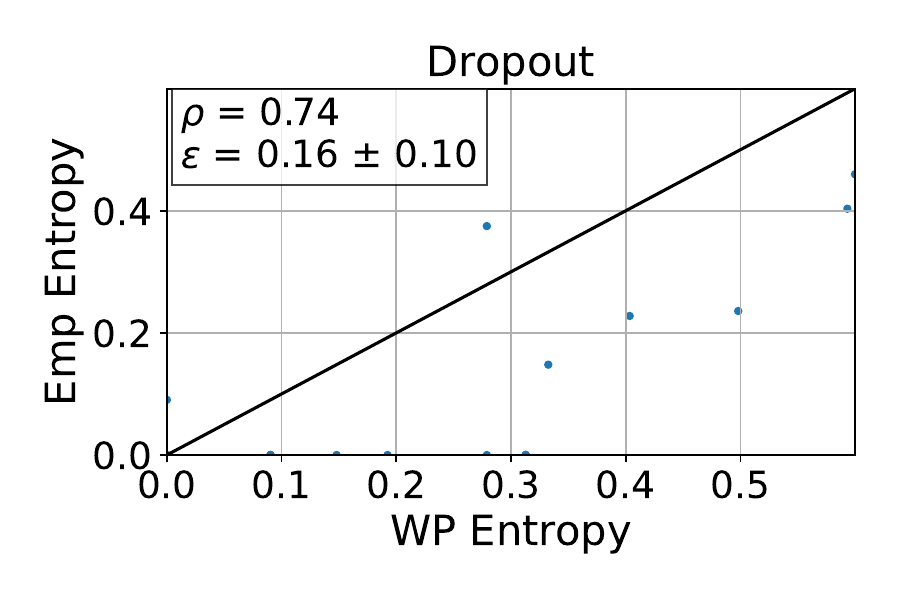}
    \includegraphics[width=0.24\textwidth]{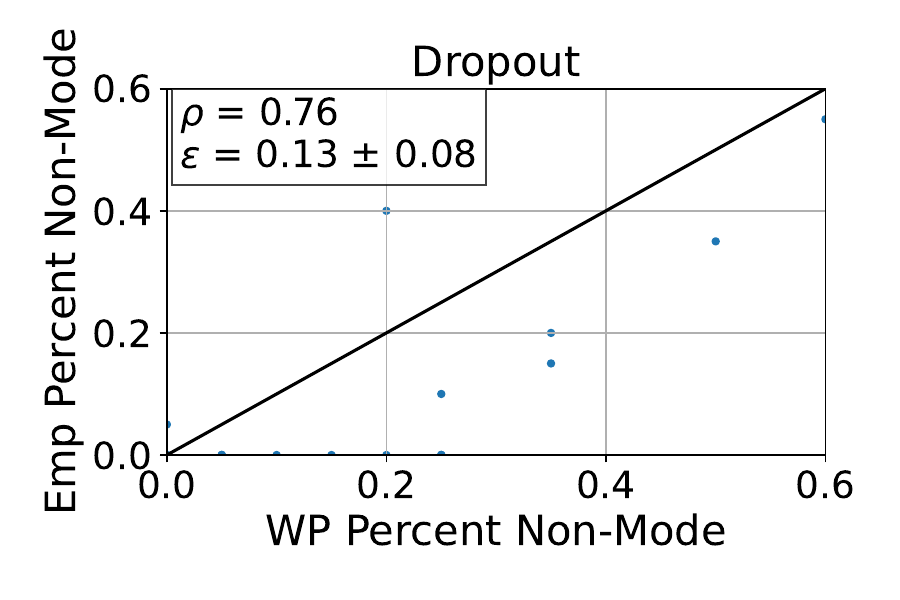}
    \includegraphics[width=0.24\textwidth]{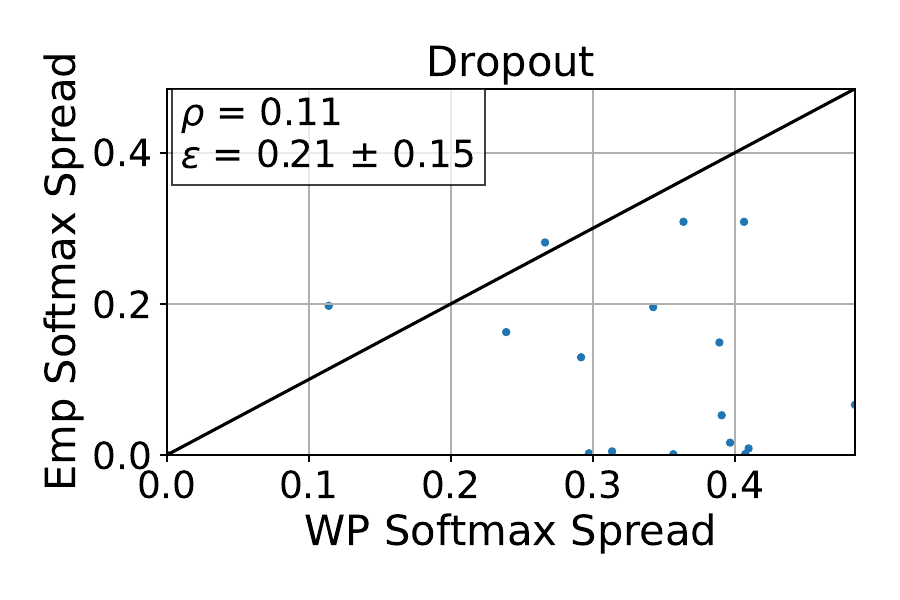}
    \includegraphics[width=0.24\textwidth]{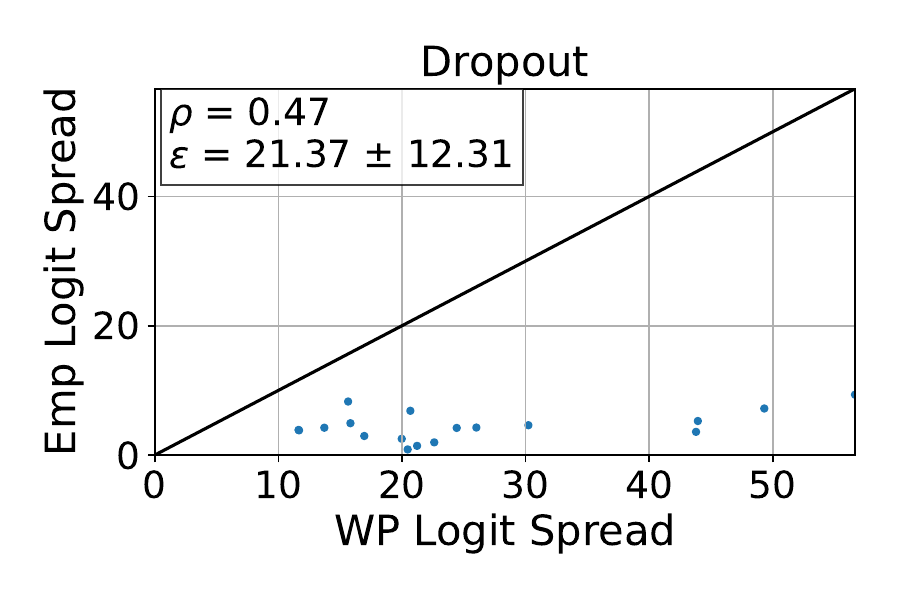}
    
    \includegraphics[width=0.24\textwidth]{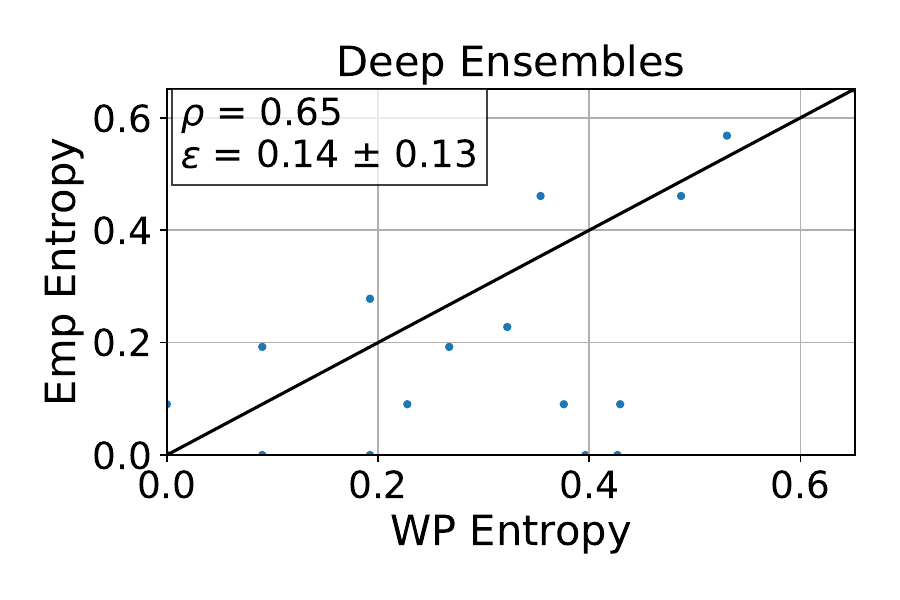}
    \includegraphics[width=0.24\textwidth]{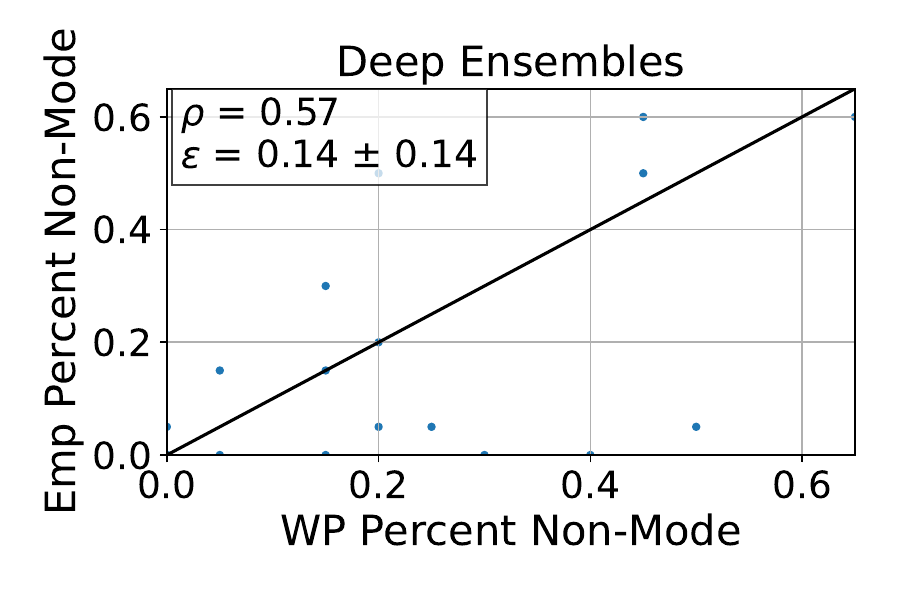}
    \includegraphics[width=0.24\textwidth]{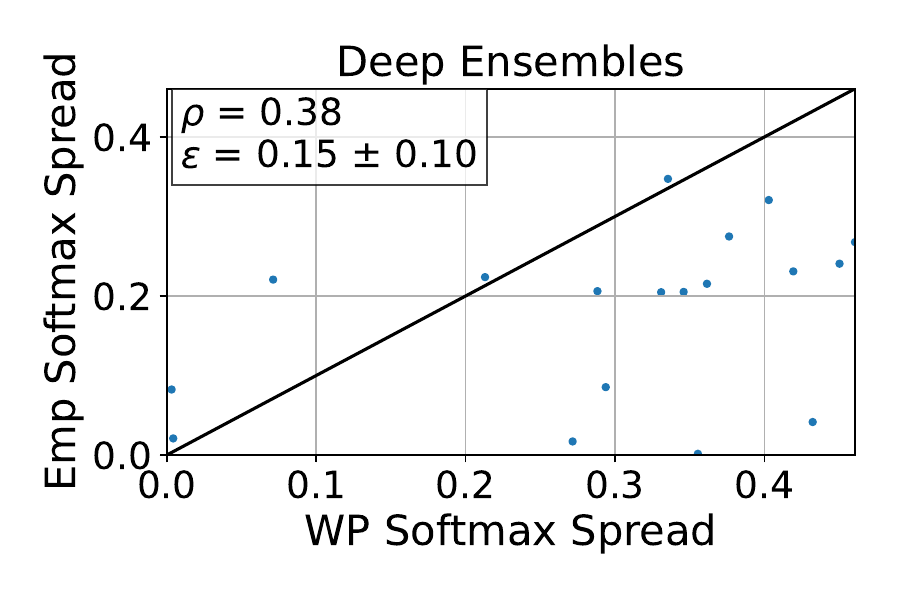}
    \includegraphics[width=0.24\textwidth]{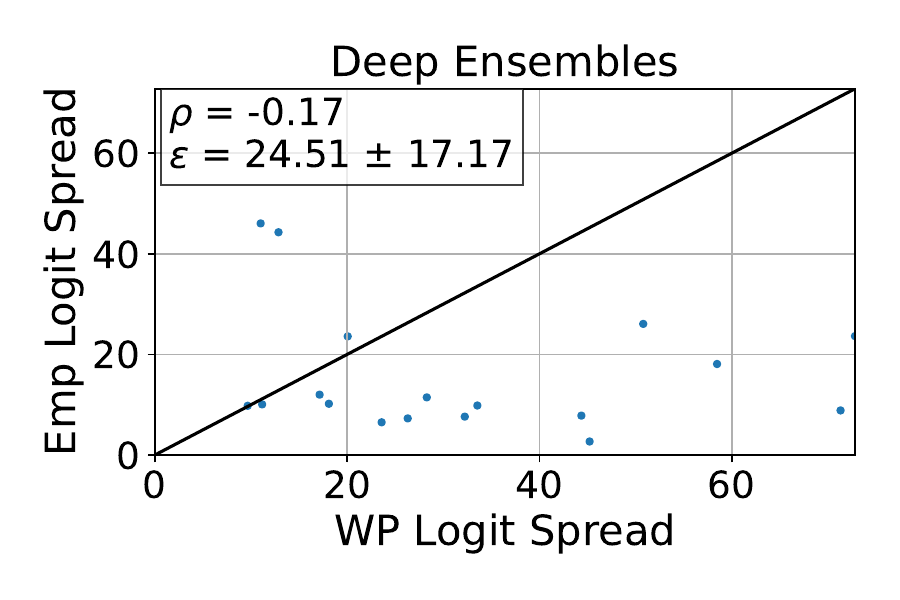}
    
    \includegraphics[width=0.24\textwidth]{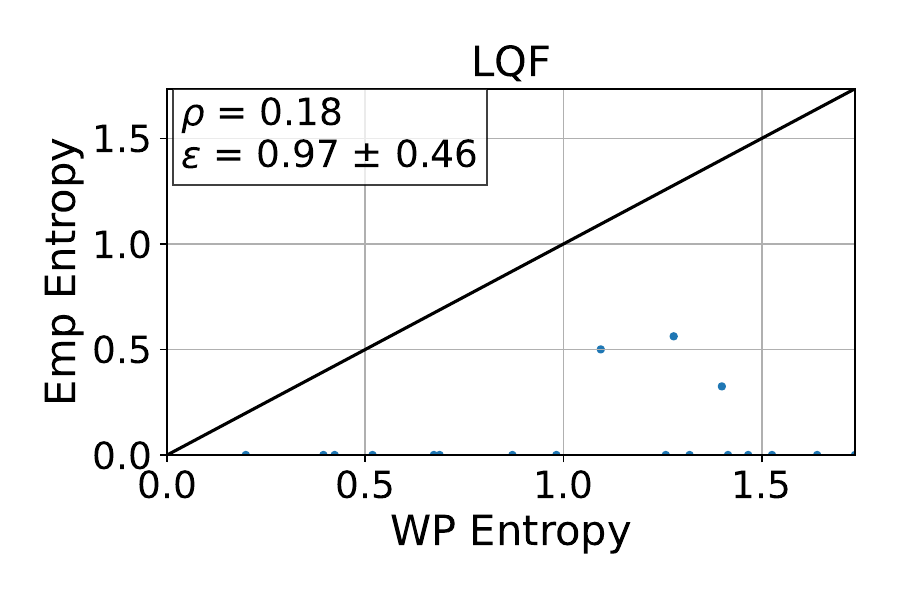}
    \includegraphics[width=0.24\textwidth]{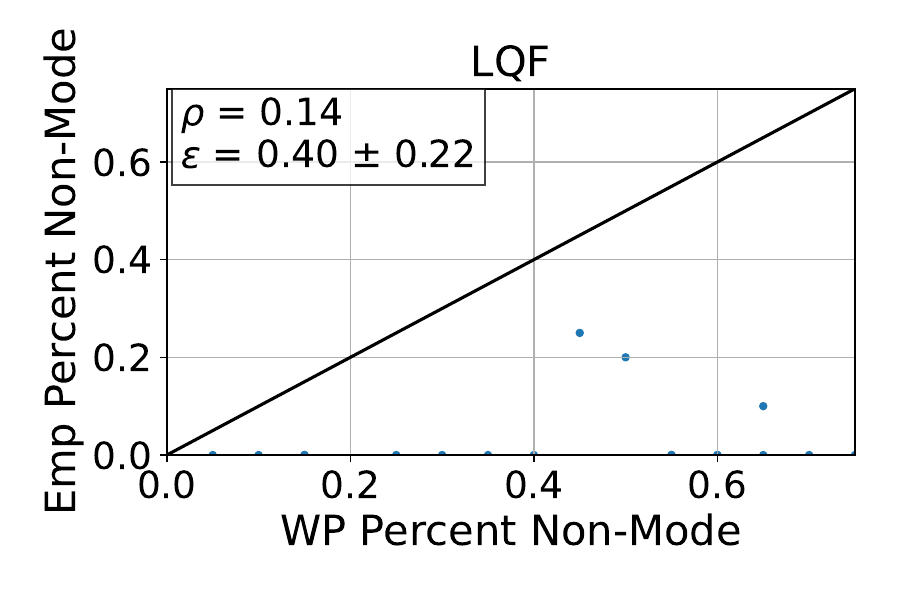}
    \includegraphics[width=0.24\textwidth]{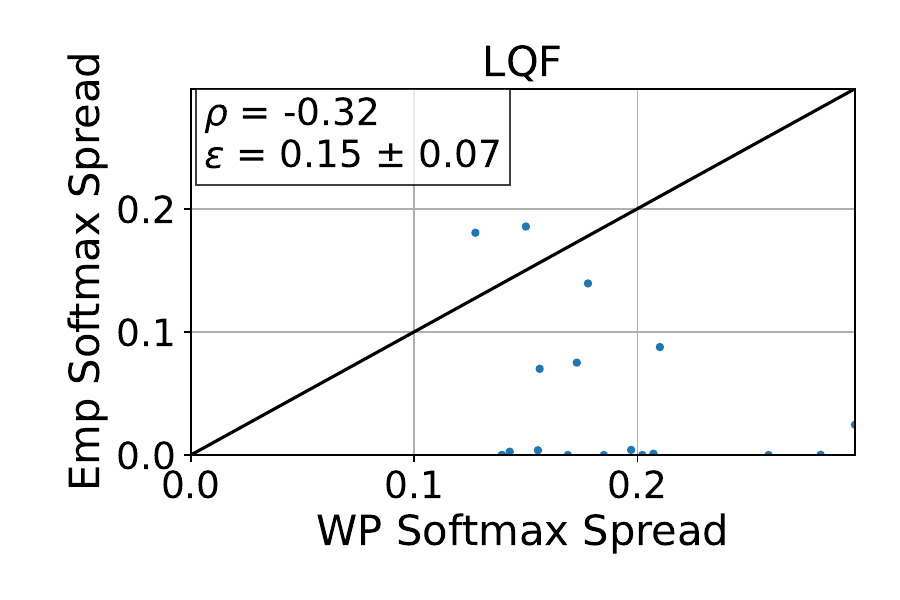}
    \includegraphics[width=0.24\textwidth]{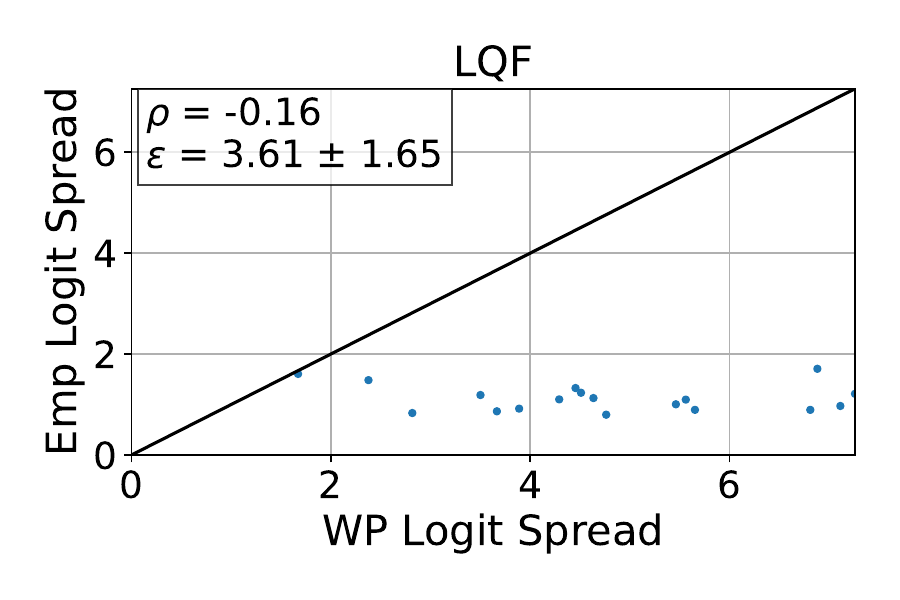}
    
    \caption{\textbf{Empirical Scene Uncertainty vs. Wellington Posterior Scene Uncertainty for Mini Objectron.}  In the above plots, each point is one of 18 scenes from Mini Objectron run through a ResNet-50 network trained on its training set. The point's x-value is an uncertainty measure computed from the Wellington Posterior and the point's y-axis is the uncertainty measure computed from the empirical paragon. The correlation coefficient $\rho$ and the MAE $\varepsilon$ are also displayed. Both MAE and correlation coefficients are poor.}
    \label{fig:WP_performance_MiniObjectron}
\end{figure*}

\begin{table*}[ht]
\small
    \centering
    \begin{tabular}{llcccc}
    \toprule
        & & \multicolumn{2}{c}{\textbf{ResNet-101 (ImgNetVid)}} & \multicolumn{2}{c}{\textbf{ResNet-50 (Objectron)}} \\
    \midrule
        & & Mean Abs. Err & Corr. Coef. & Mean Abs. Err & Corr. Coef \\ 
    \midrule
     \multirow{5}{*}{Logit Spread}
    & Data Augmentation  & 6.1119 $\pm$ 0.2404 & 0.3344 $\pm$ 0.0072 & 9.0711 $\pm$ 0.1610 & \textbf{0.2717 $\pm$ 0.0190} \\
    & Dropout            & 5.4904 $\pm$ 0.1778 & \textbf{0.3787 $\pm$ 0.0104} & 10.0664 $\pm$ 0.5685 & 0.1722 $\pm$ 0.0366 \\
    & Deep Ensembles     & 5.5730 $\pm$ 0.1804 & 0.3307 $\pm$ 0.0152 & 10.0612 $\pm$ 0.3875 & 0.2063 $\pm$ 0.0293 \\
    & LQF                & \textbf{2.2147 $\pm$ 0.0183} & 0.0526 $\pm$ 0.0164 & \textbf{6.7333 $\pm$ 0.2500} & 0.1223 $\pm$ 0.0066 \\   
    \midrule
    \multirow{5}{*}{Softmax Spread}
    & Data Augmentation  & 0.0659 $\pm$ 0.0009 & 0.4984 $\pm$ 0.0124 & 0.0494 $\pm$ 0.0017 & 0.5015 $\pm$ 0.0288 \\
    & Dropout            & 0.0639 $\pm$ 0.0004 & 0.5900 $\pm$ 0.0091 & \textbf{0.0308 $\pm$ 0.0011} & \textbf{0.5886 $\pm$ 0.0115} \\
    & Deep Ensembles     & 0.0622 $\pm$ 0.0022 & 0.5867 $\pm$ 0.0263 & 0.0398 $\pm$ 0.0002 & 0.5298 $\pm$ 0.0168 \\
    & LQF                & \textbf{0.0445 $\pm$ 0.0010} & \textbf{0.6421 $\pm$ 0.0150} & 0.0346 $\pm$ 0.0022 & 0.5298 $\pm$ 0.0203 \\
    & Depth Completion   & - & - & 0.0321 $\pm$ 0.0018 & 0.5494 $\pm$ 0.0418 \\      
    \midrule
    \multirow{5}{*}{Percent Non-Mode}
    & Data Augmentation  & 0.1296 $\pm$ 0.0061 & 0.5453 $\pm$ 0.0121 & 0.0418 $\pm$ 0.0021 & 0.5687 $\pm$ 0.0530 \\
    & Dropout            & 0.1113 $\pm$ 0.0036 & 0.6630 $\pm$ 0.0104 & 0.0286 $\pm$ 0.0026 & 0.5859 $\pm$ 0.0689 \\
    & Deep Ensembles     & 0.1177 $\pm$ 0.0018 & 0.5694 $\pm$ 0.0220 & 0.0311 $\pm$ 0.0014 & \textbf{0.6362 $\pm$ 0.0222} \\
    & LQF                & \textbf{0.0670 $\pm$ 0.0005} & \textbf{0.6787 $\pm$ 0.0259} & 0.0312 $\pm$ 0.0028 & 0.4872 $\pm$ 0.0365 \\
    & Depth Completion   & - & - & \textbf{0.0277 $\pm$ 0.0017} & 0.5560 $\pm$ 0.0491 \\
    \midrule
    \multirow{5}{*}{Scene Entropy}
    & Data Augmentation  & 0.2994 $\pm$ 0.0119 & 0.5991 $\pm$ 0.0153 & 0.0997 $\pm$ 0.0016 & 0.5644 $\pm$ 0.0256 \\
    & Dropout            & 0.2676 $\pm$ 0.0074 & 0.6872 $\pm$ 0.0148 & \textbf{0.0606 $\pm$ 0.0058} & 0.6435 $\pm$ 0.0512 \\
    & Deep Ensembles     & 0.2806 $\pm$ 0.0082 & 0.6120 $\pm$ 0.0295 & 0.0734 $\pm$ 0.0010 & \textbf{0.6583 $\pm$ 0.0052} \\
    & LQF                & \textbf{0.1441 $\pm$ 0.0029} & \textbf{0.7292 $\pm$ 0.0164} & 0.0694 $\pm$ 0.0066 & 0.5543 $\pm$ 0.0133 \\
    & Depth Completion   & - & - & 0.0623 $\pm$ 0.0046 & 0.5940 $\pm$ 0.0473 \\
    \midrule 
        & & \multicolumn{2}{c}{\textbf{ResNet-50 (MiniObjectron)}} & \multicolumn{2}{c}{\textbf{ResNet-50 (SyntheticObjectron)}} \\
    \midrule
        & & Mean Abs. Err & Corr. Coef. & Mean Abs. Err & Corr. Coef \\
    \midrule
    
    \multirow{5}{*}{Logit Spread}
    & Data Augmentation  & 23.5624 $\pm$ 0.6279 & 0.1563 $\pm$ 0.1188 & 25.4542 $\pm$ 2.0959 & \textbf{0.4523 $\pm$ 0.0622} \\
    & Dropout            & 21.3486 $\pm$ 0.4000 & 0.\textbf{2994 $\pm$ 0.1608} & 25.3325 $\pm$ 2.7967 & 0.1561 $\pm$ 0.1306 \\
    & Deep Ensembles     & 22.5527 $\pm$ 1.4580 & 0.0807 $\pm$ 0.1808 & 21.5104 $\pm$ 2.0708 & 0.3648 $\pm$ 0.1205 \\
    & LQF                & \textbf{3.7012 $\pm$ 0.1527} & -0.0218 $\pm$ 0.2006 & \textbf{5.0102 $\pm$ 0.4929} & 0.0573 $\pm$ 0.1499 \\
    \midrule   
    \multirow{5}{*}{Softmax Spread}
    & Data Augmentation  & 0.1653 $\pm$ 0.0105 & 0.2340 $\pm$ 0.1055 & 0.1482 $\pm$ 0.0071 & -0.1280 $\pm$ 0.1384 \\
    & Dropout            & 0.1763 $\pm$ 0.0242 & 0.2999 $\pm$ 0.1407 & 0.1457 $\pm$ 0.0078 & \textbf{0.1410 $\pm$ 0.0926} \\
    & Deep Ensembles     & 0.1508 $\pm$ 0.0260 & \textbf{0.3771 $\pm$ 0.0376} & 0.1277 $\pm$ 0.0131 & 0.0543 $\pm$ 0.1931 \\
    & LQF                & \textbf{0.1433 $\pm$ 0.0084} & -0.1308 $\pm$ 0.0907 & \textbf{0.0932 $\pm$ 0.0130} & 0.0502 $\pm$ 0.2252 \\
    \midrule
    \multirow{5}{*}{Percent Non-Mode}
    & Data Augmentation  & 0.1639 $\pm$ 0.0186 & 0.4484 $\pm$ 0.0436 & 0.2447 $\pm$ 0.0406 & 0.1053 $\pm$ 0.2191 \\
    & Dropout            & 0.1426 $\pm$ 0.0125 & 0.6060 $\pm$ 0.1068 & 0.2542 $\pm$ 0.0186 & 0.0521 $\pm$ 0.1182 \\
    & Deep Ensembles     & \textbf{0.1361 $\pm$ 0.0060} & \textbf{0.6283 $\pm$ 0.0656} & \textbf{0.2382 $\pm$ 0.0192} & 0.0686 $\pm$ 0.0795 \\
    & LQF                & 0.4102 $\pm$ 0.0142 & 0.1503 $\pm$ 0.1005 & 0.4251 $\pm$ 0.0056 & \textbf{0.2259 $\pm$ 0.0452} \\
    \midrule
    \multirow{5}{*}{Scene Entropy}
    & Data Augmentation  & 0.1689 $\pm$ 0.0068 & 0.4975 $\pm$ 0.1185 & 0.2319 $\pm$ 0.0123 & 0.2321 $\pm$ 0.0975 \\
    & Dropout            & 0.1574 $\pm$ 0.0209 & \textbf{0.6594 $\pm$ 0.0592} & 1.2316 $\pm$ 0.0149 & 0.2581 $\pm$ 0.0945 \\
    & Deep Ensembles     & \textbf{0.1395 $\pm$ 0.0139} & 0.6193 $\pm$ 0.1270 & \textbf{0.2116 $\pm$ 0.0132} & 0.2920 $\pm$ 0.0710 \\
    & LQF                & 1.0048 $\pm$ 0.0585 & 0.1325 $\pm$ 0.1932 & 0.9993 $\pm$ 0.0267 & \textbf{0.3277 $\pm$ 0.0968} \\
    \bottomrule
    \end{tabular}
    \caption{ \textbf{Accuracy of Uncertainty Measures Computed from Wellington Posteriors.} 
    This table accompanies \cref{fig:WP_performance_ImgNetVid}, \cref{fig:WP_performance_Objectron}, \cref{fig:WP_performance_MiniObjectron}, and \cref{fig:WP_performance_SyntheticObjectron}. Entries are means and standard deviations of metrics measuring the difference between uncertainty measures computed using the Wellington Posterior and the empirical paragon. Baseline values for each metric are given in \cref{tab:vanilla_spread}. We find that mean absolute errors are generally about as large as the baseline values and that correlation coefficients are low, suggesting that none of the Wellington Posteriors perform well.
    }
    \label{tab:WP_performance_indist}
\end{table*}

Results are shown in \cref{tab:WP_performance_indist}. An illustration of ImageNetVid results is in \cref{fig:WP_performance_ImgNetVid} and an illustration of Objectron results is in \cref{fig:WP_performance_Objectron}. Ultimately, we find that none of the methods used to compute Wellington Posteriors match the empirical paragon particularly well. For the more difficult ImageNetVid dataset, LQF clearly outperforms the other methods. For the easier Objectron, MiniObjectron, and SyntheticObjectron datasets, the performance of all methods are relatively similar. Across the board, the simplest method, data augmentation, has the worst performance.

\section{Conclusion}
We have defined a new concept, \emph{uncertainty with respect to variation in the scene}, and the notion of the Wellington Posterior, $p(y|S(x))$. However, our attempts to model and predict the Wellington Posterior were unsuccessful. We now discuss why and offer potential solutions.

Failure to model and predict the Wellington Posterior in our experiments indicates a discrepancy between the ``ground-truth" or images (or network weights) generated by the model of the Wellington Posterior, i.e. one or both are ``wrong". In Section \ref{sec:experiments}, we noted that 20 frames from videos is surely not ``fair". For a dataset as varied as ImageNetVid, generating a fair ground-truth would require a large database of 3D scenes and a photorealistic rendering engine. For the Objectron dataset, it may suffice to train or fine-tune a NERF \cite{mildenhall_nerfs} and use the NERF's rendering ability to create ground-truth images.

But since results for our small ``fairer" test sets SyntheticObjectron and MiniObjectron are also incredibly poor, it indicates that the models of the Wellington Posterior we tested (data augmentation, ensembling, dropout, LQF, and single-view 3D reconstruction) are also insufficient. Of course, possible directions for future work include fine-tuning StyleGAN2 \cite{stylegan2}, fine-tuning and then sampling from a conditional prior network \cite{Yang_2018_ECCV}, fine-tuning a conditional GAN, or using another method altogether. Another possibility may be to use LQF, but with different values of $\Sigma_w$. These proposed methods are all difficult because they either require large amounts of computational power (StyleGAN2, LQF) or have never been attempted (sampling from a CPN).

Although this manuscript is focused exclusively on image classification, the concepts of Scene Uncertainty and the Wellington Posterior may apply to other tasks, such as object detection and image segmentation. Potential lines of future work may entail modeling the Wellington Posterior for these other tasks. The nature of those future works may be very different from what is described in this manuscript -- for example, if the distributions of intrinsic and extrinsic variabilities and nuisances are known, then it may be possible in some applications to measure Scene Uncertainty directly, so that modeling the Wellington Posterior would become unnecessary.

For most sensors used in robotics and safety critical systems, the scene uncertainty at any state can be safely assumed to be a normal distribution in algorithm design, or something. Simply knowing the distribution of measurement errors has allowed engineers to create safety-critical technologies relying on imperfect sensors. Understanding Scene Uncertainty of neural networks in this way, will be key to incorporating imperfect neural networks into safety-critical systems to perform functions that could not be performed otherwise.

{\small
\bibliographystyle{ieee_fullname}
\bibliography{references_neurips.bib, references.bib}
}

\newpage

\appendix

\section{Additional Experiment Details}

\subsection{Network Hyperparameters and Classifier Performance}
 Hyperparameters, training error, and validation error for training standard image classifiers using the cross entropy loss are given in \cref{tab:standard_network_hyperparameters}. Hyperparameters, training error, and validation error for training LQF networks are given in \cref{tab:lqf_hyperparameters}. Pretrained ImageNet networks were fine-tuned for one epoch on ImageNet with ReLUs replaced by LeakyReLUs prior to LQF with the hyperparameters in \cref{tab:lqf_imagenet_finetune}. When training all networks, we saved checkpoints along with its training and validation accuracy every 10 epochs. We used a combination of training and validation errors to select an appropriate checkpoint for the rest of the experiments. Training networks with dropout for the experiments with dropout used the same hyperparameters as standard networks. Training and validation errors are shown in \cref{tab:dropout_hyperparameters}.

To train ensembles of standard networks and LQF networks, we used the same hyperparameters and procedure as for the baseline networks. The only difference is that each ensemble saw different data shuffling and a different 90\% fraction of the data.

Test error on the Objectron test set was 0.9607 $\pm$ 0.0015 for all frames and 0.9693 $\pm$ 0.0076 for only the anchor frames. Test error on ImageNetVid-Robust was 0.7187 $\pm$ 0.0084 for all frames and 0.7207 $\pm$ 0.0139 for only the anchor frames. These test errors do not reflect the variation shown in \cref{fig:spread_is_uncertainty}.

\begin{table}[ht]
\centering
\footnotesize
\begin{tabular}{lcc}
\toprule
\textbf{Parameter} & \textbf{Objectron} & \textbf{ImageNetVid} \\
\midrule
Architecture & ResNet-50 & ResNet-101 \\
Init. Learning Rate & 0.01 & 0.001 \\
Momentum & 0.9 & 0.9 \\
Weight Decay & 1e-5 & 1e-5 \\
$\gamma$ &  0.1 & 0.1\\
Milestones &  25, 35 & 25, 35 \\
Max Epochs & 50 & 50 \\
Batch Size & 32 & 32 \\
Selected Epoch & 50, 40, 40 & 30, 50, 50 \\
Training Error & 0.0326,0.0366,0.0318 & 0.0478,0.0478,0.0476\\
Validation Error & 0.0320,0.0220,0.0320 & 0.2747,0.2711,0.2912\\
\bottomrule
\end{tabular}
\caption{Hyperparameters and decision variables used in fine-tuning image classifiers on the Objectron and ImageNetVid datasets using stochastic gradient descent and the cross-entropy loss for all three trials. Reported training and validation errors are for the selected epoch, not after the maximum number of epochs.}
\label{tab:standard_network_hyperparameters}
\end{table}

\begin{table}[ht]
\footnotesize
\centering
\begin{tabular}{lcc}
\toprule
\textbf{Parameter} & \textbf{Objectron} & \textbf{ImageNetVid} \\
\midrule
Architecture & ResNet-50 & ResNet-101 \\
Init. Learning Rate & 5e-4 & 5e-4 \\
Weight Decay & 1e-5 & 1e-5 \\
Max Epochs & 50 & 50 \\
$\alpha$ & 15.0 & 15.0 \\
Batch Size & 32 & 32 \\
Selected Epoch & 50, 40, 50 & 40, 30, 40 \\
Training Error & 0.0186,0.0220,0.0180 & 0.0500,0.0542,0.0457\\
Validation Error & 0.0000,0.0200,0.0280 & 0.2692,0.2601,0.2674\\
\bottomrule
\end{tabular}
\caption{Hyperparameters and decision variables used in fine-tuning image classifiers on the Objectron and ImageNetVid datasets using LQF. Our LQF procedure used the AdamW optimizer with a one-hot mean squared error loss. Training and validation errors are at the selected epoch, not after the maximum number of epochs.}
\label{tab:lqf_hyperparameters}
\end{table}

\begin{table}[ht]
\footnotesize
\centering
\begin{tabular}{lcc}
\toprule
\textbf{Parameter} & \textbf{Value (ResNet-50)} & \textbf{Value (ResNet-101)} \\
\midrule
Batch Size & 32 & 32 \\
Learning Rate & 1e-3 & 1e-3 \\
ReLU Leak & 0.2 & 0.2 \\
Training Error & 0.43 & 0.40 \\
Validation Error & 0.41 & 0.31 \\
\bottomrule
\end{tabular}
\caption{{\sl Hyperparameters used to fine-tune pretrained ResNet-50 and ResNet-101 networks on ImageNet with LeakyReLUs replacing ReLUs prior to LQF using the cross-entropy loss. Note that the goal of this fine-tuning was to slightly adjust pretrained weights towards values that accommodate LeakyReLUs, not to achieve low training and/or validation error, as these are not the final weights used in our experiments.}}
\label{tab:lqf_imagenet_finetune}
\end{table}

\begin{table}[ht]
\footnotesize
\centering
\begin{tabular}{lcc}
\toprule
\textbf{Parameter} & \textbf{Objectron} & \textbf{ImageNetVid} \\
\midrule
Max Epochs & 50 & 50 \\
Batch Size & 32 & 32 \\
Selected Epoch & 40, 50, 50 & 50, 50, 50 \\
Training Error & 0.0490, 0.0504, 0.0488 &  0.1110, 0.1079, 0.1084\\
Validation Error & 0.0380, 0.0240, 0.0380 & 0.2527, 0.2619, 0.2564\\
\bottomrule
\end{tabular}
\caption{{\sl Training and validation errors for networks trained with dropout. As expected, training errors are higher. Validation errors are lower than for standard networks for the more difficult ImageNetVid dataset, but comprable for the easy Objectron dataset, as is typically observed when training with dropout. Training and validation errors are reported for the selected epoch, not the maximum number of epochs.}}
\label{tab:dropout_hyperparameters}
\end{table}

\subsection{3D Reconstruction}
\label{sec:3D_reproj_imgs}

\begin{figure*}[ht]
\centering
\includegraphics[width=1.65in]{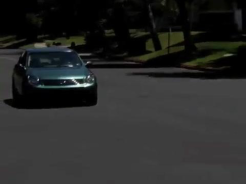}
\includegraphics[width=1.65in]{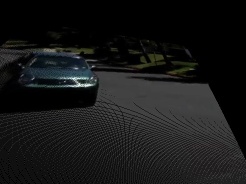}
\includegraphics[width=1.65in]{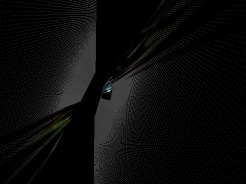}
\includegraphics[width=1.65in]{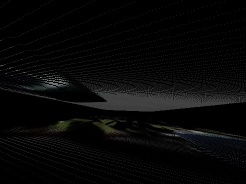}
\includegraphics[width=1.65in]{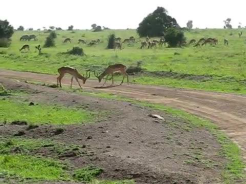}
\includegraphics[width=1.65in]{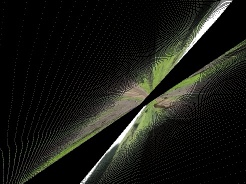}
\includegraphics[width=1.65in]{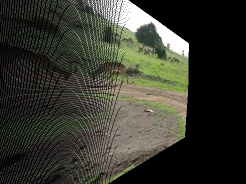}
\includegraphics[width=1.65in]{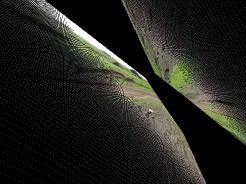}
\includegraphics[width=1.65in]{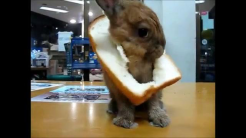}
\includegraphics[width=1.65in]{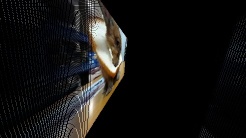}
\includegraphics[width=1.65in]{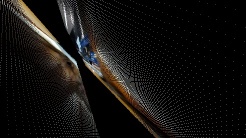}
\includegraphics[width=1.65in]{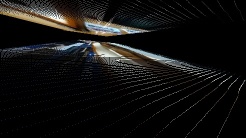}
\caption{ \textbf{Examples of reprojected ImageNetVid-Robust images after 3D monocular depth estimation.} The left column contains anchor frames from the ImageNetVid-Robust dataset. Other images are generated through depth estimation, random small camera shifts, and reprojection. The reprojections look like perspective transformations, showing that the depth estimation was incorrectly predicting near-constant depth. The presence of thick black borders around many ImageNetVid-Robust images, not present in any dataset used to train 3D monocular depth estimation networks, likely further lowered the quality of estimated depth.}
\label{fig:imgnetvid_3d_test}
\end{figure*}

We attempted to use state-of-the-art monocular 3D depth estimation \cite{Yin2019enforcing} with small camera movements drawn from the same distributions as in the previous section. The depth estimation network was pretrained on the KITTI dataset of outdoor driving scenes. Since the ImageNetVid-Robust dataset does not contain any camera intrinsics, we assumed that the center of the camera apeture was located in the middle of the frame and that the focal length was equal to 1. The results, shown in Figure \ref{fig:imgnetvid_3d_test}, led us to conclude that 3D monocular depth estimation on ImageNetVid images is not possible without fine-tuning. Furthermore, the lack of any (sparse or dense) depth information or camera intrinsics will make fine-tuning extremely difficult.

For the Objectron dataset, we trained a VGG11-based depth completion network \cite{VOICED} using hyperparameters given in \cref{tab:voiced_hyperparameters}. 
\begin{table}[]
\footnotesize
    \centering
    \begin{tabular}{lc}
    \toprule
    \textbf{Parameter} & \textbf{Value} \\
    \midrule
    Training Steps & 330,000 \\
    Learning Rate & 5e-5 \\
    Minimum Predicted Depth (m) & 0.01 \\
    Maximum Predicted Depth (m) & 5.0 \\
    Batch Size & 8 \\
    Pose Norm & Frobenius \\
    Rotation Representation & Exponential \\
    Occlusion Threshold (m) & 1.5 \\
    Occlusion Threshold Kernel Size & 7 \\ 
    $w_{ph}$ & 1.00 \\
    $w_{co}$ & 0.20 \\
    $w_{st}$ & 0.80 \\
    $w_{sm}$ & 1.00 \\
    $w_{sz}$ & 1.00 \\
    $w_{pc}$ & 0.10 \\
    \bottomrule
    \end{tabular}
    \caption{Hyperparameters used for training VOICED \cite{VOICED} on the Objectron dataset.}
    \label{tab:voiced_hyperparameters}
\end{table}

\begin{figure*}
\centering
\includegraphics[width=0.45\textwidth]{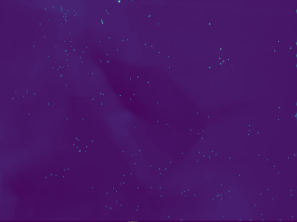}
\includegraphics[width=0.45\textwidth]{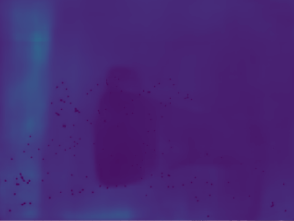}
\caption{\textbf{Examples of raw depth completion output for the Objectron dataset.} The depth completion network was not able to learn true metric depth in all scenes --- for some scenes, it would learn relative depth for all but the sparse points that existed in the dataset, and then copy the sparse points to its depth map. This resulted in a depth map that contained many specks. We believe that this was the result of large variations in distributions of depth measurements in the Objectron dataset combined with batch sizes that were too small to capture all the variation simultaneously during training. Since a larger batch would have required more computational resources, and since perfect metric depth estimation is not necessary for our experiments, we remedied the results with a median filter with window size 5.}
\label{fig:voiced_specs}
\end{figure*}

\begin{figure*}
\centering
\includegraphics[width=1.65in]{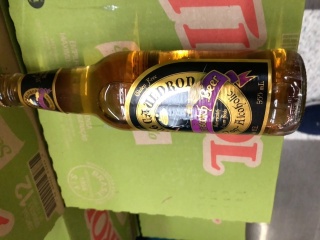}
\includegraphics[width=1.65in]{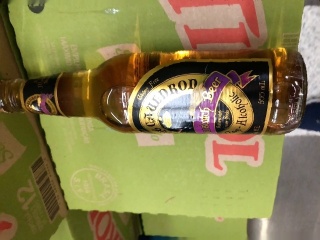}
\includegraphics[width=1.65in]{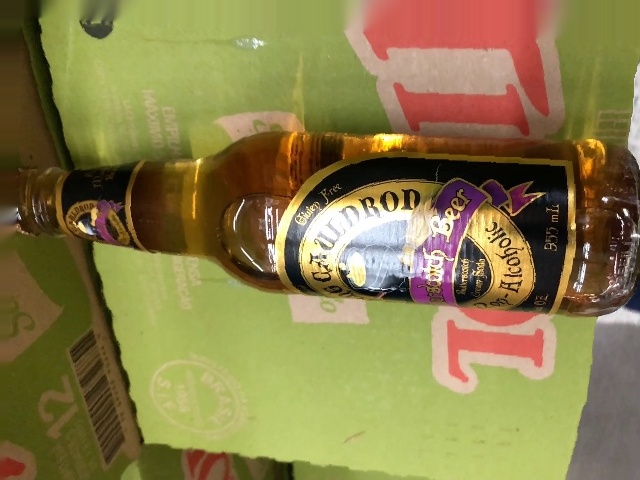}
\includegraphics[width=1.65in]{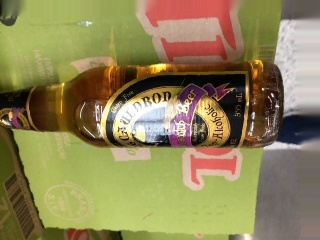}
\includegraphics[width=1.65in]{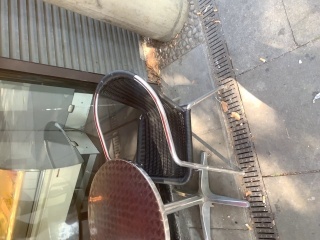}
\includegraphics[width=1.65in]{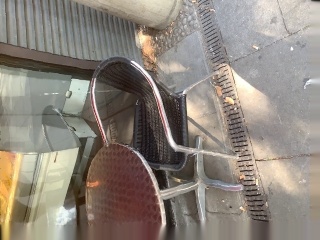}
\includegraphics[width=1.65in]{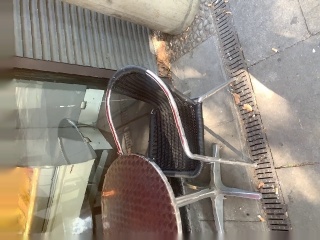}
\includegraphics[width=1.65in]{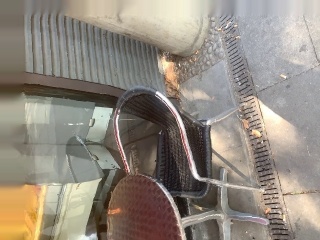}
\includegraphics[width=1.65in]{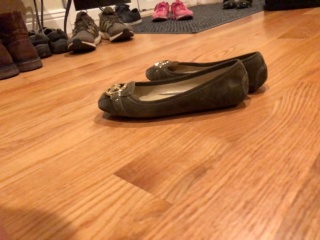}
\includegraphics[width=1.65in]{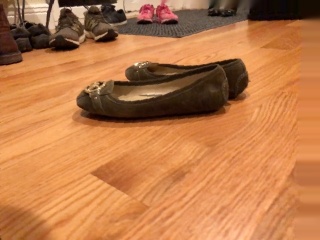}
\includegraphics[width=1.65in]{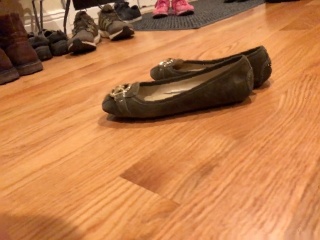}
\includegraphics[width=1.65in]{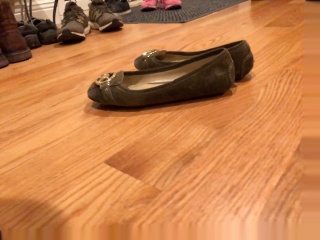}
\caption{\textbf{Examples of reprojected Objectron images after 3D monocular depth completion.} Original anchor images are shown in the left column and reprojected images with small random camera rotations and translations are shown in the rest of the columns. After postprocessing the raw depth images shown in \cref{fig:voiced_specs}, the reprojections (with some classical inpainting) do not contain any noticeable unnatural artifacts.}
\end{figure*}

\subsection{GAN Ensembling Images}
\label{sec:gans_are_lulz}

\begin{figure*}
    \centering
    \includegraphics[width=0.24\textwidth]{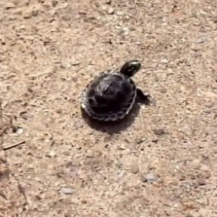}
    \includegraphics[width=0.24\textwidth]{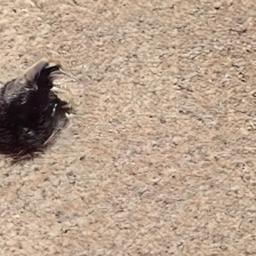}
    \includegraphics[width=0.24\textwidth]{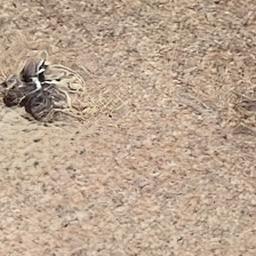}
    \includegraphics[width=0.24\textwidth]{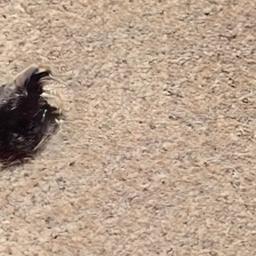}
    \includegraphics[width=0.24\textwidth]{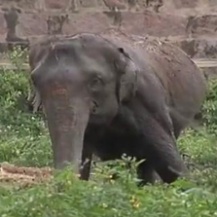}
    \includegraphics[width=0.24\textwidth]{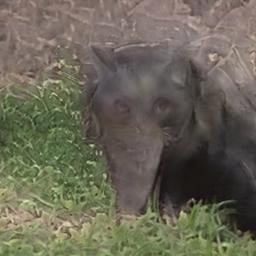}
    \includegraphics[width=0.24\textwidth]{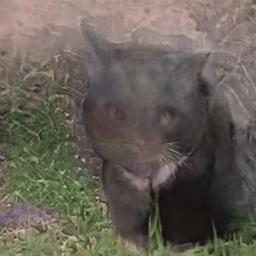}
    \includegraphics[width=0.24\textwidth]{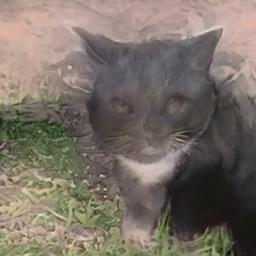}
    \includegraphics[width=0.24\textwidth]{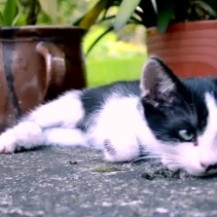}
    \includegraphics[width=0.24\textwidth]{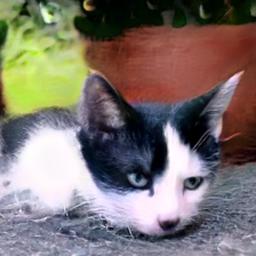}
    \includegraphics[width=0.24\textwidth]{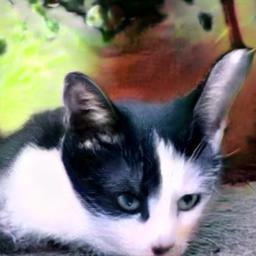}
    \includegraphics[width=0.24\textwidth]{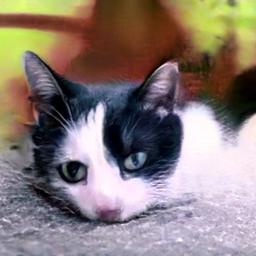}   
    \caption{\textbf{GAN-Ensembled ImageNetVid-Robust anchor images.} Original images are in the left column. The second column is the projection of the original image into the latent space of a StyleGAN2 \cite{stylegan2} network trained on cat faces. The third and fourth columns are perturbations using coarse isotropic noise and coarse principal component analysis, as described in \cite{cGAN}. Only images of cats are well-perturbed, leading to a Wellington Posterior with poor performance.}
    \label{fig:gan_imgnetvid_cats}
\end{figure*}

\begin{figure*}
    \centering
    \includegraphics[width=0.24\textwidth]{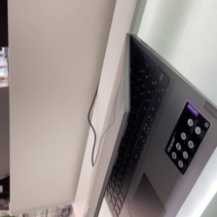}
    \includegraphics[width=0.24\textwidth]{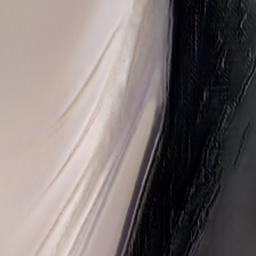}
    \includegraphics[width=0.24\textwidth]{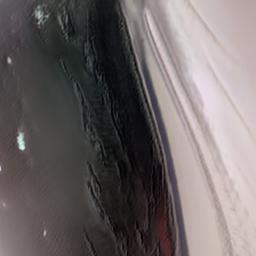}
    \includegraphics[width=0.24\textwidth]{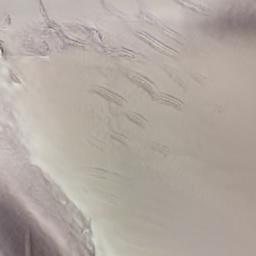}
    \includegraphics[width=0.24\textwidth]{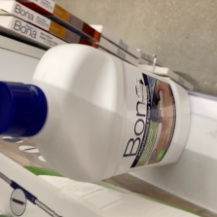}
    \includegraphics[width=0.24\textwidth]{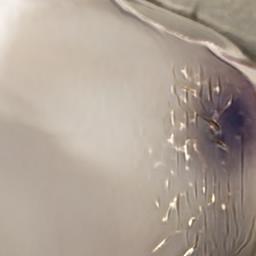}
    \includegraphics[width=0.24\textwidth]{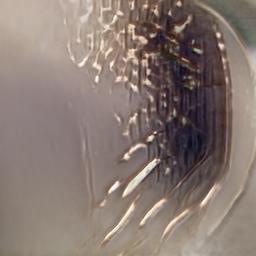}
    \includegraphics[width=0.24\textwidth]{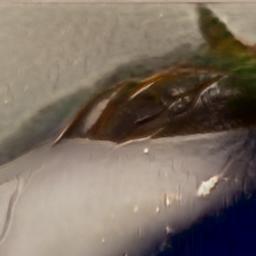}
    \includegraphics[width=0.24\textwidth]{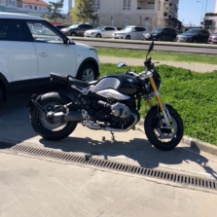}
    \includegraphics[width=0.24\textwidth]{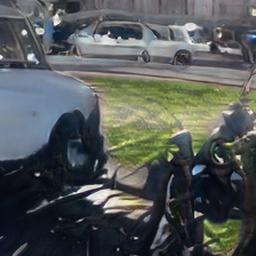}
    \includegraphics[width=0.24\textwidth]{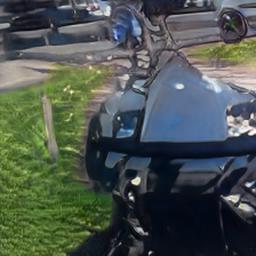}
    \includegraphics[width=0.24\textwidth]{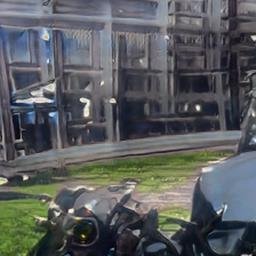}   
    \caption{\textbf{GAN-Ensembled Objectron anchor images.} Original images are in the left column. The second column is the projection of the original image into the latent space of a StyleGAN2 \cite{stylegan2} network trained on images of cars. The third and fourth columns are perturbations using coarse isotropic noise and coarse principal component analysis, as described in \cite{cGAN}. All images are poorly perturbed leading to a Wellington Posterior with poor performance. Unsurprisingly, in the image of the motorcycle, the motorcycle was erased and the automobile in the background becomes more prominent.}
    \label{fig:gan_objectron_cars}
\end{figure*}

GAN Ensembling \cite{cGAN} projects images into the latent space of a pretrained StyleGAN2 \cite{stylegan2}, and then generates new images similar to the original one through perturbations in the latent space. Results from using this technique to generate Wellington Posteriors are shown in \cref{fig:gan_imgnetvid_cats} and \cref{fig:gan_objectron_cars}. Because the GANs were not fine-tuned onto the Objectron and ImageNetVid datasets, performance of the Wellington Posteriors was so poor that we chose not to include them in \cref{tab:WP_performance_indist} and defer fine-tuning StyleGAN2 as future work.

\subsection{Computational Resources}
All experiments were performed on a workstation with the following specs:
\begin{itemize}
    \item CPU (8 cores): Intel(R) Core(TM) i7-6850K CPU @ 3.60GHz
    \item RAM: 64 GB
    \item GPU 0: NVIDIA TITAN V, 12GB RAM
    \item GPU 1: NVIDIA GeForce GTX 1080, 11GB RAM
    \item Python 3.9
    \item PyTorch 1.8
\end{itemize}

\section{Logit Covariance Using LQF}
\label{sec:lqf_details}

\paragraph{Computing Logit Covariance from Diagonal Weight Covariance}
\label{sec:lqf_compute_sigmay}
\cref{eq:lqfcov} states that the formula for the covariance of the logits $\Sigma_y$ is a linear transformation of the covariance of the weights $\Sigma_w$. Calculating $\Sigma_y$ from $\Sigma_w$ using standard matrix multiplication requires more computer memory than is available on a typical workstation. Therefore, we compute $\Sigma_y$ column by column using the following procedure that performs $2K$ forward passes and $K$ backward passes per image. Please see function \texttt{getOutCov} in our attached source code for more detail:

For each class $k$, corresponding to column $k$ of $\Sigma_y$:
\begin{itemize}
\item In the network object, set all values of $\bar w$ to 0.
\item Compute dummy output $\xi$ and dummy loss function (first forward pass), where $\tilde y_k$ is a one-hot encoding of the correct class:
    \begin{equation*}
    \begin{aligned}
    \zeta &= \nabla_w f_{w_0}(x) \cdot 0  \\
    \tilde L &= \| \tilde z - \tilde y_k \|_2^2
    \end{aligned}
    \end{equation*}
\item Run a backward pass. PyTorch will compute and store $\nabla_w f_{w_0}$, even though $\tilde L$ is not actually dependent on $\bar w$.
\item Multiply the stored values of $\nabla_w f_{w_0}$ by the values in the diagonal approximation of $\Sigma_w$.\footnote{This step is implemented as a Pytorch preconditioner that edits the values of stored gradients.} Copy these values to the variable $\bar w$.\footnote{This step is implemented using SGD with a learning rate of 1.0.}
\item Run a second forward pass to compute $\zeta' = \nabla_w f_{w_0}(x) \Sigma_w \nabla_w f_{w_0}(x)^\top$. ($\zeta'$ is column $k$ of $\Sigma_y$.)
\end{itemize}

The above procedure is complex and uses the PyTorch optimizer in a manner it was not designed for. Next, $2K$ forward passes and $K$ backward passes per image, although computationally feasible, is sufficient for our study, but too slow to be practical. Using our computational resources, computing $\Sigma_y$ for all anchor images of the ImageNetVid test set requires ~3 hours with one GPU.

\subsection{Finding a Value for Weight Covariance}
In our experiments, we set $\Sigma_w$ to the sample variance of the linear weights. The sample variance came from an ensemble of 20 trained LQF networks. However, it is also admissible to treat $\Sigma_w$ as a tuning parameter over a validation dataset, the only requirement is that $\Sigma_w$ be a positive semi-definite matrix. For now, though, the computationally slow procedure to compute $\Sigma_z$ from $\Sigma_w$ described in the previous section makes the sample variance the most practical value for $\Sigma_w$, as each iteration of tuning requires $\sim 3$ hours while training an ensemble of 20 LQF networks requires only $\sim 11$ hours and yields reasonable results.

\section{Pseudo-Uncertainties and Scene Entropy}
\label{sec:scene_entropy}
This section details preliminary work that ruled out the softmax vector and other pseudo-uncertainties as a method for computing the Wellington Posterior, as stated in Section \ref{sec:pseudo_uncertainty}. 

The maximum possible entropy for any classification problem with $K$ classes is:
\begin{equation}
\begin{aligned}
    H_{max}(K) &= - \sum_{k=1}^K \frac{1}{K} \log \left ( \frac{1}{K} \right ) \\
               &= - \log \left ( \frac{1}{K} \right )
    \label{eq:max_scene_entropy}
\end{aligned}
\end{equation}
Therefore, the normalized scene entropy
\begin{equation}
    U_{f_w}(P(\hat k | S(x))) := \frac{H(P(\hat k | S(x)))}{H_{max}(K)}.
\end{equation}
may also be taken as a one possible measure of uncertainty based on the scene distribution, in addition to the measures noted in Section \ref{sec:measuring_uncertainty}. Note that predicting scene entropy or normalized scene entropy is not the same as creating an accurate Wellington Posterior, as two different distributions may have the same entropy. Therefore, predicting normalized scene entropy is an \emph{easier} problem than creating an accurate Wellington Posterior, as two different categorical distributions may have the same entropy. We evaluate predictions of normalized scene entropy using mean absolute error.

\begin{figure*}
    \centering
    \subfloat[Softmax Vector]{\includegraphics[width=1.7in]{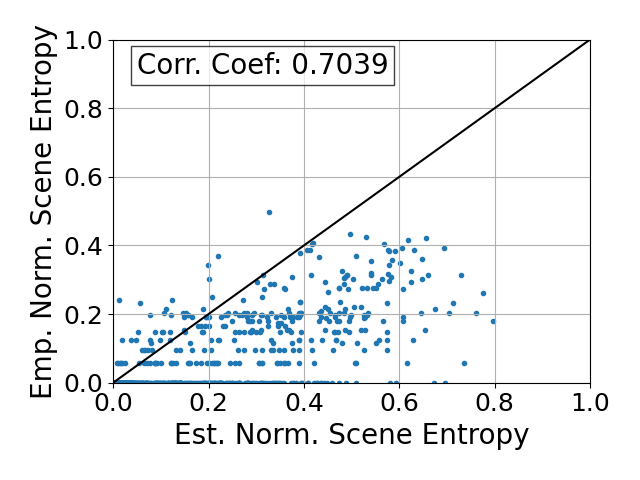}}
    \subfloat[Temp. Scaling]{\includegraphics[width=1.7in]{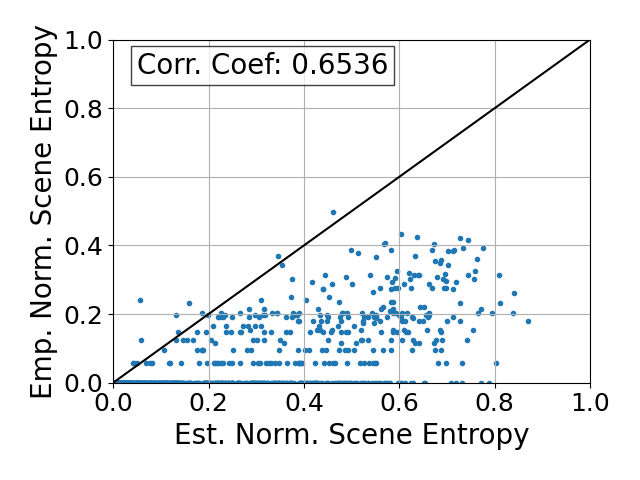}}
    \subfloat[Isotonic - max softmax]{\includegraphics[width=1.7in]{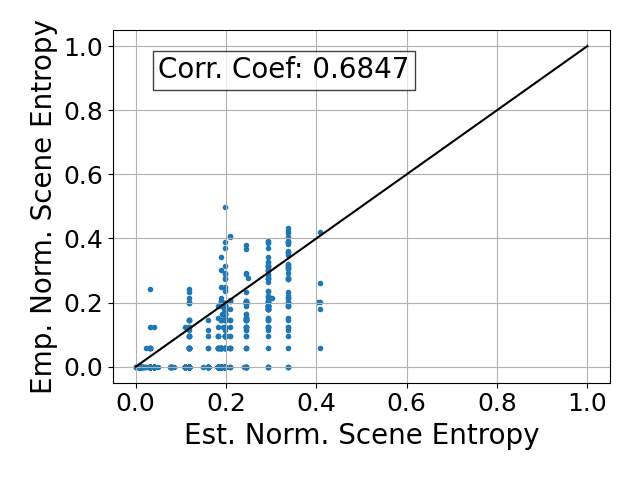}}
    \subfloat[Isotonic - entropy]{\includegraphics[width=1.7in]{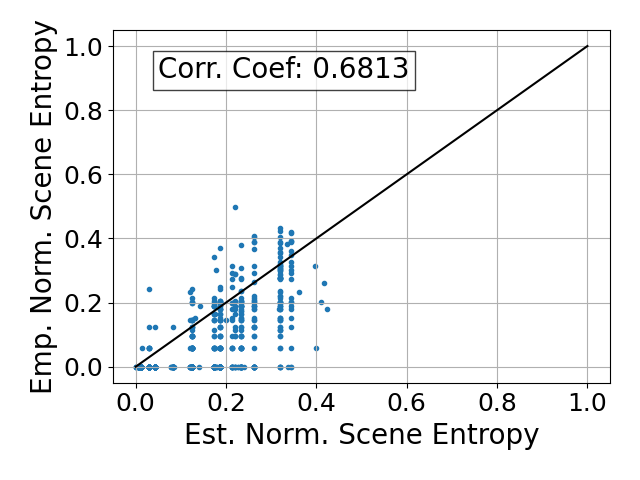}}
    \\
    \subfloat[Isotonic - energy]{\includegraphics[width=1.7in]{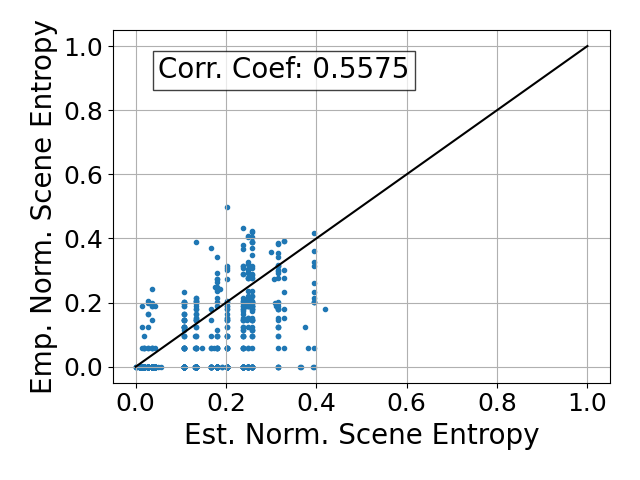}}
    \subfloat[Aux. Net - logits]{\includegraphics[width=1.7in]{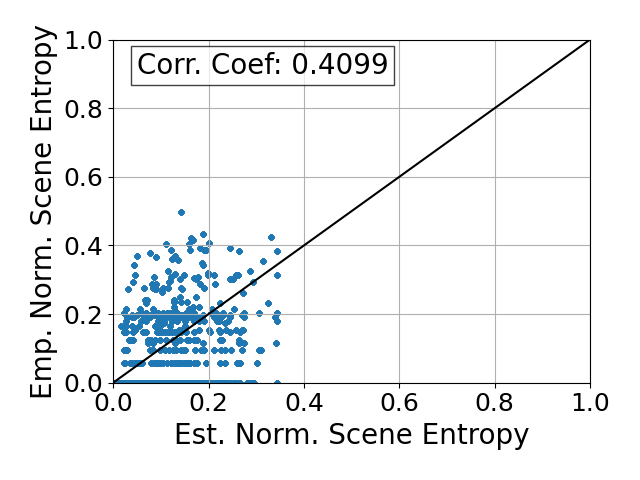}}
    \subfloat[Aux. Net - softmax]{\includegraphics[width=1.7in]{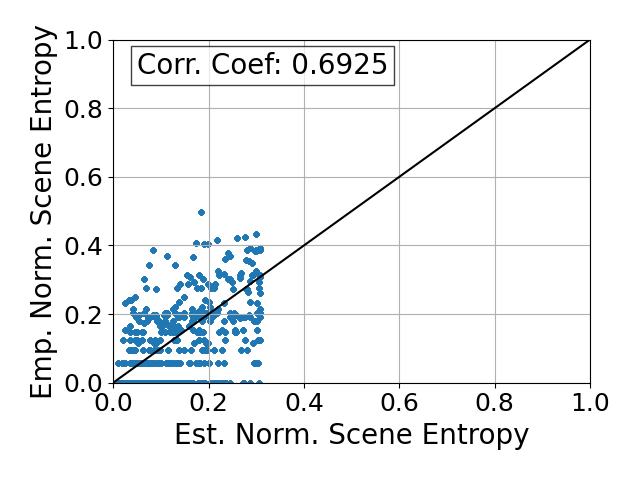}}
    \subfloat[Aux. Net - embed.]{\includegraphics[width=1.7in]{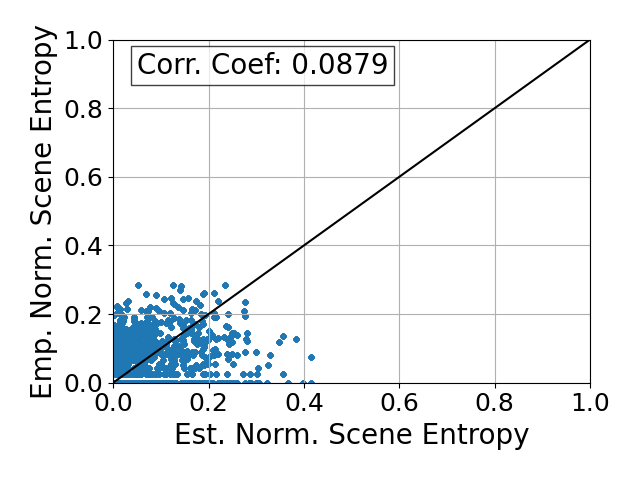}}
    \caption{\textbf{Normalized scene entropy of ImageNetVid-Robust videos cannot be predicted from the anchor frame's logits.} In the plots above, each point represents a scene. The y-axis of each point is normalized scene entropy computed using the empirical paragon while the x-axis of each point is normalized scene entropy estimated from the output of a single forward pass over the anchor frame. If normalized scene entropy were predictable, then points would lie along the black line.}
    \label{fig:ImgNetVid_SUCE}
\end{figure*}

\begin{figure*}[ht]
    \centering
    \subfloat[Softmax Vector]{\includegraphics[width=1.7in]{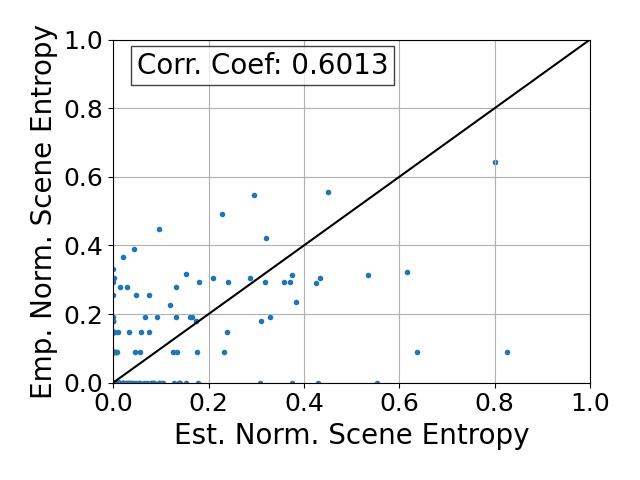}}
    \subfloat[Temp. Scaling]{\includegraphics[width=1.7in]{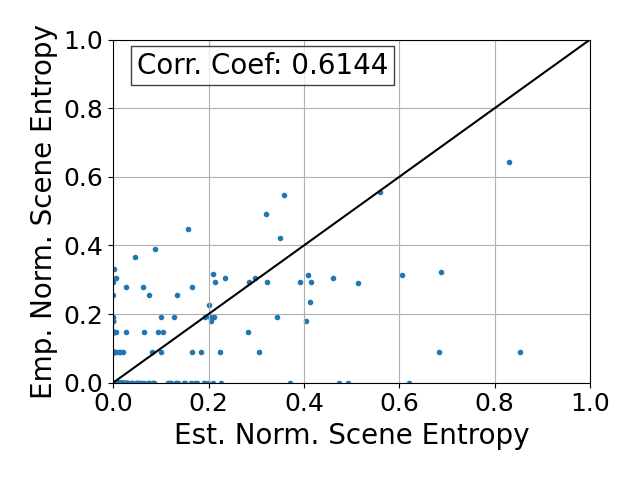}}
    \subfloat[Isotonic - max softmax]{\includegraphics[width=1.7in]{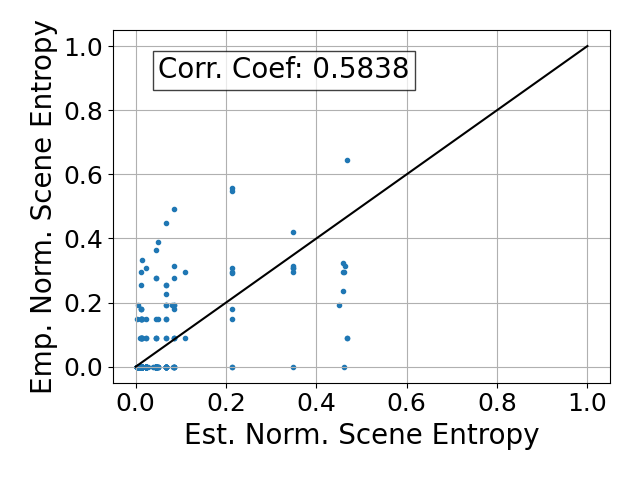}}
    \subfloat[Isotonic - entropy]{\includegraphics[width=1.7in]{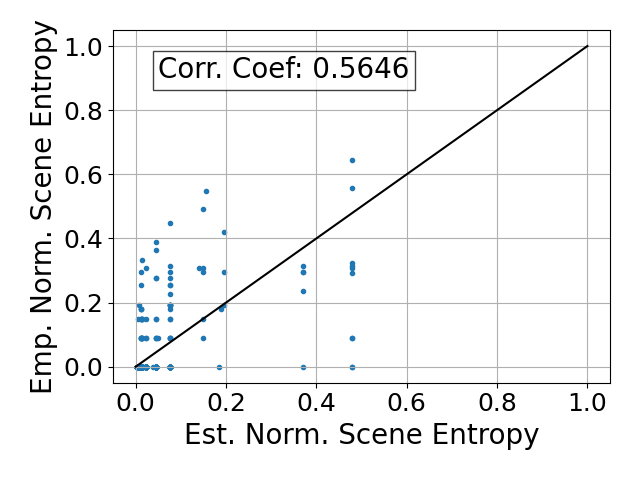}}
    \\
    \subfloat[Isotonic - energy]{\includegraphics[width=1.7in]{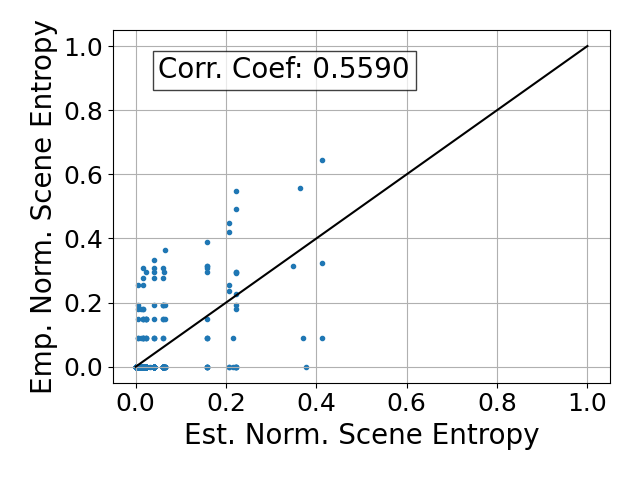}}
    \subfloat[Aux. Net - logits]{\includegraphics[width=1.7in]{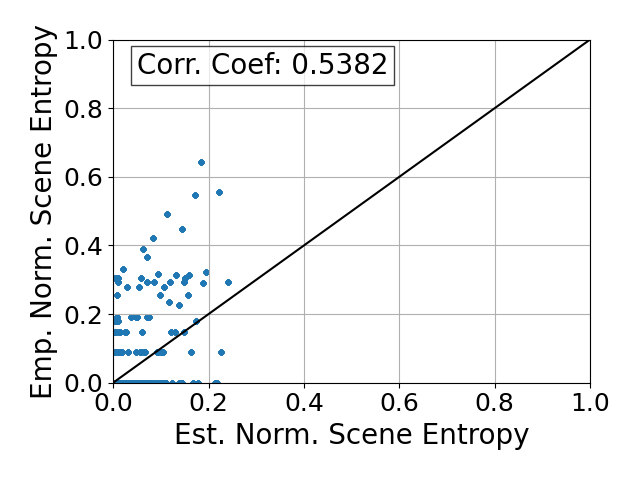}}
    \subfloat[Aux. Net - softmax]{\includegraphics[width=1.7in]{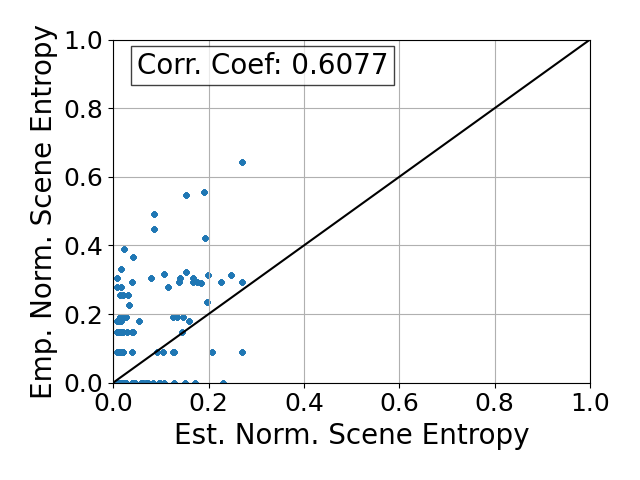}}
    \subfloat[Aux. Net - embed.]{\includegraphics[width=1.7in]{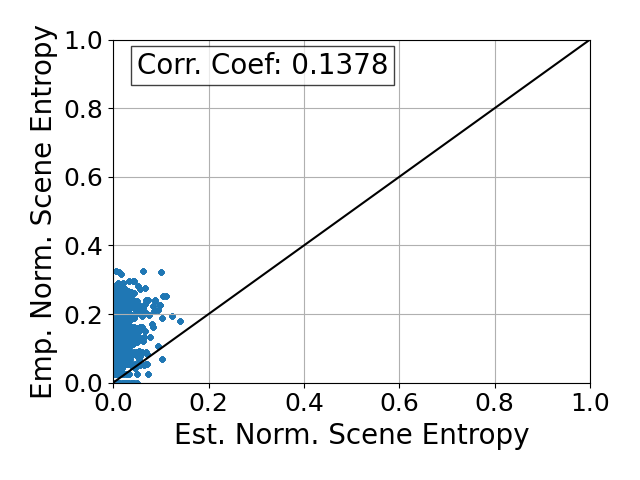}}
    \caption{\textbf{Normalized scene entropy of Objectron videos cannot be predicted from the anchor frame's logits.} In the plots above, each point represents a scene. The y-axis of each point is normalized scene entropy computed using the empirical paragon while the x-axis of each point is normalized scene entropy estimated from the output of a single forward pass over the anchor frame. Compared to ImageNetVid-Robust (Figure \ref{fig:ImgNetVid_SUCE}), the graphs appear sparse because most frames are in the bottom-right corner. This is due to the fact that Objectron is a much easier dataset, and almost all frames of all videos are correctly classified, which makes a typically overconfident softmax vector close to correct. However, for videos for which empirical normalized scene entropy is nonzero, i.e. there is some variation in prediction, the methods above are not able to predict the amount of variation.}
    \label{fig:obj_SUCE}
\end{figure*}

To estimate normalized scene entropy, we consider the softmax vector, temperature scaled softmax vectors, isotonic regression \cite{isotonic_regression}, and auxiliary networks. Lackluster summary results for both ImageNetVid and Objectron are shown in Figure \ref{fig:ImgNetVid_SUCE}, Figure \ref{fig:obj_SUCE}, and Table \ref{tab:SUCE_table}. More details on each particular method are given in the subsections below.

\begin{table}[ht]
\centering
\footnotesize
\begin{tabular}{lcc}
\toprule
\textbf{Method} & \textbf{Objectron} & \textbf{ImageNetVid} \\
\midrule
Softmax Vector & \textbf{0.0317	$\pm$ 0.0025} & 0.1148 $\pm$ 0.0058 \\
Temperature Scaling & 0.0347 $\pm$ 0.0023 & 0.2043 $\pm$ 0.0063\\
Isotonic Reg. (max softmax) & 0.0587 $\pm$ 0.0020 & 0.1268 $\pm$ 0.0024 \\
Isotonic Reg. (entropy) & 0.0590 $\pm$ 0.0020 & 0.1266 $\pm$ 0.0026 \\
Isotonic Reg. (energy) & 0.0579 $\pm$ 0.0026 & 0.1240	$\pm$ 0.0035 \\
Aux. Net (logit input) & 0.0401 $\pm$ 0.0017 & 0.1081 $\pm$ 0.0453 \\
Aux. Net (softmax input) & 0.0396 $\pm$ 0.0003 & \textbf{0.0589 $\pm$ 0.0040} \\
Aux. Net (embedding input) & 0.1251 $\pm$ 0.0123 & 0.0854 $\pm$ 0.0102 \\
\bottomrule
\end{tabular}
\caption{\textbf{Normalized scene entropy cannot be predicted from a single image.} Entries in the table are mean absolute errors averaged over three trials and accompany Figures \ref{fig:ImgNetVid_SUCE}  and \ref{fig:obj_SUCE}. MAE for ImageNetVid-Robust videos are improved, but the plots in Figure \ref{fig:ImgNetVid_SUCE} show that this improvement is still not an accurate estimate of scene uncertainty, because correlation coefficients between empirical and estimated normalized scene entropy are at most $\sim 0.7$. The MAE for Objectron videos appear low compared to the MAE for ImageNetVid-Robust videos because nearly all frames of most videos are correctly classified. The fact that all methods fail to improve upon the typically overconfident softmax vector from a network trained with the cross-entropy loss shows that information about scene uncertainty cannot be obtained without imputation.}
\label{tab:SUCE_table}
\end{table}

\paragraph{Temperature Scaling}
Our temperature scaling procedure consisted of adding one additional temperature parameter to each network and fine-tuning on the anchor frames of a validation dataset for 100 epochs using the cross-entropy loss and the Adam optimizer. The temperature at the epoch that produced the lowest cross-entropy loss was chosen for each trial. Temperatures reported for the three baseline networks on the Objectron dataset were 1.1761, 1.0909, and 1.1715. Temperatures reported for the three baseline networks on the ImageNetVid dataset were 1.3043, 1.3205, and 1.3121.

\begin{figure*}[ht]
    \centering
    \includegraphics[width=1.8in]{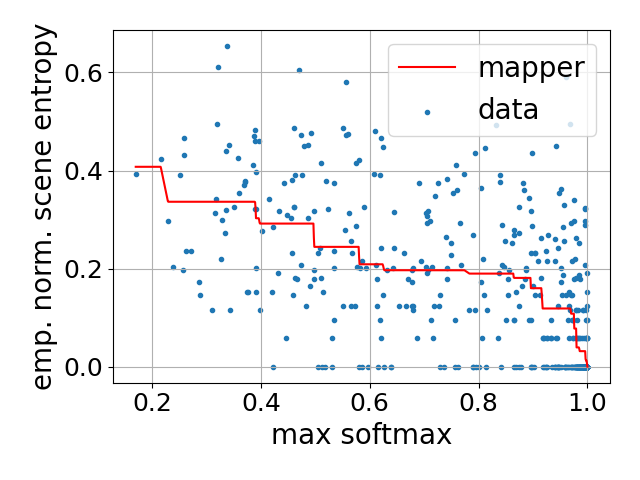}
    \includegraphics[width=1.8in]{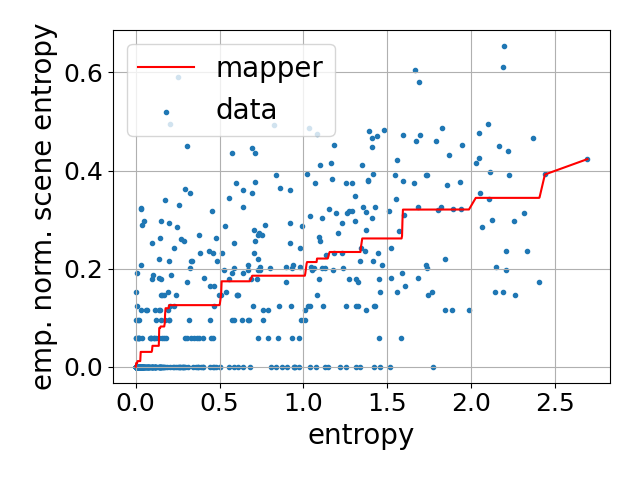}
    \includegraphics[width=1.8in]{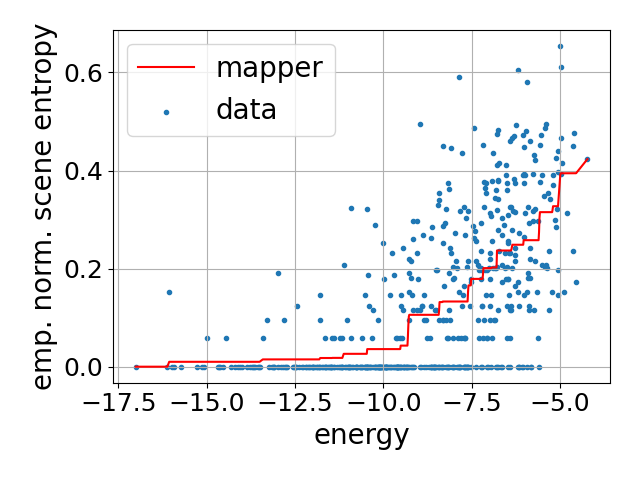}
    \includegraphics[width=1.8in]{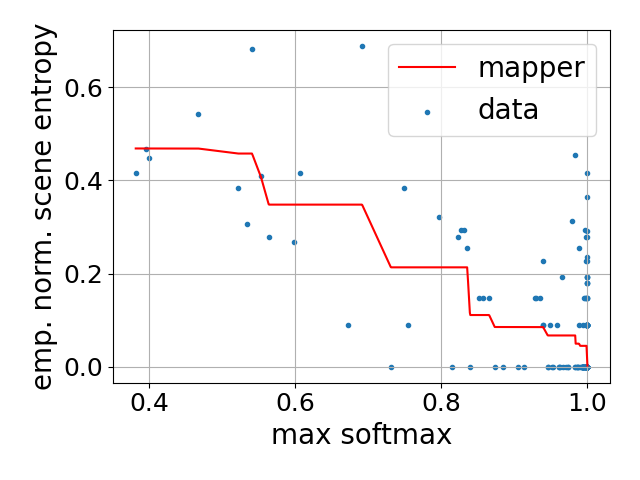}
    \includegraphics[width=1.8in]{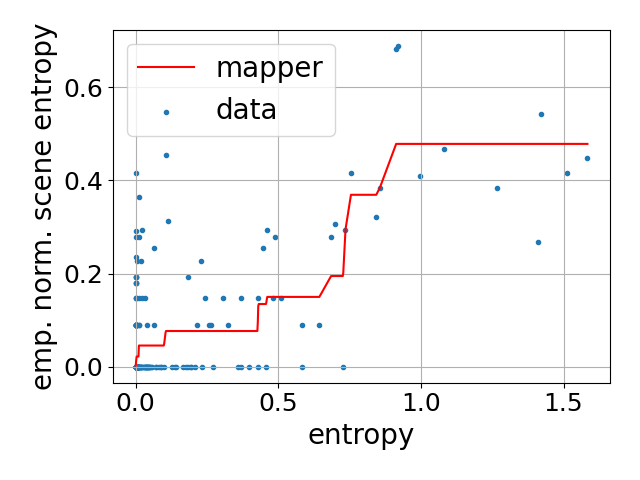}
    \includegraphics[width=1.8in]{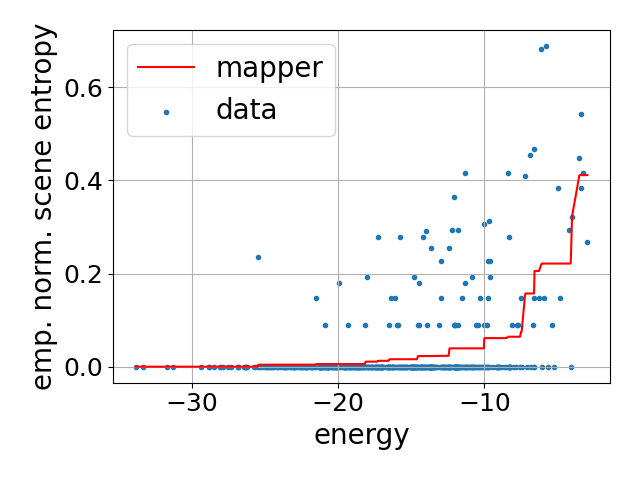}
    \caption{\textbf{Isotonic curves bear little correlation to the data.} In the plots above, each blue dot corresponds to a scene. The x-axis is the value of a simple statistic of the anchor frame's softmax vector (maximum value, entropy, or energy) and the y-axis is the normalized scene entropy computed from the empirical paragon for the scene. Red lines are the learned isotonic functions that attempt to map simple statistics to normalized scene entropy. The top row contains plots for the ImageNetVid validation set and the bottom row contains plots for the Objectron validation set. The red curves do seem to fit the data, i.e. there is some correlation between these simple statistics and empirical scene entropy. However, Table \ref{tab:SUCE_table} shows that these isotonic curves do not generalize well, and have higher MAE than normalized scene entropy computed using the softmax vector.  }
    \label{fig:isotonic_curves}
\end{figure*}

\paragraph{Isotonic Regression}
We isotonically mapped several simple statistics of the softmax vector (the maximum softmax value, entropy, and energy \cite{liu_energy-based_2020}) of the anchor frames of a validation dataset onto the empirical normalized scene entropy. Isotonic curves that reflect the results of Table \ref{tab:SUCE_table} are shown in Figure \ref{fig:isotonic_curves}.

\subsection{Auxiliary Networks}

We trained simple feedforward neural networks $\phi(\cdot)$ to map the softmax vector, logits, or embedding (the output of the second-to-last layer of ResNet) of an anchor frame $x$ to the scene entropy. Networks that mapped logits and softmax vectors had hidden layers with widths 512, 512, 256, 128, 64 and those that mapped embeddings had hidden layers with widths 4096, 2048, 1024, 512, 256. Other hyperparameters, training error, and validation error are shown in Table \ref{tab:auxnet_params}. The loss function $\mathcal L_{aux}$ was the MSE loss between the output of the auxiliary network and the empirical scene uncertainty:
\begin{equation}
    \mathcal{L}_{aux} = \left ( \sigma(\phi(y)) - U_{f_w}(P(\hat k|S(x)) \right )^2 
\end{equation}
where $\phi(\cdot)$ is the auxiliary network, $y$ is its input (logits, softmax vector, or embedding), and $\sigma(\cdot)$ is the sigmoid function. Results listed in Tables \ref{tab:SUCE_table} and \ref{tab:auxnet_params} show that auxiliary networks effectly overfit the training data, but do not generalize to the validation or test dataset. This lack of generalization indicates that normalized scene entropy cannot be estimated using auxiliary networks.

\begin{table*}[ht]
\footnotesize
\centering
\begin{tabular}{lcccccc}
\toprule
& \multicolumn{3}{c}{\textbf{Objectron}} & \multicolumn{3}{c}{\textbf{ImageNetVid}} \\
\midrule
\textbf{Parameter} & \textbf{Logit Input} & \textbf{Softmax Input} & \textbf{Embed. Input} & \textbf{Logit Input} & \textbf{Softmax Input} & \textbf{Embed. Input} \\
\midrule
Weight Decay & 1e-05 & 1e-05 & 1e-04 & 1e-04 & 1e-04 & 1e-05 \\
Selected Epoch & 50, 50, 50 & 50, 50, 50 & 70, 50, 50 & 30, 20, 20, & 60, 20, 30 & 30, 80, 70 \\
Training Loss at Epoch & 0.0025 $\pm$ 0.0002 & 0.0025 $\pm$ 0.0002 & 0.0029 $\pm$ 0.0002 & 0.0047 $\pm$ 0.0001 & 0.0086 $\pm$ 0.0068 & 0.0030 $\pm$ 0.0003 \\
Validation Loss at Epoch & 0.0051 $\pm$ 0.0010 & 0.0049	$\pm$ 0.0012 & 0.0062 $\pm$ 0.0015 & 0.0189 $\pm$ 0.0014 & 0.0123 $\pm$0.0064 & 0.0200 $\pm$ 0.0005 \\
\bottomrule
\end{tabular}
\caption{\textbf{Auxiliary networks do not generalize.} Hyperparameters, training error, and validation error for three separate trials are shown above after up to 100 epochs of Adam with a default initial learning rate of 1e-3. The simple feedforward auxiliary network is able to fit the training data, but does not generalize well to the validation set, which is also reflected in the test set results. These results further imply that the information required to predict normalized scene entropy, let along the Wellington Posterior, is not contained in the output of a single network on a single image.
}
\label{tab:auxnet_params}
\end{table*}

\section{Broader Impact and Licensing}
Uncertainty quantification is a fundamental building block for explainable and interpretable machine learning. The ideas in our paper can help improve uncertainty estimate for the outcome of deep neural network image classifiers, contributing to progress towards reliable models and trustworthy AI. As our work is basic research, it is possible for it to be incorporated in a wide variety of applications. Finally, the datasets and resources we chose to use were explicitly crafted to be neutral and not offensive.

The resources we used and their licenses are listed below:
\begin{itemize}
    \item ImageNetVid Dataset: \\
    \url{https://image-net.org/challenges/LSVRC/2015/index.php}
    \item ImageNetVid-Robust Dataset (BSD-2):\\
    \url{https://modestyachts.github.io/natural-perturbations-website}
    \item Objectron Dataset (C-UDA 1.0)\\ \url{https://github.com/google-research-datasets/Objectron}
    \item VOICED (no commercial use)\\ { \url{https://github.com/alexklwong/unsupervised-depth-completion-visual-inertial-odometry}}
    \item Pretrained Pytorch Models (BSD-3): \\ \url{https://pytorch.org/hub/pytorch_vision_resnet}
\end{itemize}

\end{document}